\documentclass{article}

\usepackage{microtype}
\usepackage{graphicx}
\usepackage{subcaption}
\usepackage{booktabs} 

\usepackage{hyperref}

\usepackage[preprint]{icml2026}

\usepackage{amsmath}
\usepackage{amssymb}
\usepackage{mathtools}
\usepackage{amsthm}

\usepackage[capitalize,noabbrev]{cleveref}

\usepackage{multirow}
\usepackage{multicol}
\usepackage[table]{xcolor}
\usepackage{colortbl}
\usepackage{amssymb}
\usepackage{booktabs}
\usepackage{tabularx} 
\usepackage{makecell} 
\usepackage{xcolor}
\usepackage{pifont}
\usepackage[most]{tcolorbox}
\newcommand{\cmark}{{\bfseries\ding{51}}}
\newcommand{\xmark}{\textcolor{gray!60}{\ding{55}}}

\usepackage[most]{tcolorbox}
\usepackage{listings}
\usepackage{xcolor}

\definecolor{boxblue}{RGB}{15, 85, 180}
\definecolor{titleblue}{RGB}{10, 60, 145}

\lstdefinestyle{mycode}{
  basicstyle=\ttfamily\small,
  numbers=left,
  numberstyle=\scriptsize\color{gray!70},
  numbersep=8pt,
  xleftmargin=2.2em,
  frame=none,
  breaklines=true,
  columns=fullflexible,
}

\newtcolorbox{fancycodebox}[2][]{%
  enhanced,
  colback=white,
  colframe=blue!70!black,
  boxrule=1pt,
  arc=8pt,
  left=6pt,right=6pt,
  top=10pt,bottom=6pt,
  before skip=14pt, 
  after skip=6pt,
  listing only,
  listing options={style=mycode},
  overlay={%
    \node[
      fill=blue!70!black,
      text=white,
      font=\bfseries\small,
      rounded corners=6pt,
      inner xsep=10pt,
      inner ysep=5pt,
      anchor=west
    ] at ([xshift=10pt,yshift=3pt]frame.north west) {#2};
  },
  #1
}

\definecolor{lightblue}{RGB}{150,255,255}
\theoremstyle{plain}

\theoremstyle{definition}

\theoremstyle{remark}

\usepackage[textsize=tiny]{todonotes}

\icmltitlerunning{M3-AD for Industrial Anomaly Detection}

\begin{document}

\twocolumn[
  \icmltitle{M3-AD: Reflection-aware Multi-modal, Multi-category, and Multi-dimensional Benchmark and Framework for Industrial Anomaly Detection}



  \icmlsetsymbol{equal}{*}

  \begin{icmlauthorlist}
    \icmlauthor{Chao Huang}{sysu}
    \icmlauthor{Yanhui Li}{sysu}
    \icmlauthor{Yunkang Cao}{hnu}
    \icmlauthor{Wei Wang}{sysu} \\
    \icmlauthor{Hongxi Huang}{sysu}
    \icmlauthor{Jie Wen}{hit}
    \icmlauthor{Wenqi Ren}{sysu}
    \icmlauthor{Xiaochun Cao}{sysu}
  \end{icmlauthorlist}

  \icmlaffiliation{sysu}{Shenzhen Campus of Sun Yat-sen University}
  \icmlaffiliation{hnu}{Hunan University}
  \icmlaffiliation{hit}{Harbin Institute of Technology, Shenzhen}

  \icmlcorrespondingauthor{Chao Huang}{huangch253@mail.sysu.edu.cn}

  \icmlkeywords{Machine Learning, ICML}
  \begin{center}
    \setkeys{Gin}{width=\textwidth}
    \includegraphics{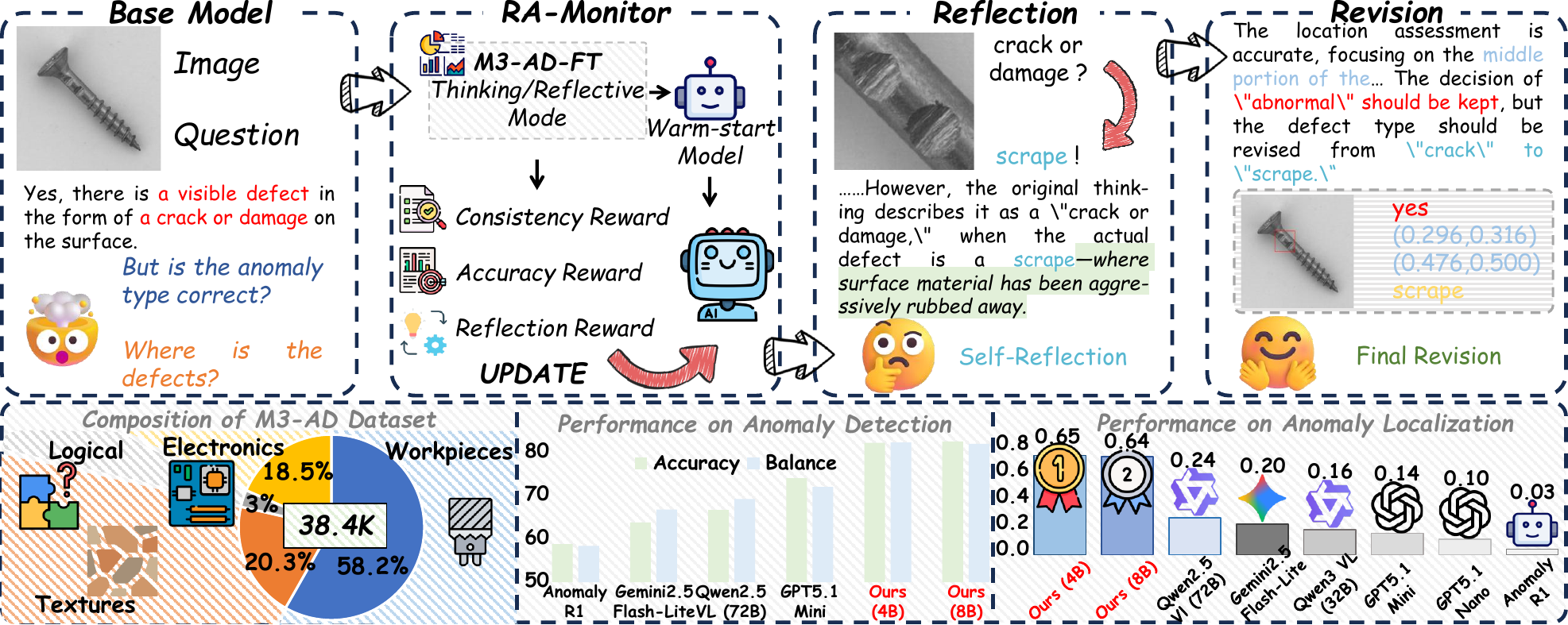}
      \captionof{figure}{%
      M3-AD enables self-correction of unreliable initial predictions through a reflection-aware mechanism, significantly improving anomaly type recognition and spatial localization in industrial anomaly detection compared to base models.%
      }
    \label{teaser}
  \end{center}

  \vskip 0.3in
  
]



\printAffiliationsAndNotice{}  

\begin{abstract}
Although multimodal large language models (MLLMs) have advanced industrial anomaly detection toward a zero-shot paradigm, they still tend to produce high-confidence yet unreliable decisions in fine-grained and structurally complex industrial scenarios, and lack effective self-corrective mechanisms.
To address this issue, we propose M3-AD, a unified reflection-aware multimodal framework for industrial anomaly detection. M3-AD comprises two complementary data resources: M3-AD-FT, designed for reflection-aligned fine-tuning, and M3-AD-Bench, designed for systematic cross-category evaluation, together providing a foundation for reflection-aware learning and reliability assessment.
Building upon this foundation, we propose RA-Monitor, which models reflection as a learnable decision revision process and guides models to perform controlled self-correction when initial judgments are unreliable, thereby improving decision robustness. Extensive experiments conducted on M3-AD-Bench demonstrate that RA-Monitor outperforms multiple open-source and commercial MLLMs in zero-shot anomaly detection and anomaly analysis tasks. Code will be released at \url{https://github.com/Yanhui-Lee/M3-AD}.

\end{abstract}

\section{Introduction}
Industrial anomaly detection is a fundamental problem in intelligent manufacturing and quality control \cite{li2025survey,chandola2009anomaly,vaikundam2016anomaly,liu2024deep,pang2021deep}. Its goal is to accurately determine whether a product is defective in complex and diverse industrial scenarios. In recent years, the rapid development of Multimodal Large Language Models (MLLMs) has introduced new possibilities for industrial anomaly detection \cite{Vlm-r1,visual_Planning,vision-r1}. By leveraging their strong cross-modal perception and reasoning capabilities, prior works have advanced industrial inspection beyond traditional supervised or unsupervised anomaly detection frameworks, which primarily formulate anomaly detection as representation learning or similarity-based discrimination~\cite{winclip,anomalyclip,adaclip,adaptclip,patchcore,defard2021padim}, toward more general and flexible solutions \cite{Anomalygpt,towards_zero-shot_anomaly_detection_and_reasoning,myriad,mmad}. While this trend brings new opportunities to industrial quality inspection, it also raises a critical question: do MLLMs possess sufficiently reliable decision-making capabilities to support deployment in real-world industrial scenarios?

Existing studies have explored the application of MLLMs to industrial anomaly detection by jointly reasoning over visual inputs and language prompts, enabling models to directly output anomaly judgments, descriptive explanations, or localization results \cite{anomalyr1,omniad,li2025iad}. Such approaches demonstrate strong performance in scenarios with simple structures and salient anomalous regions, suggesting that large models exhibit considerable anomaly perception capability. However, through systematic evaluations of multiple mainstream foundation models across diverse industrial scenarios (as shown in \cref{tab:anomaly_detection,tab:anomaly_analysis}), MLLMs perform relatively stably on low-level visual anomalies, yet suffer significant performance degradation in scenarios involving high-level semantic anomalies such as structural relationships and component absence. This observation indicates that the core bottleneck of MLLMs in industrial anomaly detection does not primarily stem from insufficient visual perception, but rather from the absence of mechanisms that can explicitly inspect their own reasoning processes, identify potential decision errors, and correct them accordingly. 

Motivated by these observations, we propose M3-AD, a unified industrial anomaly detection framework that integrates a reflection-aware modeling approach, RA-Monitor, with a structured data resource consisting of M3-AD-FT for reflection-aligned fine-tuning and M3-AD-Bench for systematic evaluation. 
At the data level, M3-AD-FT and M3-AD-Bench are constructed from a unified data pool with language-aligned structures, covering a complete spectrum of anomalies ranging from low-level visual defects to high-level structural anomalies. To support structured supervision and analysis, we design a hierarchical anomaly taxonomy that is consistently adopted in both M3-AD-FT and M3-AD-Bench. Moreover, the data construction is aligned with reflection-aware learning, enabling M3-AD-FT to provide high-quality supervision for reflection-aware post-training. 
At the methodological level, RA-Monitor models reflection as a learnable decision revision process and jointly optimize it through Reflection-Aware Warm Start and Reflection-Cognitive Reinforcement Learning.
At the evaluation level, M3-AD-Bench defines fine-grained evaluation protocols to comprehensively assess models’ zero-shot anomaly detection and anomaly analysis capabilities.

Our main contributions can be summarized as follows:
\begin{itemize}
    \item We propose M3-AD Dataset, a structured data resource consisting of M3-AD-FT for reflection-aligned fine-tuning and M3-AD-Bench for systematic cross-category evaluation.
    \item We propose RA-Monitor, a reflection-aware learning framework that enables MLLMs to revise unreliable initial decisions and improve their decision reliability.
    \item Extensive experiments demonstrate consistent improvements in zero-shot anomaly detection and anomaly analysis across diverse industrial scenarios.
\end{itemize}

\begin{figure*}[t]
  \begin{center}
    \centerline{\includegraphics[width=\linewidth]{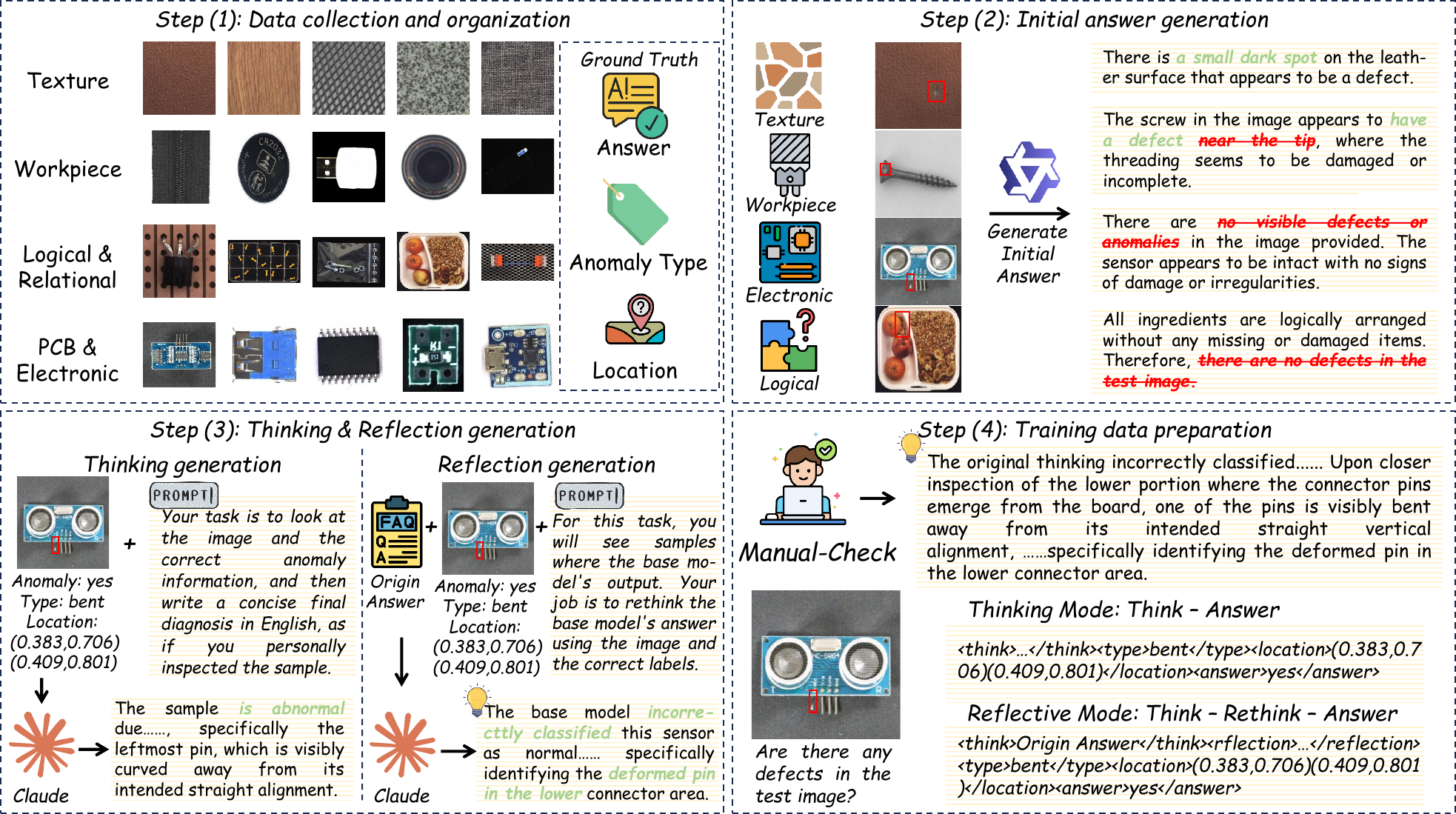}}
    \caption{
      Overview of M3-AD-FT data construction pipeline. The pipeline consists of four stages: (1) collecting and organizing industrial images across multiple scenarios with structured anomaly annotations; (2) classifying data by scenario and generating initial model answers; (3) constructing thinking and reflective captions; (4) preparing training data through manual verification.
    }
    \label{construction}
  \end{center}
\end{figure*}

\section{Related Work}
\subsection{Zero-shot Anomaly Detection}
Zero-shot anomaly detection has developed into a well-established research paradigm under VLMs. Representative CLIP-based approaches, such as WinCLIP \cite{winclip}, as well as later works in this line \cite{adaclip,filo,anomalyclip,Bayesian_Prompt_Flow}, typically determine image normality by designing textual prompts and computing visual–textual similarity, benefiting from weak annotation requirements and strong cross-domain generalization. However, this paradigm essentially formulates anomaly detection as a similarity measurement problem, often lacking explicit interpretability in anomaly judgment and decision-making. With the rapid advancement of MLLMs, an increasing number of studies have begun to apply MLLMs to industrial anomaly detection \cite{Anomalygpt,anomalyr1,lad-reasoner,Logicad,LR-IAD}, leveraging their instruction-following and reasoning capabilities to produce more descriptive decisions, fine-grained anomaly semantics, and interactive explanations. Nevertheless, in zero-shot anomaly detection scenarios, models remain prone to producing highly confident but incorrect predictions when confronted with subtle visual variations or structural ambiguities.
\subsection{Reasoning and Self-Reflection in Large Models}
Recent studies on large language models have begun to explore adaptive inference strategies \cite{PAG,ReCoT,Reflect_Retry_Reward,srpo,vl-rethinker,training_language_models_to_self-correct,Fast-Slow_Thinking_GRPO}, in which models dynamically decide whether to adopt ``fast thinking” or invoke more computationally expensive ``slow thinking,” as well as whether to trigger self-reflection and revision during the reasoning process. One line of work designs confidence based gating mechanisms to control the activation of Chain-of-Thought reasoning, enabling more elaborate inference only when the initial judgment is deemed unreliable \cite{Reasoning_in_Flux,CER}. Another line of research allocates reasoning depth based on input difficulty or feedback signals, or introduces self-critique and iterative revision following an initial response \cite{Adaption-of-Thought,DAST,CoTFormer}, in order to improve decision quality in complex scenarios. However, these approaches are predominantly validated on textual or symbolic reasoning tasks, where the triggering criteria rely on the model’s self-estimated confidence or language-level error signals. Such assumptions do not readily transfer to industrial anomaly detection, which is grounded in visual evidence and is prone to high-confidence perceptual errors. This gap indicates that existing fast–slow thinking and reflection mechanisms remain insufficient for supporting reliable decision evaluation in industrial anomaly detection.

\section{M3-AD Dataset}
Existing fine-tuning datasets \cite{li2025iad,anomalyr1} typically provide only a single reasoning trajectory, forcing models to uniformly apply complex reasoning to all samples and resulting in redundant reasoning.
To address this limitation, we construct the M3-AD Dataset, a unified resource for industrial anomaly detection. 
The M3-AD Dataset is constructed by re-annotating visual images from multiple public industrial anomaly detection datasets \cite{manta,MulSen_AD,mvtec_locoAD,mpdd,dagm,dtd,sdd,mvtec,visa,real-iad,shi2016automatic,mishra21-vt-adl,Bozic2021COMIND,AITEX_AFID_dataset} under a unified anomaly taxonomy. It covers four industrial scenarios: surface textures, industrial workpieces, electronic components, and logical scenes, and consists of M3-AD-FT for reflection-aligned fine-tuning and M3-AD-Bench for systematic cross-category evaluation. More details can be found in Appendix \cref{m3-ad-dataset}.

To enable models to learn when reflection should be triggered, we design the difficulty-aware data construction based on decision correctness. As shown in \cref{construction}, we employ Qwen2.5-VL-72B \cite{bai2025qwen2} as the base model and use its anomaly detection results as the criterion for difficulty assignment. Samples correctly classified by the base model are treated as easy, while misclassified samples are treated as hard. Based on this difficulty partition, M3-AD-FT incorporates two distinct reasoning trajectories into the training data: \textit{Thinking Mode}, which directly produces a prediction, and \textit{Reflective Mode}, which performs reflection before generating the final decision.

For easy samples, the thinking mode is primarily adopted, while 30\% of the samples are assigned to the reflective mode. In this case, reflection is not intended for error correction, but rather for explicitly strengthening the model’s articulation of the evidence underlying correct decisions. 

For hard samples, the reflective mode is dominant, accounting for 70\% of the instances. This design explicitly models the learning process of error–reflection–correction, helping the model understand the causes of incorrect predictions. Meanwhile, a small portion of samples retain the thinking mode as a control, preventing reflection from degenerating into the only decision-making pathway.

\begin{figure*}[htbp]
  \begin{center}
    \centerline{\includegraphics[width=\linewidth]{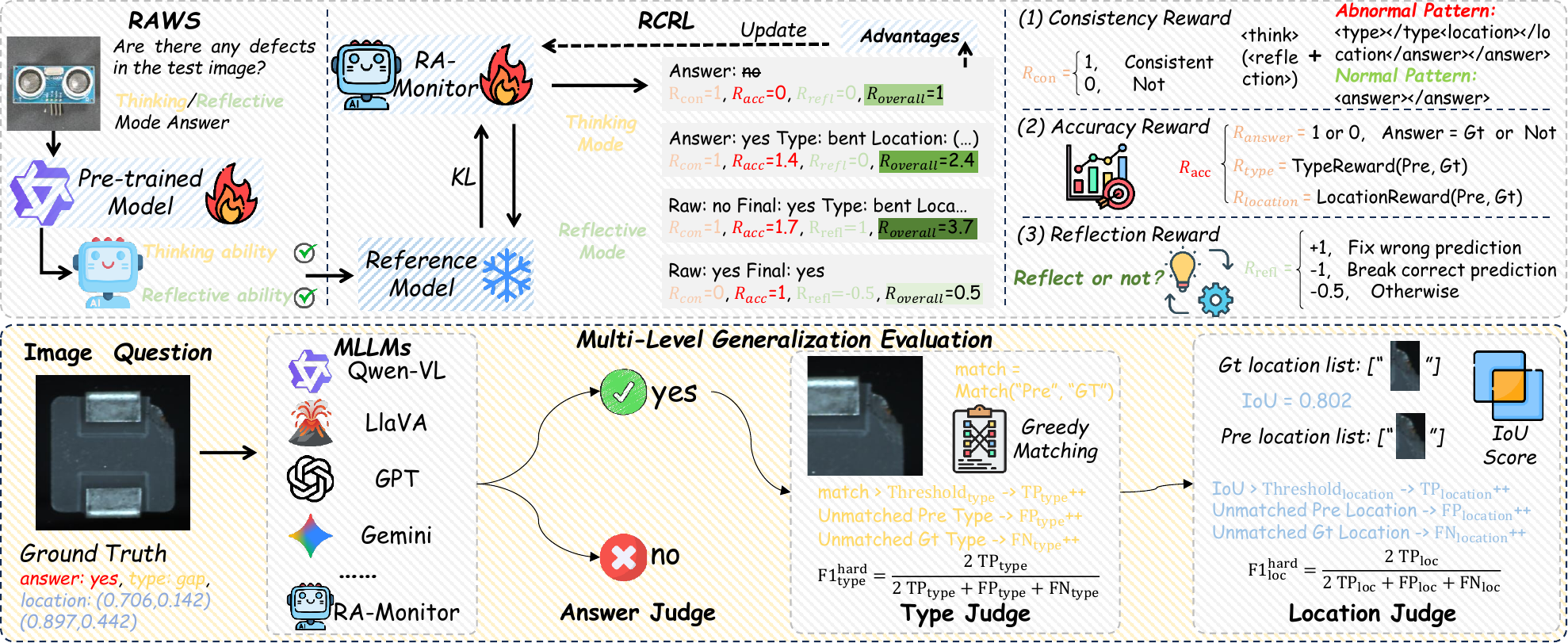}}
    \caption{
      Overview of RA-Monitor. RAWS equips the pre-trained model with both thinking and reflective abilities, while RCRL further optimizes the model via consistency, accuracy, and reflection rewards. The lower part illustrates the unified metric computation used for multi-level evaluation of anomaly detection, type recognition, and localization.
    }
    \label{framework}
  \end{center}
\end{figure*}

\section{RA-Monitor}
\subsection{Reflection-Aware Warm Start}
In the RAWS stage, we perform supervised fine-tuning on the M3-AD-FT. The supervision in RAWS corresponds to two distinct behavioral pathways, namely \textit{Thinking Mode} and \textit{Reflective Mode}.
\textit{Thinking Mode} corresponds to samples that do not contain reflection annotations. In this mode, the \texttt{<think>} field captures a single-pass forward reasoning process based on salient visual cues, enabling the model to directly produce reliable predictions in scenarios where anomaly patterns are clear and visual evidence is sufficient. \textit{Reflective Mode} corresponds to samples that explicitly include reflection annotations. In this case, the \texttt{<think>} field represents the model’s original reasoning and initial prediction, while the \texttt{<reflection>} field describes the model’s analysis and decision revision after identifying potential uncertainty or errors in the initial output.

Given an input image and instruction, the model is trained to generate a structured output sequence, whose specific form depends on the behavioral mode of the sample.
For \textit{Thinking Mode} samples, the output sequence is defined as:
\begin{small}
\begin{equation}
O = (\langle \texttt{think} \rangle, A),
\end{equation}
\end{small}where $\langle \texttt{think} \rangle$ denotes the model’s discriminative reasoning process and $A$ represents the final answers.
For \textit{Reflective Mode} samples, the output sequence is defined as:
\begin{small}
\begin{equation}
O = (\langle \texttt{think} \rangle, \langle \texttt{reflection} \rangle, A),
\end{equation}
\end{small}

where $\langle \texttt{think} \rangle$ corresponds to the initial decision process, and $\langle \texttt{reflection} \rangle$ captures the model’s self-reflection and decision revision behavior.

The training objective of RAWS is to maximize the conditional likelihood of generating the corresponding structured output sequence given the input.
Specifically, the supervised fine-tuning loss is defined as:
\begin{small}
\begin{equation}
\mathcal{L}_{\text{RAWS}}
=
-\mathbb{E}_{(x,O)\sim \mathcal{D}_{\text{M3-AD-FT}}}
\sum_{t=1}^{|O|}
\log \pi_\theta(o_t \mid x, o_{<t}),
\end{equation}
\end{small}where $\mathcal{D}_{\text{M3-AD-FT}}$ denotes the M3-AD-FT data distribution, $x$ represents the input image,
$O = \{o_1, \ldots, o_{|O|}\}$ is the target output sequence, and $\pi_\theta$ denotes the model parameterized by $\theta$.

Under this unified sequence modeling framework, the RAWS objective jointly covers both \textit{Thinking Mode} and \textit{Reflective Mode}, enabling the model to acquire stable initial decision-making capabilities while establishing explicit semantic alignment for reflection-based decision revision.
\subsection{Reflection-Cognitive Reinforcement Learning}
\label{sec:reflection-crl}
In the RCRL stage, we further optimize the model’s decision-making and reflection strategies through reinforcement fine-tuning (RFT). We formulate industrial anomaly detection with reflection as a sequential decision-making problem, and formalize the model output as a policy $\pi_\theta(y \mid x)$, where $\theta$ denotes the model parameters, $x$ represents the input image, and $y$ denotes the output sequence containing structured reasoning traces and anomaly predictions. The objective of RCRL is to maximize the expected reward under the given input:
\begin{small}
\begin{equation}
J(\theta) = \mathbb{E}_{y \sim \pi_\theta(y \mid x)} [R(x, y)].
\end{equation}
\end{small}
The overall reward function $R(x, y)$ evaluates the overall quality of the model output for the input $x$, which is defined as a weighted combination of three components:
\begin{small}
\begin{equation}
R
=
\lambda_c R_{\text{cons}}
+
\lambda_a R_{\text{acc}}
+
\lambda_r R_{\text{refl}},
\end{equation}
\end{small}where $R_{\text{cons}}$, $R_{\text{acc}}$, and $R_{\text{refl}}$ denote the consistency reward, accuracy reward, and reflection reward, respectively. $\lambda_c$, $\lambda_a$, and $\lambda_r$ are the corresponding weighting coefficients.

\textbf{The consistency reward} enforces semantic coherence between the reasoning process and the final decision. We define a minimum structural constraint requiring all outputs to explicitly generate reasoning and final decision fields, corresponding to the tags \texttt{<think>(<reflection>)} and \texttt{<answer>}. For normal samples, the model output should not include anomaly-related descriptions. For anomalous samples, beyond satisfying the minimum structure, the model is required to provide anomaly type \texttt{<type>} and spatial localization information \texttt{<location>}.

\textbf{The accuracy reward} encourages correct anomaly detection and anomaly analysis. It adopts a gated design: only when the model correctly predicts anomaly existence is it further rewarded for anomaly type and localization predictions. Formally, the accuracy reward is defined as:
\begin{small}
\begin{equation}
R_{\text{acc}}(y,g)
=
R_{\text{ans}}(y,g)
+
\frac{1}{2}\Big(R_{\text{type}}(y,g)+R_{\text{loc}}(y,g)\Big),
\end{equation}
\end{small}where $y$ denotes the model output text and $g$ denotes the ground-truth annotation. Here, $R_{\text{ans}}(y,g)$ evaluates the correctness of anomaly detection, while $R_{\text{type}}(y,g)$ and $R_{\text{loc}}(y,g)$ reward the model’s anomaly analysis capability in terms of type identification and spatial localization.

\textbf{The reflection reward} explicitly evaluates whether introducing reflection leads to a meaningful improvement over the initial prediction. Let $y_0 \in \{0,1\}$ denote the prediction obtained before reflection, and $y_1 \in \{0,1\}$ denote the final prediction after reflection, where $1$ indicates anomaly and $0$ indicates normal, and let $y$ denote the ground-truth label. The reflection reward is defined based on the change in prediction correctness:

\begin{small}
\begin{equation}
R_{\text{refl}}(y_0,y_1,y)=
\begin{cases}
+1.0, & y_0 \neq y \;\land\; y_1 = y, \\
-1.0, & y_0 = y \;\land\; y_1 \neq y, \\
-0.5, & y_0 = y \;\land\; y_1 = y, \\
-0.5, & y_0 \neq y \;\land\; y_1 \neq y.
\end{cases}
\end{equation}
\end{small}

\begin{table*}[t]
  \caption{Accuracy and balanced accuracy. The best and second-best results are highlighted in \textbf{bold} and \underline{underlined}, respectively.}
  \label{tab:anomaly_detection}
  \begin{center}
    \begin{small}
      \begin{sc}
        \begin{tabular}{lcc cc cc cc cc}
          \toprule
          & & & 
          \multicolumn{8}{c}{Industrial Scene} \\
          \cmidrule(lr){4-11}
          Category & Model & Param &
          \multicolumn{2}{c}{Texture} &
          \multicolumn{2}{c}{Workpiece} &
          \multicolumn{2}{c}{Electronic} &
          \multicolumn{2}{c}{Average} \\
          \cmidrule(lr){4-5}
          \cmidrule(lr){6-7}
          \cmidrule(lr){8-9}
          \cmidrule(lr){10-11}
          & & &
          Acc & Bal &
          Acc & Bal &
          Acc & Bal &
          Acc & Bal \\
          \midrule
          \multirow{3}{*}{Commercial}
            & GPT-5.1-nano        & /  & 83.3 & 83.2 & 68.2 & 67.7 & 63.5 & 61.9 & 71.7 & 70.9 \\
            & GPT-5.1-mini       & /  & 83.9 & 84.2 & 53.6 & 56.3 & 59.6 & 63.9 & 65.7 & 68.1 \\
            & Gemini-2.5-Flash-Lite  & /  & 81.1 & 81.5 & 56.4 & 59.0 & 51.5 & 56.9 & 63.0 & 65.8 \\
          \midrule
          \multirow{11}{*}{Open-source}
            & Anomaly-R1   & 3B & 65.8 & 65.5 & 55.5 & 55.5 & 53.6 & 52.6 & 58.3 & 57.9 \\
            & \multirow[c]{2}{*}{Qwen-2-VL-Instruct}   & 2B & 41.9 & 41.4 & 47.6 & 46.2 & 46.1 & 42.7 & 45.2 & 43.4 \\
            &                      & 7B & 64.3 & 64.1 & 59.1 & 59.1 & 55.7 & 53.8 & 59.7 & 59.0 \\
            & \multirow[c]{2}{*}{Qwen-2.5-VL-Instruct} & 3B & 66.1 & 65.8 & 57.1 & 56.2 & 55.7 & 52.9 & 59.6 & 58.3 \\
            &                      & 7B & 76.4 & 75.9 & 60.5 & 58.8 & 57.0 & 52.3 & 64.6 & 62.3 \\
            & LlaVA-OneVision-SI   & 7B & 68.1 & 67.6 & 59.6 & 57.7 & 61.2 & 55.9 & 63.0 & 60.4 \\
            & \multirow[c]{2}{*}{Intern-VL-3.5}   & 4B & 60.7 & 61.5 & 51.0 & 53.8 & 45.6 & 52.8 & 52.4 & 56.0 \\
            &                      & 8B & 77.8 & 77.6 & 57.0 & 58.1 & 56.8 & 58.9 & 63.9 & 64.9 \\
            & \multirow[c]{3}{*}{Qwen-3-VL-Instruct}   & 4B & 83.1 & 83.1 & 70.0 & 69.4 & 64.6 & 63.7 & 72.6 & 72.1 \\
            &                      & 8B & 71.1 & 70.5 & 65.3 & 63.2 & 63.3 & 57.3 & 66.6 & 63.7 \\
            &                      & 32B & 87.4 & 87.3 & 73.7 & \underline{73.8} & \underline{78.6} & \textbf{77.5} & 79.9 & 79.5 \\
            & Qwen-2.5-VL-Instruct & 72B & 80.6 & 80.4 & 70.3 & 69.2 & 66.8 & 62.6 & 72.6 & 70.7 \\
            & \multirow{2}{*}{IAD-R1} & 3B & 85.6 & 85.5 & 66.4 & 64.7 & 75.7 & 69.7 & 75.9 & 73.3\\
            & & 7B & 87.0 & 87.0 & \underline{74.1} & 71.6 & 77.0 & 71.6 & 79.4 & 76.7\\
          \midrule
          \multirow{2}{*}{Thinking}
            & \multirow[c]{2}{*}{Qwen-3-VL-Thinking} & 4B & 67.3 & 68.0 & 53.9 & 55.9 & 41.5 & 48.1 & 54.2 & 57.3 \\
            &                    & 8B & 86.0 & 86.0 & 54.7 & 56.7 & 40.0 & 46.0 & 60.2 & 62.9 \\
          \midrule
          \multirow{2}{*}{RA-Monitor}
            & \multirow[c]{2}{*}{Qwen-3-VL-Instruct} & \cellcolor{lightblue!25}4B & \cellcolor{lightblue!25}\textbf{91.2} & \cellcolor{lightblue!25}\textbf{91.1} & \cellcolor{lightblue!25}72.5 & 
            \cellcolor{lightblue!25}72.9 & \cellcolor{lightblue!25}77.1 & \cellcolor{lightblue!25}77.1 & \cellcolor{lightblue!25}\underline{80.3} & \cellcolor{lightblue!25}\textbf{80.4} \\
            &  & \cellcolor{lightblue!25}8B & \cellcolor{lightblue!25}\underline{88.3} & \cellcolor{lightblue!25}\underline{88.2} & \cellcolor{lightblue!25}\textbf{74.3} & \cellcolor{lightblue!25}\textbf{74.7} & \cellcolor{lightblue!25}\textbf{79.1} & \cellcolor{lightblue!25}\underline{77.4} & \cellcolor{lightblue!25}\textbf{80.6} & \cellcolor{lightblue!25}\underline{80.1} \\
          \bottomrule
        \end{tabular}
      \end{sc}
    \end{small}
  \end{center}
\end{table*}

\begin{table}[t]
    \centering
    \caption{Effectiveness of reflection-aware warm-start data.}
    \label{tab:warm_start_ablation}
    \small
    \begin{tabular}{lcccc}
        \toprule
        Model & Accuracy & Balance & Type & Loc \\
        \midrule
        \rowcolor{gray!15}
        \multicolumn{5}{l}{\textbf{Qwen-3-VL-Instruct-4B}} \\
        \midrule
         None & 72.6 & 72.1 & 0.326 & 0.140 \\
         Thinking Mode Only & 74.5 & 75.6 & 0.452 & 0.597 \\
         Reflective Mode & \textbf{75.7} & \textbf{77.1} & \textbf{0.462} & \textbf{0.611} \\
        \midrule
        \rowcolor{gray!15}
        \multicolumn{5}{l}{\textbf{Qwen-3-VL-Instruct-8B}} \\
        \midrule
         None & 66.6 & 63.7 & 0.271 & 0.274 \\
        Thinking Mode Only& 73.7 & 74.7 & 0.442 & 0.525 \\
        Reflective Mode & \textbf{77.4} & \textbf{78.3} & \textbf{0.478} & \textbf{0.603} \\
        \bottomrule
    \end{tabular}
\end{table}

\begin{table}[t]
\centering
\small
\caption{Ablation study on reward composition.}
\label{tab:reward_ablation}
\begin{tabular}{ccc cccc}
\toprule
$R_{\text{con}}$ & $R_{\text{acc}}$ & $R_{\text{ref1}}$ &
Accuracy & Balance & Type & Loc \\
\midrule
\xmark & \xmark & \xmark & 75.7 & 77.1 & 0.462 & 0.611 \\
\xmark & \xmark & \cmark & 76.0 & 76.9 & 0.503 & 0.605 \\
\xmark & \cmark & \xmark & 77.7 & 78.8 & \textbf{0.549} & 0.631 \\
\xmark & \cmark & \cmark & 74.3 & 75.1 & 0.548 & 0.619  \\
\cmark & \xmark & \xmark & 77.9 & 78.8 & 0.497 & 0.588 \\
\cmark & \xmark & \cmark & 75.6 & 77.1 & 0.448 & 0.625 \\
\cmark & \cmark & \xmark & 76.1 & 77.5 & 0.524 & 0.635 \\
\cmark & \cmark & \cmark & \textbf{80.3} & \textbf{80.4} & 0.527 & \textbf{0.653} \\
\bottomrule
\end{tabular}
\end{table}

\begin{table*}[t]
  \caption{Fine-grained anomaly analysis performance. The best and second-best results are highlighted in \textbf{bold} and \underline{underlined}, respectively.}
  \label{tab:anomaly_analysis}
  \begin{center}
    \begin{small}
      \begin{sc}
        \begin{tabular}{lcc cc cc cc cc}
          \toprule
          & & &
          \multicolumn{8}{c}{Industrial Scene} \\
          \cmidrule(lr){4-11}
          Category & Model & Param &
          \multicolumn{2}{c}{Texture} &
          \multicolumn{2}{c}{Workpiece} &
          \multicolumn{2}{c}{Electronic} &
          \multicolumn{2}{c}{Average} \\
          \cmidrule(lr){4-5}
          \cmidrule(lr){6-7}
          \cmidrule(lr){8-9}
          \cmidrule(lr){10-11}
          & & &
          Type & Loc &
          Type & Loc &
          Type & Loc &
          Type & Loc \\
          \midrule
          \multirow{3}{*}{Commercial}
            & GPT-5.1-nano        & /  & 0.425 & 0.173 & 0.356 & 0.088 & 0.340 & 0.048 & 0.374 & 0.103 \\
            & GPT-5.1-mini       & /  & \underline{0.523} & 0.240 & 0.481 & 0.112 & 0.489 & 0.072 & 0.498 & 0.141 \\
            & Gemini-2.5-Flash-Lite  & /  & \textbf{0.556} & 0.194 & 0.422 & 0.211 & 0.354 & 0.198 & 0.444 & 0.201 \\
          \midrule
          \multirow{11}{*}{Open-source}
            & Anomaly-R1   & 3B & 0.320 & 0.056 & 0.178 & 0.034 & 0.120 & 0.017 & 0.206 & 0.036 \\
            & \multirow[c]{2}{*}{Qwen-2-VL-Instruct}   & 2B & 0.007 & 0.002 & 0.019 & 0.005 & 0.010 & 0.003 & 0.012 & 0.003 \\
            &                                        & 7B & 0.234 & 0.125 & 0.149 & 0.101 & 0.125 & 0.051 & 0.169 & 0.092 \\
            & \multirow[c]{2}{*}{Qwen-2.5-VL-Instruct} & 3B & 0.327 & 0.046 & 0.159 & 0.030 & 0.123 & 0.023 & 0.203 & 0.033 \\
            &                                        & 7B & 0.332 & 0.050 & 0.209 & 0.023 & 0.133 & 0.020 & 0.225 & 0.031 \\
            & LlaVA-OneVision-SI  & 7B & 0.387 & 0.099 & 0.186 & 0.051 & 0.152 & 0.027 & 0.242 & 0.059 \\
            & \multirow[c]{2}{*}{Intern-VL-3.5}   & 4B & 0.454 & 0.426 & 0.346 & 0.058 & 0.285 & 0.045 & 0.362 & 0.176 \\
            &                                        & 8B & 0.406 & 0.053 & 0.320 & 0.033 & 0.255 & 0.021 & 0.327 & 0.036 \\
            & \multirow[c]{3}{*}{Qwen-3-VL-Instruct}   & 4B & 0.312 & 0.010 & 0.385 & 0.166 & 0.280 & 0.154 & 0.326 & 0.110 \\
            &                                        & 8B & 0.348 & 0.424 & 0.289 & 0.217 & 0.176 & 0.180 & 0.271 & 0.274 \\
            &                                        & 32B & 0.325 & 0.444 & 0.399 & 0.016 & 0.381 & 0.032 & 0.368 & 0.164 \\
            & Qwen-2.5-VL-Instruct & 72B & 0.292 & 0.347 & 0.269 & 0.226 & 0.203 & 0.160 & 0.255 & 0.244 \\
          \midrule
          \multirow{2}{*}{Thinking}
            & \multirow[c]{2}{*}{Qwen-3-VL-Thinking} & 4B & 0.464 & 0.190 & 0.302 & 0.167 & 0.280 & 0.111 & 0.349 & 0.156 \\
            &                                      & 8B & 0.415 & 0.290 & 0.333 & 0.216 & 0.278 & 0.150 & 0.342 & 0.219 \\
          \midrule
          \multirow{2}{*}{RA-Monitor}
            & \multirow[c]{2}{*}{Qwen-3-VL-Instruct} & \cellcolor{lightblue!25}4B & \cellcolor{lightblue!25}0.497 & \cellcolor{lightblue!25}\textbf{0.788} & \cellcolor{lightblue!25}\underline{0.498} & \cellcolor{lightblue!25}\underline{0.579} & \cellcolor{lightblue!25}\textbf{0.587} & \cellcolor{lightblue!25}\textbf{0.591} & \cellcolor{lightblue!25}\textbf{0.527} & \cellcolor{lightblue!25}\textbf{0.653} \\
            &                                      & \cellcolor{lightblue!25}8B & \cellcolor{lightblue!25}0.490 & \cellcolor{lightblue!25}\underline{0.764} & \cellcolor{lightblue!25}\textbf{0.522} & \cellcolor{lightblue!25}\textbf{0.607} & \cellcolor{lightblue!25}\underline{0.523} & \cellcolor{lightblue!25}\underline{0.571} & \cellcolor{lightblue!25}\underline{0.512} & \cellcolor{lightblue!25}\underline{0.647} \\
          \bottomrule
        \end{tabular}
      \end{sc}
    \end{small}
  \end{center}
\end{table*}
This design introduces an explicit reflection cost, granting positive reward only when reflection successfully corrects an initially incorrect prediction. More details of the design can be found in Appendix \cref{sec:reflection_reawrd}.

Building on the above formulation, RCRL is expressed as a reward-regularized policy optimization problem with prior constraints. Let $\mathcal{X}$ denote the input space of industrial images and $\mathcal{Y}$ denote the output space of structured reasoning sequences and anomaly predictions. Using the model trained in the RAWS stage as a reference policy $\pi_{\text{ref}}$, the optimization objective is defined as:
\begin{small}
\begin{equation}
\begin{aligned}
\pi^\star
=
&\arg\max_{\pi}
\mathbb{E}_{x \sim \mathcal{D}_{\text{M3-AD-FT}}}
\\
&\mathbb{E}_{y \sim \pi(y \mid x)}
\big[ R(x, y) \big]
-
\beta \cdot
D_{\mathrm{KL}}\!\left(
\pi(y \mid x) \,\|\, \pi_{\text{ref}}(y \mid x)
\right),
\end{aligned}
\end{equation}
\end{small}where the KL divergence term constrains the current policy to remain close to the reference policy, and $\beta$ controls the trade-off between reward maximization and policy stability. We adopt a reflection-aware GRPO algorithm to sample multiple candidate outputs and update the policy using the best-performing samples. Through RCRL, the model gradually learns to actively revise unreliable initial predictions while avoiding unnecessary reflection in simple or high-confidence scenarios.

\section{Experiment}
\subsection{Experiment Setup}
We employ Qwen-3VL-4B-Instruct and Qwen-3VL-8B-Instruct \cite{bai2025qwen3vltechnicalreport} as the base models.
In the RAWS stage, we perform full-parameter fine-tuning with a learning rate of $1\times10^{-5}$, using a cosine learning rate scheduler with 100 warm-up steps. The per-device batch size is set to 1, and an effective larger batch size is achieved via gradient accumulation over 2 steps. In the RCRL stage, we adopt GRPO as the advantage estimator. For each input prompt, we sample 4 trajectories with a temperature of 1.0 and top-$p$ set to 1.0. Rewards are computed using the task-specific reward functions defined in \cref{sec:reflection-crl}. All experiments are conducted using PyTorch on a computing platform equipped with 4 $\times$ NVIDIA A100 80G GPUs. Details of the compared methods and evaluation metrics are provided in \cref{detail_experiments} and \cref{detail_inference}.

\subsection{Main Results}
We evaluate different models on M3-AD-Bench from two perspectives: anomaly detection and anomaly analysis. The anomaly detection results are reported in \cref{tab:anomaly_detection}. Most models achieve strong performance on Texture scenes, but their performance drops significantly on structurally complex scenarios such as Workpiece and Electronic. This indicates that, in realistic industrial settings, relying solely on generic vision–language capabilities is insufficient for achieving stable and reliable anomaly detection. \cref{tab:anomaly_analysis} summarizes the results on anomaly analysis. Compared with binary anomaly detection, model performance on anomaly type recognition and anomaly localization is substantially limited, particularly for localization metrics. This suggests that although current models may produce anomaly judgments, they often fail to accurately characterize the semantic categories and spatial regions of anomalies.

Notably, thinking-based models bring only marginal improvements in anomaly detection, while their gains in anomaly analysis remain unstable. In contrast, models fine-tuned with our proposed RA-Monitor achieve the best overall performance on both anomaly detection and anomaly analysis. These results indicate that extending the reasoning process alone does not effectively enhance model capability. By incorporating reflection-aware optimization, the model is able to correct unreliable reasoning outcomes and achieve more consistent alignment between anomaly semantics and spatial evidence, thereby improving both anomaly detection and anomaly analysis performance.

\begin{figure}[t]
  \begin{center}
    \centerline{\includegraphics[width=\linewidth]{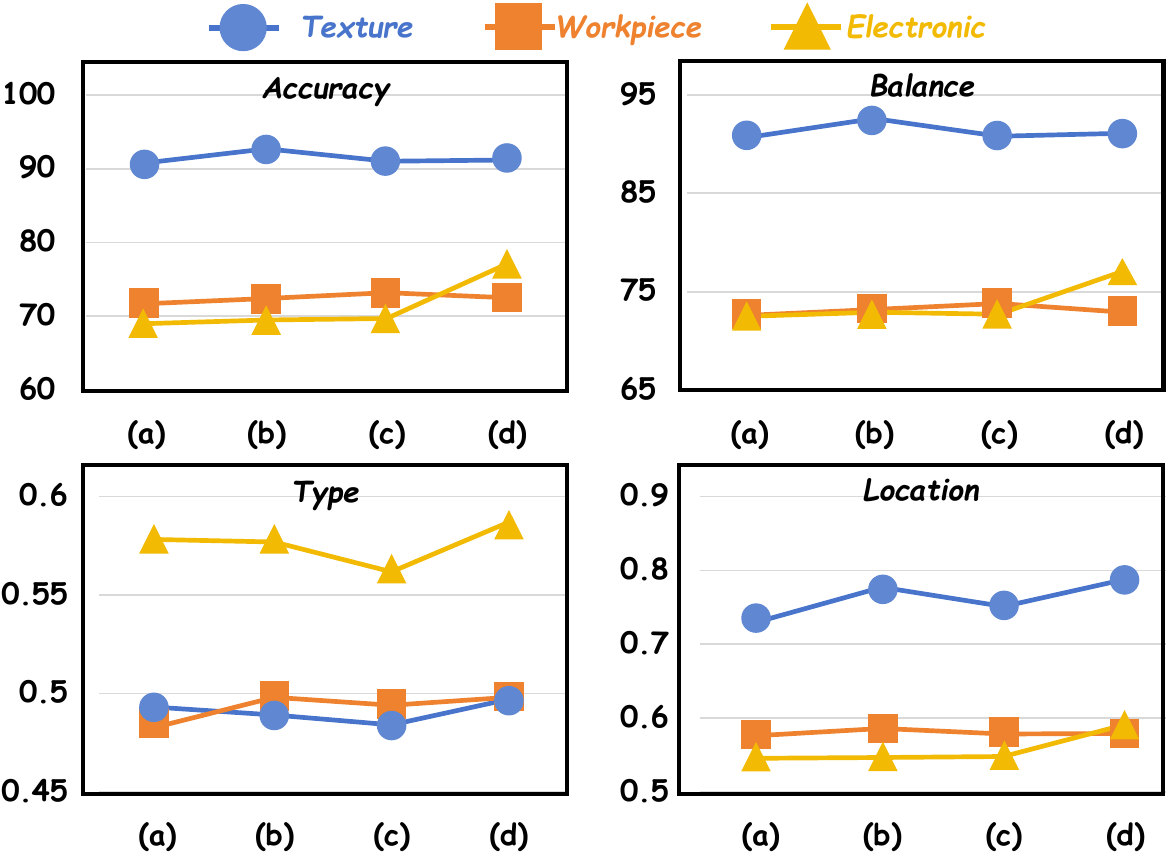}}
    \caption{
      Ablation of reflection reward.
    }
    \label{refl_ablation}
  \end{center}
\end{figure}

\begin{figure}[t]
  \begin{center}
    \centerline{\includegraphics[width=\linewidth]{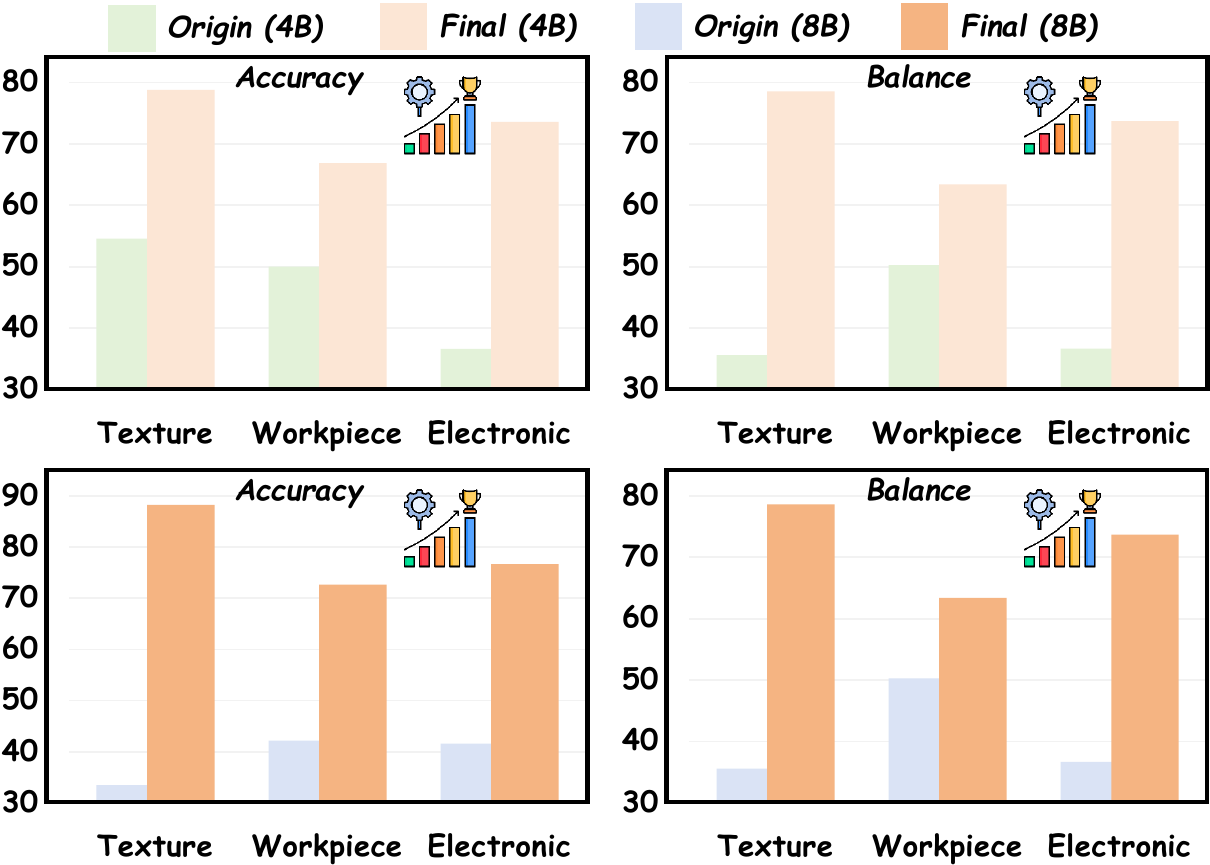}}
    \caption{
      Performance improvement after reflection.
    }
    \label{improverment}
  \end{center}
\end{figure}

\begin{figure*}[htbp]
  \begin{center}
    \centerline{\includegraphics[width=\linewidth]{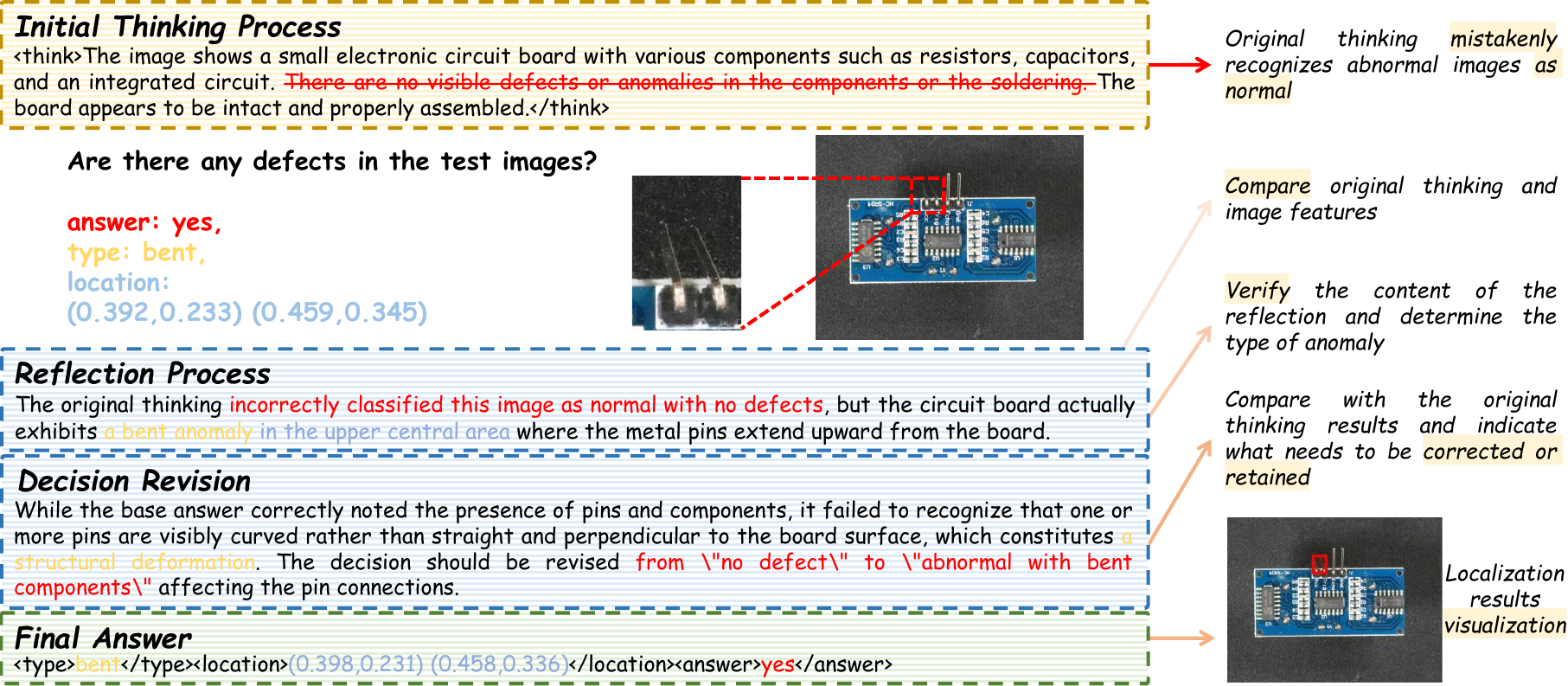}}
    \caption{
      Case study of RA-Monitor output.
    }
    \label{visualization}
  \end{center}
\end{figure*}

\subsection{Ablation Results}
\textbf{Warm-Start Data Ablation.}
The effectiveness of Reflective Mode is investigated in \cref{tab:warm_start_ablation}. None denotes the base model without fine-tuning, Thinking Mode Only indicates fine-tuning using data that contain only the thinking process, while Reflective Mode further proposes conditional decision revision trajectories for samples where stable judgments cannot be obtained from thinking alone. The results show that using thinking data alone significantly improves both anomaly detection and anomaly analysis performance; in comparison, incorporating reflection-based corrections leads to further performance gains. These findings suggest that Reflective Mode not only strengthens the model’s initial reasoning capability, but also explicitly models error identification and self-correction, thereby preventing early decision biases from being propagated into the final predictions.

\textbf{Reward Ablation.}
\cref{tab:reward_ablation} presents the ablation study on reward composition.
The consistency reward $R_{\text{con}}$ helps stabilize model decisions. By comparing settings with and without the accuracy reward $R_{\text{acc}}$, we observe that $R_{\text{acc}}$ not only improves anomaly detection accuracy but also significantly enhances performance in terms of both type classification and localization.
Building upon this, introducing the reflection reward $R_{\text{ref}}$ further strengthens the model's fine-grained anomaly analysis capability.
When all three rewards are combined, the model achieves the best performance across both anomaly detection and anomaly analysis metrics, demonstrating their complementarity.
Further ablation on the reflection reward is shown in \cref{refl_ablation}.
Configurations that provide only positive incentives ((a), (b), and (c)) tend to induce redundant or ineffective reflections, whereas incorporating both a reflection cost and penalties for erroneous reflections ((d)) effectively constrains reflection triggering, guiding the model to perform beneficial self-corrections only when necessary and yielding more robust detection and analysis performance in complex scenarios.

\begin{table}[htbp]
    \centering
    \caption{Sensitivity analysis on IoU threshold.}
    \label{threshold}
    \small
    \begin{tabular}{ccccccc}
        \toprule
        \multirow[c]{2}{*}{IoU} & \multicolumn{2}{c}{Texture} & \multicolumn{2}{c}{Workpiece} & \multicolumn{2}{c}{Electronic} \\
            & 4B & 8B & 4B & 8B & 4B & 8B  \\
        \midrule
        0.1 & 0.856 & 0.833 & 0.650 & 0.678 & 0.675 & 0.638 \\
        0.2 & 0.843 & 0.816 & 0.616 & 0.646 & 0.616 & 0.600 \\
        0.3 & 0.788 & 0.764 & 0.579 & 0.607 & 0.591 & 0.571 \\
        0.4 & 0.693 & 0.664 & 0.543 & 0.568 & 0.560 & 0.544 \\
        0.5 & 0.577 & 0.546 & 0.497 & 0.514 & 0.523 & 0.503 \\
        \bottomrule
    \end{tabular}
\end{table}

\textbf{Performance Gain from Reflection.}
As shown in \cref{improverment}, we compare the accuracy and balanced accuracy of the model before and after reflection following RA-Monitor fine-tuning. Across the three industrial scenarios, incorporating reflection consistently improves detection performance, with more pronounced gains observed in complex scenes such as Workpiece and Electronic.

\textbf{Effect of IoU Threshold Selection.}
We investigate whether the localization results are sensitive to the choice of IoU threshold in hard-F1–based localization evaluation. \cref{threshold} shows that while absolute hard-F1 scores change with the IoU threshold, the relative performance relationships among models remain stable, demonstrating that our localization results do not rely on a specific threshold setting.

\subsection{Case Study}
As shown in \cref{visualization}, we present a case study of the model’s output. The model initially misclassifies the sample as normal, but during the reflection stage, it revisits the original analysis by examining fine-grained visual features in the image. This process enables the model to identify a bent pin located in the upper-middle region of the component and accordingly revise its initial decision, resulting in a correct final prediction. Additional cases can be found in \cref{supply_case}.

\section{Conclusion}
We propose M3-AD, a reflection-aware multimodal unified framework for industrial anomaly detection, which comprises a reflection-aligned fine-tuning dataset (M3-AD-FT), a systematic evaluation benchmark (M3-AD-Bench), and a learnable reflection-based decision correction module (RA-Monitor). M3-AD models reflection requirements at the data and methodological levels, enabling models to perform self-correction when initial predictions are unreliable. Experimental results demonstrate that our approach achieves state-of-the-art performance on zero-shot anomaly detection and anomaly analysis tasks, providing a unified training and evaluation foundation for reliable and interpretable multimodal systems in industrial anomaly detection.

\section*{Impact Statement}
This paper presents M3-AD, which aims to advance the reliability and interpretability of multimodal large language models in industrial anomaly detection. By constructing a systematic cross-category evaluation benchmark and reflection-aware training data, and by introducing a learnable decision correction mechanism, M3-AD enables models to identify unreliable initial predictions and perform controlled self-correction in complex industrial scenarios, thereby improving the stability and robustness of anomaly detection and analysis results. From an application perspective, this work has the potential to support more reliable intelligent assistance in industrial quality inspection, equipment monitoring, and production safety. By reducing false positives and false negatives, it may help mitigate potential economic losses and safety risks in practical industrial settings.
\bibliography{main}

@article{Fast-Slow_Thinking_GRPO,
  title={Fast-slow thinking for large vision-language model reasoning},
  author={Xiao, Wenyi and Gan, Leilei and Dai, Weilong and He, Wanggui and Huang, Ziwei and Li, Haoyuan and Shu, Fangxun and Yu, Zhelun and Zhang, Peng and Jiang, Hao and others},
  journal={arXiv preprint arXiv:2504.18458},
  year={2025}
}

@inproceedings{Reasoning_in_Flux,
  title={Reasoning in flux: Enhancing large language models reasoning through uncertainty-aware adaptive guidance},
  author={Yin, Zhangyue and Sun, Qiushi and Guo, Qipeng and Zeng, Zhiyuan and Li, Xiaonan and Dai, Junqi and Cheng, Qinyuan and Huang, Xuan-Jing and Qiu, Xipeng},
  booktitle={Proceedings of the 62nd Annual Meeting of the Association for Computational Linguistics (Volume 1: Long Papers)},
  pages={2401--2416},
  year={2024}
}

@article{CER,
  title={Cer: Confidence enhanced reasoning in llms},
  author={Razghandi, Ali and Hosseini, Seyed Mohammad Hadi and Baghshah, Mahdieh Soleymani},
  journal={arXiv preprint arXiv:2502.14634},
  year={2025}
}

@inproceedings{Adaption-of-Thought,
  title={Adaption-of-thought: Learning question difficulty improves large language models for reasoning},
  author={Xu, Mayi and Li, Yongqi and Sun, Ke and Qian, Tieyun},
  booktitle={Proceedings of the 2024 Conference on Empirical Methods in Natural Language Processing},
  pages={5468--5495},
  year={2024}
}

@article{DAST,
  title={Dast: Difficulty-adaptive slow-thinking for large reasoning models},
  author={Shen, Yi and Zhang, Jian and Huang, Jieyun and Shi, Shuming and Zhang, Wenjing and Yan, Jiangze and Wang, Ning and Wang, Kai and Liu, Zhaoxiang and Lian, Shiguo},
  journal={arXiv preprint arXiv:2503.04472},
  year={2025}
}

@article{CoTFormer,
  title={CoTFormer: A Chain-of-Thought Driven Architecture with Budget-Adaptive Computation Cost at Inference},
  author={Mohtashami, Amirkeivan and Pagliardini, Matteo and Jaggi, Martin},
  journal={arXiv preprint arXiv:2310.10845},
  year={2023}
}

@inproceedings{adaclip,
  title={Adaclip: Adapting clip with hybrid learnable prompts for zero-shot anomaly detection},
  author={Cao, Yunkang and Zhang, Jiangning and Frittoli, Luca and Cheng, Yuqi and Shen, Weiming and Boracchi, Giacomo},
  booktitle={European Conference on Computer Vision},
  pages={55--72},
  year={2024},
  organization={Springer}
}

@article{adaptclip,
  title={AdaptCLIP: Adapting CLIP for Universal Visual Anomaly Detection},
  author={Gao, Bin-Bin and Zhou, Yue and Yan, Jiangtao and Cai, Yuezhi and Zhang, Weixi and Wang, Meng and Liu, Jun and Liu, Yong and Wang, Lei and Wang, Chengjie},
  journal={arXiv preprint arXiv:2505.09926},
  year={2025}
}

@article{anomalyclip,
  title={Anomalyclip: Object-agnostic prompt learning for zero-shot anomaly detection},
  author={Zhou, Qihang and Pang, Guansong and Tian, Yu and He, Shibo and Chen, Jiming},
  journal={arXiv preprint arXiv:2310.18961},
  year={2023}
}

@inproceedings{Bayesian_Prompt_Flow,
  title={Bayesian Prompt Flow Learning for Zero-Shot Anomaly Detection},
  author={Qu, Zhen and Tao, Xian and Gong, Xinyi and Qu, Shichen and Chen, Qiyu and Zhang, Zhengtao and Wang, Xingang and Ding, Guiguang},
  booktitle={Proceedings of the Computer Vision and Pattern Recognition Conference},
  pages={30398--30408},
  year={2025}
}

@inproceedings{filo,
  title={Filo: Zero-shot anomaly detection by fine-grained description and high-quality localization},
  author={Gu, Zhaopeng and Zhu, Bingke and Zhu, Guibo and Chen, Yingying and Li, Hao and Tang, Ming and Wang, Jinqiao},
  booktitle={Proceedings of the 32nd ACM International Conference on Multimedia},
  pages={2041--2049},
  year={2024}
}

@inproceedings{winclip,
  title={Winclip: Zero-/few-shot anomaly classification and segmentation},
  author={Jeong, Jongheon and Zou, Yang and Kim, Taewan and Zhang, Dongqing and Ravichandran, Avinash and Dabeer, Onkar},
  booktitle={Proceedings of the IEEE/CVF Conference on Computer Vision and Pattern Recognition},
  pages={19606--19616},
  year={2023}
}

@article{PAG,
  title={PAG: Multi-Turn Reinforced LLM Self-Correction with Policy as Generative Verifier},
  author={Jiang, Yuhua and Xiong, Yuwen and Yuan, Yufeng and Xin, Chao and Xu, Wenyuan and Yue, Yu and Zhao, Qianchuan and Yan, Lin},
  journal={arXiv preprint arXiv:2506.10406},
  year={2025}
}

@inproceedings{ReCoT,
  title={ReCoT: Reflective Self-Correction Training for Mitigating Confirmation Bias in Large Vision-Language Models},
  author={Qu, Mengxue and Hu, Yibo and Han, Kunyang and Wei, Yunchao and Zhao, Yao},
  booktitle={Proceedings of the IEEE/CVF International Conference on Computer Vision},
  pages={9147--9157},
  year={2025}
}

@article{Reflect_Retry_Reward,
  title={Reflect, Retry, Reward: Self-Improving LLMs via Reinforcement Learning},
  author={Bensal, Shelly and Jamil, Umar and Bryant, Christopher and Russak, Melisa and Kamble, Kiran and Mozolevskyi, Dmytro and Ali, Muayad and AlShikh, Waseem},
  journal={arXiv preprint arXiv:2505.24726},
  year={2025}
}

@article{srpo,
  title={Srpo: Enhancing multimodal llm reasoning via reflection-aware reinforcement learning},
  author={Wan, Zhongwei and Dou, Zhihao and Liu, Che and Zhang, Yu and Cui, Dongfei and Zhao, Qinjian and Shen, Hui and Xiong, Jing and Xin, Yi and Jiang, Yifan and others},
  journal={arXiv preprint arXiv:2506.01713},
  year={2025}
}

@article{training_language_models_to_self-correct,
  title={Training language models to self-correct via reinforcement learning},
  author={Kumar, Aviral and Zhuang, Vincent and Agarwal, Rishabh and Su, Yi and Co-Reyes, John D and Singh, Avi and Baumli, Kate and Iqbal, Shariq and Bishop, Colton and Roelofs, Rebecca and others},
  journal={arXiv preprint arXiv:2409.12917},
  year={2024}
}

@article{vl-rethinker,
  title={Vl-rethinker: Incentivizing self-reflection of vision-language models with reinforcement learning},
  author={Wang, Haozhe and Qu, Chao and Huang, Zuming and Chu, Wei and Lin, Fangzhen and Chen, Wenhu},
  journal={arXiv preprint arXiv:2504.08837},
  year={2025}
}

@inproceedings{anomalygpt,
  title={Anomalygpt: Detecting industrial anomalies using large vision-language models},
  author={Gu, Zhaopeng and Zhu, Bingke and Zhu, Guibo and Chen, Yingying and Tang, Ming and Wang, Jinqiao},
  booktitle={Proceedings of the AAAI conference on artificial intelligence},
  volume={38},
  pages={1932--1940},
  year={2024}
}

@article{anomalyr1,
  title={Anomalyr1: A grpo-based end-to-end mllm for industrial anomaly detection},
  author={Chao, Yuhao and Liu, Jie and Tang, Jie and Wu, Gangshan},
  journal={arXiv preprint arXiv:2504.11914},
  year={2025}
}

@article{lad-reasoner,
  title={Lad-reasoner: Tiny multimodal models are good reasoners for logical anomaly detection},
  author={Li, Weijia and Chu, Guanglei and Chen, Jiong and Xie, Guo-Sen and Shan, Caifeng and Zhao, Fang},
  journal={arXiv preprint arXiv:2504.12749},
  year={2025}
}

@inproceedings{Logicad,
  title={Logicad: Explainable anomaly detection via vlm-based text feature extraction},
  author={Jin, Er and Feng, Qihui and Mou, Yongli and Lakemeyer, Gerhard and Decker, Stefan and Simons, Oliver and Stegmaier, Johannes},
  booktitle={Proceedings of the AAAI Conference on Artificial Intelligence},
  volume={39},
  pages={4129--4137},
  year={2025}
}

@article{LR-IAD,
  title={LR-IAD: Mask-Free Industrial Anomaly Detection with Logical Reasoning},
  author={Zeng, Peijian and Pang, Feiyan and Wang, Zhanbo and Yang, Aimin},
  journal={arXiv preprint arXiv:2504.19524},
  year={2025}
}

@article{mmad,
  title={Mmad: A comprehensive benchmark for multimodal large language models in industrial anomaly detection},
  author={Jiang, Xi and Li, Jian and Deng, Hanqiu and Liu, Yong and Gao, Bin-Bin and Zhou, Yifeng and Li, Jialin and Wang, Chengjie and Zheng, Feng},
  journal={arXiv preprint arXiv:2410.09453},
  year={2024}
}

@article{myriad,
  title={Myriad: Large multimodal model by applying vision experts for industrial anomaly detection},
  author={Li, Yuanze and Wang, Haolin and Yuan, Shihao and Liu, Ming and Zhao, Debin and Guo, Yiwen and Xu, Chen and Shi, Guangming and Zuo, Wangmeng},
  journal={arXiv preprint arXiv:2310.19070},
  year={2023}
}

@article{omniad,
  title={OmniAD: Detect and Understand Industrial Anomaly via Multimodal Reasoning},
  author={Zhao, Shifang and Lin, Yiheng and Han, Lu and Zhao, Yao and Wei, Yunchao},
  journal={arXiv preprint arXiv:2505.22039},
  year={2025}
}

@inproceedings{towards_zero-shot_anomaly_detection_and_reasoning,
  title={Towards zero-shot anomaly detection and reasoning with multimodal large language models},
  author={Xu, Jiacong and Lo, Shao-Yuan and Safaei, Bardia and Patel, Vishal M and Dwivedi, Isht},
  booktitle={Proceedings of the Computer Vision and Pattern Recognition Conference},
  pages={20370--20382},
  year={2025}
}

@article{vision-r1,
  title={Vision-r1: Incentivizing reasoning capability in multimodal large language models},
  author={Huang, Wenxuan and Jia, Bohan and Zhai, Zijie and Cao, Shaosheng and Ye, Zheyu and Zhao, Fei and Xu, Zhe and Hu, Yao and Lin, Shaohui},
  journal={arXiv preprint arXiv:2503.06749},
  year={2025}
}

@article{visual_Planning,
  title={Visual Planning: Let's Think Only with Images},
  author={Xu, Yi and Li, Chengzu and Zhou, Han and Wan, Xingchen and Zhang, Caiqi and Korhonen, Anna and Vuli{\'c}, Ivan},
  journal={arXiv preprint arXiv:2505.11409},
  year={2025}
}

@article{Vlm-r1,
  title={Vlm-r1: A stable and generalizable r1-style large vision-language model},
  author={Shen, Haozhan and Liu, Peng and Li, Jingcheng and Fang, Chunxin and Ma, Yibo and Liao, Jiajia and Shen, Qiaoli and Zhang, Zilun and Zhao, Kangjia and Zhang, Qianqian and others},
  journal={arXiv preprint arXiv:2504.07615},
  year={2025}
}

@article{li2025iad,
  title={Iad-r1: Reinforcing consistent reasoning in industrial anomaly detection},
  author={Li, Yanhui and Cao, Yunkang and Liu, Chengliang and Xiong, Yuan and Dong, Xinghui and Huang, Chao},
  journal={arXiv preprint arXiv:2508.09178},
  year={2025}
}

@article{li2025survey,
  title={A survey of deep learning for industrial visual anomaly detection},
  author={Li, Zhuo and Yan, Yuhao and Wang, Xiangheng and Ge, Yifei and Meng, Lin},
  journal={Artificial Intelligence Review},
  volume={58},
  number={9},
  pages={279},
  year={2025},
  publisher={Springer}
}

@article{chandola2009anomaly,
  title={Anomaly detection: A survey},
  author={Chandola, Varun and Banerjee, Arindam and Kumar, Vipin},
  journal={ACM computing surveys (CSUR)},
  volume={41},
  number={3},
  pages={1--58},
  year={2009},
  publisher={ACM New York, NY, USA}
}

@inproceedings{vaikundam2016anomaly,
  title={Anomaly region detection and localization in metal surface inspection},
  author={Vaikundam, Sriram and Hung, Tzu-Yi and Chia, Liang Tien},
  booktitle={2016 IEEE International Conference on Image Processing (ICIP)},
  pages={759--763},
  year={2016},
  organization={IEEE}
}

@article{liu2024deep,
  title={Deep industrial image anomaly detection: A survey},
  author={Liu, Jiaqi and Xie, Guoyang and Wang, Jinbao and Li, Shangnian and Wang, Chengjie and Zheng, Feng and Jin, Yaochu},
  journal={Machine Intelligence Research},
  volume={21},
  number={1},
  pages={104--135},
  year={2024},
  publisher={Springer}
}

@article{pang2021deep,
  title={Deep learning for anomaly detection: A review},
  author={Pang, Guansong and Shen, Chunhua and Cao, Longbing and Hengel, Anton Van Den},
  journal={ACM computing surveys (CSUR)},
  volume={54},
  number={2},
  pages={1--38},
  year={2021},
  publisher={ACM New York, NY, USA}
}

@article{taylor2024visionad,
  title={VisionAD, a software package of performant anomaly detection algorithms, and proportion localised, an interpretable metric},
  author={Taylor, Alexander DJ and Tregidgo, Phillip and Morrison, Jonathan James and Campbell, Neill DF},
  journal={Transactions on Machine Learning Research},
  year={2024}
}

@inproceedings{patchcore,
  title={Towards total recall in industrial anomaly detection},
  author={Roth, Karsten and Pemula, Latha and Zepeda, Joaquin and Sch{\"o}lkopf, Bernhard and Brox, Thomas and Gehler, Peter},
  booktitle={Proceedings of the IEEE/CVF conference on computer vision and pattern recognition},
  pages={14318--14328},
  year={2022}
}

@inproceedings{defard2021padim,
  title={Padim: a patch distribution modeling framework for anomaly detection and localization},
  author={Defard, Thomas and Setkov, Aleksandr and Loesch, Angelique and Audigier, Romaric},
  booktitle={International conference on pattern recognition},
  pages={475--489},
  year={2021},
  organization={Springer}
}

@article{team2024qwen2,
  title={Qwen2 technical report},
  author={Team, Qwen and others},
  journal={arXiv preprint arXiv:2407.10671},
  volume={2},
  number={3},
  year={2024}
}

@article{bai2025qwen2,
  title={Qwen2. 5-vl technical report},
  author={Bai, Shuai and Chen, Keqin and Liu, Xuejing and Wang, Jialin and Ge, Wenbin and Song, Sibo and Dang, Kai and Wang, Peng and Wang, Shijie and Tang, Jun and others},
  journal={arXiv preprint arXiv:2502.13923},
  year={2025}
}

@misc{bai2025qwen3vltechnicalreport,
      title={Qwen3-VL Technical Report}, 
      author={Shuai Bai and Yuxuan Cai and Ruizhe Chen and Keqin Chen and Xionghui Chen and Zesen Cheng and Lianghao Deng and Wei Ding and Chang Gao and Chunjiang Ge and Wenbin Ge and Zhifang Guo and Qidong Huang and Jie Huang and Fei Huang and Binyuan Hui and Shutong Jiang and Zhaohai Li and Mingsheng Li and Mei Li and Kaixin Li and Zicheng Lin and Junyang Lin and Xuejing Liu and Jiawei Liu and Chenglong Liu and Yang Liu and Dayiheng Liu and Shixuan Liu and Dunjie Lu and Ruilin Luo and Chenxu Lv and Rui Men and Lingchen Meng and Xuancheng Ren and Xingzhang Ren and Sibo Song and Yuchong Sun and Jun Tang and Jianhong Tu and Jianqiang Wan and Peng Wang and Pengfei Wang and Qiuyue Wang and Yuxuan Wang and Tianbao Xie and Yiheng Xu and Haiyang Xu and Jin Xu and Zhibo Yang and Mingkun Yang and Jianxin Yang and An Yang and Bowen Yu and Fei Zhang and Hang Zhang and Xi Zhang and Bo Zheng and Humen Zhong and Jingren Zhou and Fan Zhou and Jing Zhou and Yuanzhi Zhu and Ke Zhu},
      year={2025},
      eprint={2511.21631},
      archivePrefix={arXiv},
      primaryClass={cs.CV},
      url={https://arxiv.org/abs/2511.21631}, 
}

@misc{anthropic2025claudesonnet45,
  author       = {Anthropic},
  title        = {Introducing Claude Sonnet 4.5},
  howpublished = {\url{https://www.anthropic.com/news/claude-sonnet-4-5}},
  year         = {2025},
  note         = {Accessed: 2026-01-28}
}

@misc{openai2025gpt51,
  author       = {OpenAI},
  title        = {GPT-5.1: A smarter, more conversational ChatGPT},
  howpublished = {\url{https://openai.com/index/gpt-5-1/}},
  year         = {2025},
  note         = {Accessed: 2026-01-28}
}

@misc{openai2025gpt52,
  author       = {OpenAI},
  title        = {Introducing GPT-5.2},
  howpublished = {\url{https://openai.com/index/introducing-gpt-5-2/}},
  year         = {2025},
  note         = {Accessed: 2026-02-10}
}

@misc{google2025gemini25flashlite,
  author       = {Google},
  title        = {Gemini 2.5 Flash-Lite Model | Vertex AI Documentation},
  howpublished = {\url{https://docs.cloud.google.com/vertex-ai/generative-ai/docs/models/gemini/2-5-flash-lite}},
  year         = {2025},
  note         = {Accessed: 2026-01-28}
}

@misc{gemini3,
  author = {{Google DeepMind}},
  title  = {Gemini-3-Pro-Preview},
  year   = {2025},
  howpublished = {\url{https://deepmind.google}},
  note   = {Accessed: 2026-02-10}
}

@misc{wang2025internvl35advancingopensourcemultimodal,
      title={InternVL3.5: Advancing Open-Source Multimodal Models in Versatility, Reasoning, and Efficiency}, 
      author={Weiyun Wang and Zhangwei Gao and Lixin Gu and Hengjun Pu and Long Cui and Xingguang Wei and Zhaoyang Liu and Linglin Jing and Shenglong Ye and Jie Shao and Zhaokai Wang and Zhe Chen and Hongjie Zhang and Ganlin Yang and Haomin Wang and Qi Wei and Jinhui Yin and Wenhao Li and Erfei Cui and Guanzhou Chen and Zichen Ding and Changyao Tian and Zhenyu Wu and Jingjing Xie and Zehao Li and Bowen Yang and Yuchen Duan and Xuehui Wang and Zhi Hou and Haoran Hao and Tianyi Zhang and Songze Li and Xiangyu Zhao and Haodong Duan and Nianchen Deng and Bin Fu and Yinan He and Yi Wang and Conghui He and Botian Shi and Junjun He and Yingtong Xiong and Han Lv and Lijun Wu and Wenqi Shao and Kaipeng Zhang and Huipeng Deng and Biqing Qi and Jiaye Ge and Qipeng Guo and Wenwei Zhang and Songyang Zhang and Maosong Cao and Junyao Lin and Kexian Tang and Jianfei Gao and Haian Huang and Yuzhe Gu and Chengqi Lyu and Huanze Tang and Rui Wang and Haijun Lv and Wanli Ouyang and Limin Wang and Min Dou and Xizhou Zhu and Tong Lu and Dahua Lin and Jifeng Dai and Weijie Su and Bowen Zhou and Kai Chen and Yu Qiao and Wenhai Wang and Gen Luo},
      year={2025},
      eprint={2508.18265},
      archivePrefix={arXiv},
      primaryClass={cs.CV},
      url={https://arxiv.org/abs/2508.18265}, 
}

@misc{li2024llavaonevisioneasyvisualtask,
      title={LLaVA-OneVision: Easy Visual Task Transfer}, 
      author={Bo Li and Yuanhan Zhang and Dong Guo and Renrui Zhang and Feng Li and Hao Zhang and Kaichen Zhang and Peiyuan Zhang and Yanwei Li and Ziwei Liu and Chunyuan Li},
      year={2024},
      eprint={2408.03326},
      archivePrefix={arXiv},
      primaryClass={cs.CV},
      url={https://arxiv.org/abs/2408.03326}, 
}

@inproceedings{manta,
  title={Manta: A large-scale multi-view and visual-text anomaly detection dataset for tiny objects},
  author={Fan, Lei and Fan, Dongdong and Hu, Zhiguang and Ding, Yiwen and Di, Donglin and Yi, Kai and Pagnucco, Maurice and Song, Yang},
  booktitle={Proceedings of the Computer Vision and Pattern Recognition Conference},
  pages={25518--25527},
  year={2025}
}

@inproceedings{MulSen_AD,
  title={Multi-sensor object anomaly detection: Unifying appearance, geometry, and internal properties},
  author={Li, Wenqiao and Zheng, Bozhong and Xu, Xiaohao and Gan, Jinye and Lu, Fading and Li, Xiang and Ni, Na and Tian, Zheng and Huang, Xiaonan and Gao, Shenghua and others},
  booktitle={Proceedings of the Computer Vision and Pattern Recognition Conference},
  pages={9984--9993},
  year={2025}
}

@article{mvtec_locoAD,
  title={Beyond dents and scratches: Logical constraints in unsupervised anomaly detection and localization},
  author={Bergmann, Paul and Batzner, Kilian and Fauser, Michael and Sattlegger, David and Steger, Carsten},
  journal={International Journal of Computer Vision},
  volume={130},
  number={4},
  pages={947--969},
  year={2022},
  publisher={Springer}
}

@misc{AITEX_AFID_dataset,
  title        = {AITEX Fabric Image Database (AFID)},
  author       = {{AITEX — Textile Technological Institute}},
  year         = {2021},
  note         = {Accessed: 2025-01-28},
  howpublished = {\url{https://www.aitex.es/afid/}},
  url          = {https://www.aitex.es/afid/}
}

@inproceedings{
  mishra21-vt-adl,
  author = {Mishra, Pankaj and Verk, Riccardo and Fornasier, Daniele and Piciarelli, Claudio and Foresti, Gian Luca},
  title = {{VT-ADL}: A Vision Transformer Network for Image Anomaly Detection and Localization},
  booktitle = {30th IEEE/IES International Symposium on Industrial Electronics (ISIE)},
  year = {2021},
  month = {June},
  location = {Kyoto, Japan}
}

@inproceedings{mpdd,
  title={Visual prompt tuning},
  author={Jia, Menglin and Tang, Luming and Chen, Bor-Chun and Cardie, Claire and Belongie, Serge and Hariharan, Bharath and Lim, Ser-Nam},
  booktitle={European conference on computer vision},
  pages={709--727},
  year={2022},
  organization={Springer}
}

@article{Bozic2021COMIND,
  author = {Bo{\v{z}}i{\v{c}}, Jakob and Tabernik, Domen and 
  Sko{\v{c}}aj, Danijel},
  journal = {Computers in Industry},
  title = {{Mixed supervision for surface-defect detection:
from weakly to fully supervised learning}},
  year = {2021}
}

@inproceedings{dagm,
  title={Weakly supervised learning for industrial optical inspection},
  author={Wieler, Matthias and Hahn, Tobias},
  booktitle={DAGM symposium in},
  volume={6},
  pages={11},
  year={2007}
}

@inproceedings{dtd,
  title={Zero-shot versus many-shot: Unsupervised texture anomaly detection},
  author={Aota, Toshimichi and Tong, Lloyd Teh Tzer and Okatani, Takayuki},
  booktitle={Proceedings of the IEEE/CVF Winter Conference on Applications of Computer Vision},
  pages={5564--5572},
  year={2023}
}

@article{sdd,
  title={Segmentation-based deep-learning approach for surface-defect detection},
  author={Tabernik, Domen and {\v{S}}ela, Samo and Skvar{\v{c}}, Jure and Sko{\v{c}}aj, Danijel},
  journal={Journal of Intelligent Manufacturing},
  volume={31},
  number={3},
  pages={759--776},
  year={2020},
  publisher={Springer}
}

@inproceedings{mvtec,
  title={MVTec AD--A comprehensive real-world dataset for unsupervised anomaly detection},
  author={Bergmann, Paul and Fauser, Michael and Sattlegger, David and Steger, Carsten},
  booktitle={Proceedings of the IEEE/CVF conference on computer vision and pattern recognition},
  pages={9592--9600},
  year={2019}
}

@inproceedings{visa,
  title={Spot-the-difference self-supervised pre-training for anomaly detection and segmentation},
  author={Zou, Yang and Jeong, Jongheon and Pemula, Latha and Zhang, Dongqing and Dabeer, Onkar},
  booktitle={European conference on computer vision},
  pages={392--408},
  year={2022},
  organization={Springer}
}

@inproceedings{real-iad,
  title={Real-iad: A real-world multi-view dataset for benchmarking versatile industrial anomaly detection},
  author={Wang, Chengjie and Zhu, Wenbing and Gao, Bin-Bin and Gan, Zhenye and Zhang, Jiangning and Gu, Zhihao and Qian, Shuguang and Chen, Mingang and Ma, Lizhuang},
  booktitle={Proceedings of the IEEE/CVF Conference on Computer Vision and Pattern Recognition},
  pages={22883--22892},
  year={2024}
}

@article{shi2016automatic,
  title={Automatic road crack detection using random structured forests},
  author={Shi, Yong and Cui, Limeng and Qi, Zhiquan and Meng, Fan and Chen, Zhensong},
  journal={IEEE Transactions on Intelligent Transportation Systems},
  volume={17},
  number={12},
  pages={3434--3445},
  year={2016},
  publisher={IEEE}
}

@misc{arodi2024cableinspectadexpertannotatedanomalydetection,
      title={CableInspect-AD: An Expert-Annotated Anomaly Detection Dataset}, 
      author={Akshatha Arodi and Margaux Luck and Jean-Luc Bedwani and Aldo Zaimi and Ge Li and Nicolas Pouliot and Julien Beaudry and Gaétan Marceau Caron},
      year={2024},
      eprint={2409.20353},
      archivePrefix={arXiv},
      primaryClass={cs.CV},
      url={https://arxiv.org/abs/2409.20353}, 
}
\bibliographystyle{icml2026}

\newpage
\appendix
\onecolumn
\begin{center}
{\Large \bfseries Appendix}
\end{center}

\setlength{\parindent}{0pt}
\setlength{\parskip}{0.6em}  

A\quad \hyperref[m3-ad-dataset]{Details of the M3-AD Dataset} \dotfill \pageref{m3-ad-dataset}

B\quad \hyperref[detail_reward]{Details of Reward Function Design} \dotfill \pageref{detail_reward}

C\quad \hyperref[detail_experiments]{Details of Experiments} \dotfill \pageref{detail_experiments}

D\quad \hyperref[detail_inference]{Details of Inference Metrics} \dotfill \pageref{detail_inference}

E\quad \hyperref[system_prompt]{System Prompt} \dotfill \pageref{system_prompt}

F\quad \hyperref[supply_case]{Case Study} \dotfill \pageref{supply_case}

\newpage

\section{Details of the M3-AD Dataset}
\label{m3-ad-dataset}
\subsection{Overview of M3-AD Dataset}
As shown in \cref{dataset_compare}, most existing industrial anomaly detection datasets focus on a single aspect of the task, such as binary anomaly classification or spatial localization, and generally lack systematic support for semantic anomaly understanding, reasoning process modeling, and reflective behavior. Mainstream datasets, including MVTec-AD \cite{mvtec}, VisA \cite{visa}, MVTec LOCO-AD \cite{mvtec_locoAD}, CableInspect-AD \cite{arodi2024cableinspectadexpertannotatedanomalydetection}, Real-IAD \cite{real-iad}, and MulSen-AD \cite{MulSen_AD}, although providing pixel-level or region-level annotations for anomaly localization, typically do not support anomaly type classification and do not include any form of reasoning or reflection supervision. As a result, they are insufficient for training and evaluating MLLMs that require semantic understanding, explanation generation, and self-correction capabilities.

MMAD \cite{mmad} further extends anomaly type classification in existing works; however, it relies on an option-based evaluation protocol and does not provide explicit annotations for reasoning processes or reflective behavior, which limits its ability to support reflection-aware learning paradigms. Moreover, during the anomaly analysis and localization stages, MMAD explicitly informs the model in advance that the input image contains an anomaly, and then asks the model to select the anomaly type or localization result from predefined options. This evaluation setting deviates from real-world industrial inspection scenarios, where the presence of anomalies is unknown a priori and must be autonomously determined by the model.

Expert-AD \cite{li2025iad} partially addresses these limitations by introducing anomaly type annotations and basic reasoning information. Nevertheless, it does not explicitly model reflective processes, nor does it provide a unified evaluation benchmark for systematically assessing anomaly localization performance and anomaly analysis capability. In addition, its dataset scale and coverage of industrial scenarios remain relatively limited.

\begin{figure}[htbp]
  \begin{center}
    \centerline{\includegraphics[width=\linewidth]{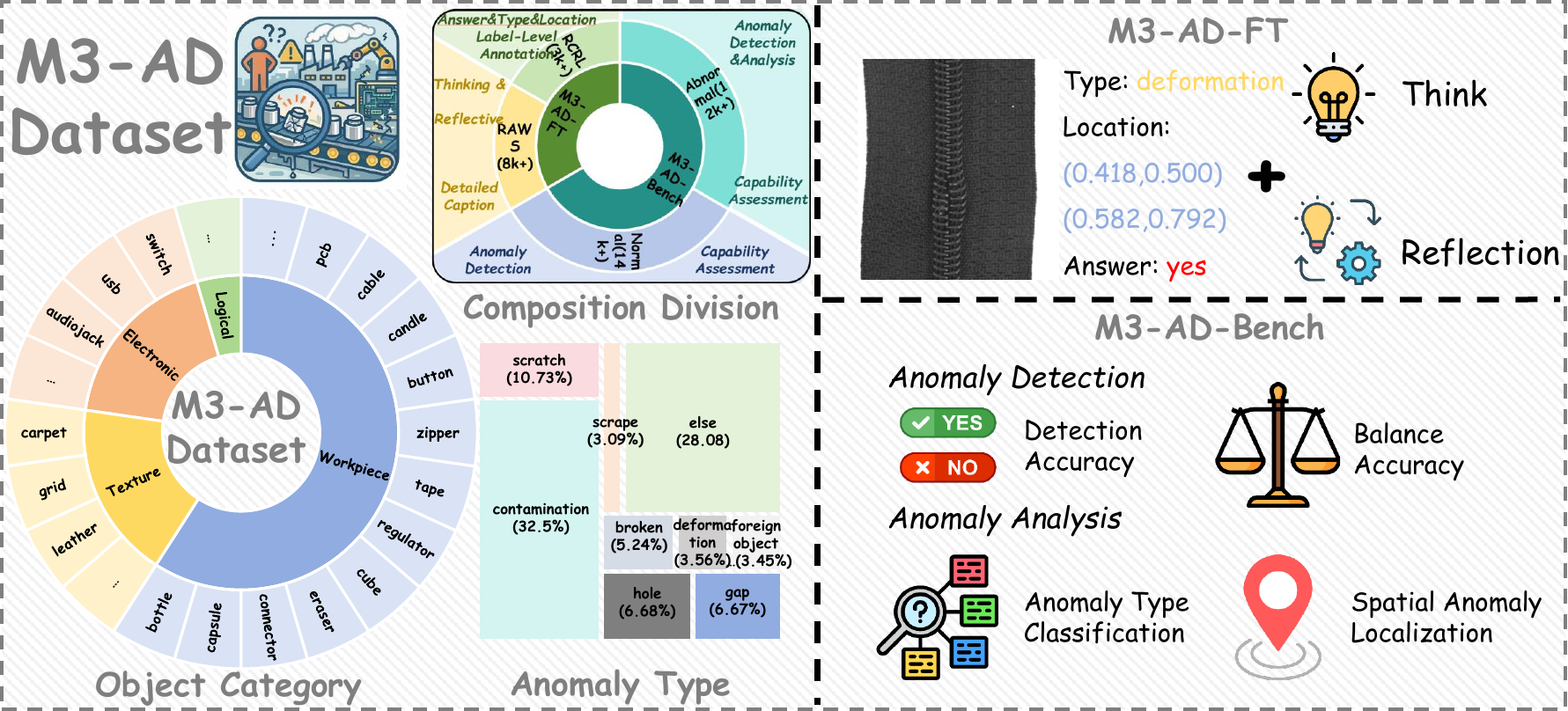}}
    \caption{
      Overview of M3-AD Dataset.
    }
    \label{m3-ad_dataset_overview}
  \end{center}
\end{figure}

In contrast, M3-AD Dataset is currently the only industrial anomaly detection dataset that simultaneously supports reasoning, reflection, anomaly type classification, and precise anomaly localization within a unified framework. \cref{m3-ad_dataset_overview} provides an overview of the structure and composition of the M3-AD dataset. It covers 140 industrial categories across diverse representative industrial scenarios and explicitly annotates both reasoning trajectories and reflective correction processes, enabling models to learn when to trigger reflection and how to perform effective self-correction under uncertain conditions. Furthermore, M3-AD unifies anomaly detection, anomaly analysis, and anomaly localization within a single evaluation protocol, providing a solid foundation for the systematic assessment of reflection-aware multimodal models.

Overall, M3-AD Dataset not only significantly surpasses existing datasets in terms of category coverage and annotation richness, but also explicitly models decision uncertainty and reflection requirements at the data level, offering a comprehensive and unified training and evaluation resource for building reliable, interpretable, and self-corrective industrial anomaly detection systems.
\begin{table}[htpb]
  \centering
  \small
  \caption{Comparison of industrial anomaly detection datasets.}
  \label{tab:dataset_comparison}
  \setlength{\tabcolsep}{4pt}
  \renewcommand{\arraystretch}{1.15}
  \begin{tabular}{l c c c c c c c c c}
    \toprule
    \multirow{2}{*}{\textbf{Dataset}} & \multirow{2}{*}{\textbf{Total}} & \multirow{2}{*}{\textbf{Normal}} & \multirow{2}{*}{\textbf{Abnormal}} &
    \multirow{2}{*}{\textbf{Category}} & \textbf{Anomaly} & \multirow{2}{*}{\textbf{Reasoning}} & \multirow{2}{*}{\textbf{Reflective}} &
    \textbf{Type} & \multirow{2}{*}{\textbf{Localization}} \\
    & & & &  & \textbf{Category} & & & \textbf{Classification} & \\
    \midrule
    MVTec-AD & 5,253 & 4,096 & 1,258 & 15 & 73 & \xmark & \xmark & \xmark & \cmark\\
    VisA & 10821 & 9,621 & 1,200 & 12 & - & \xmark & \xmark & \xmark & \cmark\\
    MVTec LOCO AD & 3,651 & 2,658 & 993 & 5 & 89 & \xmark & \xmark & \xmark & \cmark\\
    CableInspect-AD & 4,798 & 2,159 & 2,639 & 3 & 7 & \xmark & \xmark & \xmark & \cmark\\
    Real-IAD & 151,050 & 99,721 & 51,329 & 30 & 8 & \xmark & \xmark & \xmark & \cmark\\
    Expert-AD & 5,935 & 2,998 & 2,937 & 30 & 63 & \cmark & \xmark & \cmark & \cmark\\
    MulSen-AD & 2035 & 1541 & 494 & 15 & 72 & \xmark & \xmark & \xmark & \cmark\\
    MMAD & 8366 & 5720 & 2646 & 28 & 244 & \xmark & \xmark & \cmark & \cmark\\
    \rowcolor{gray!10}
    M3-AD Dataset & 38,448 & 19,972 & 18,476 & 140 & 67 & \cmark & \cmark & \cmark & \cmark\\
    \bottomrule
    \label{dataset_compare}
  \end{tabular}
\end{table}

\subsection{Construction of M3-AD Dataset}
This section provides detailed descriptions of the M3-AD construction process that is briefly introduced in the main paper. We elaborate on the data sources, the anomaly type unification procedure, the definition of sample difficulty, and the construction of reflection-aware reasoning trajectories, in order to improve reproducibility and clarity.

M3-AD Dataset is constructed by unifying multiple public industrial anomaly detection datasets and covers four representative industrial scenarios, including surface texture anomalies, industrial workpiece anomalies, electronic component and PCB anomalies, as well as assembly and logical anomalies. Based on this unified data pool, we construct two functionally distinct components: M3-AD-FT and M3-AD-Bench. M3-AD-FT is designed for reflection-aligned fine-tuning and contains categories selected to provide diverse anomaly patterns and reasoning trajectories. In contrast, M3-AD-Bench serves as a dedicated evaluation benchmark, where the remaining unseen categories within each high-level scenario are reserved exclusively for cross-category evaluation. This functional separation prevents performance gains from memorizing category-specific appearance patterns and better reflects real-world industrial settings, where models are required to generalize to previously unseen anomaly categories.

\subsection{Unified Anomaly Type Definition}
\begin{figure}
  \begin{center}
    \centerline{\includegraphics[width=\linewidth]{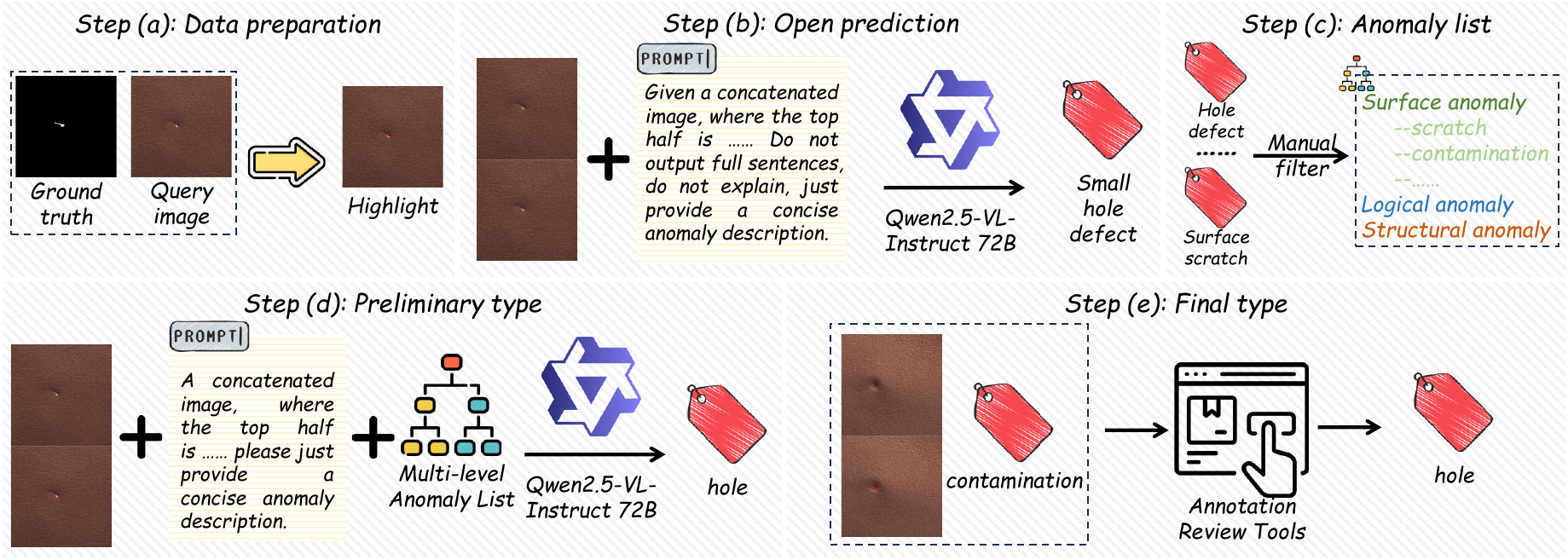}}
    \caption{
      Unified exception type
    }
    \label{anomaly_type}
  \end{center}
\end{figure}

\begin{figure}
  \begin{center}
    \centerline{\includegraphics[width=\linewidth]{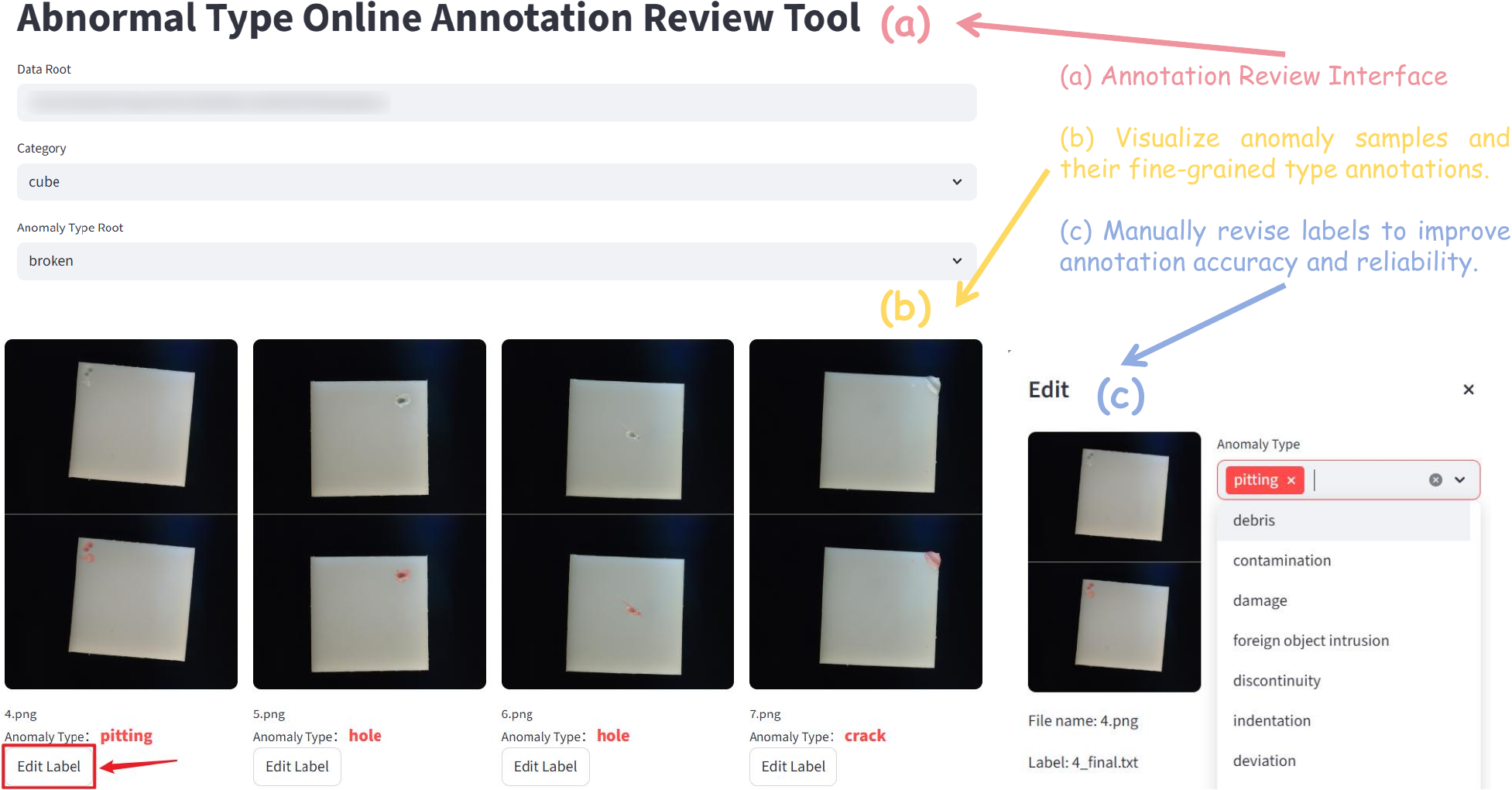}}
    \caption{
      Abnormal Type Online Annotation Review Tool
    }
    \label{online_tool}
  \end{center}
\end{figure}

Due to the significant inconsistency in anomaly type definitions across different source datasets, directly inheriting original labels would lead to semantic ambiguity and hinder reasoning-oriented learning. As shown in \cref{anomaly_type}, M3-AD Dataset redefines anomaly types in a unified manner.
In practice, we first highlight the defect regions in all anomalous images and feed them into a strong multimodal model, Qwen2.5-VL-72B \cite{bai2025qwen2}, to generate open-ended anomaly descriptions. This process is not constrained by any predefined anomaly taxonomy. The model produces concise natural language descriptions characterizing defects in terms of visual appearance, structural disruption, or logical inconsistency.
Next, the generated descriptions are systematically consolidated through a combination of manual review and rule-assisted grouping. Semantically similar descriptions are merged and abstracted to form a hierarchical anomaly taxonomy. Each anomaly type is associated with a clear textual definition, enabling discrimination among different defect patterns in terms of structural damage, appearance variation, or logical violation.
After the taxonomy is established, anomalous images (together with highlighted regions) are paired with the anomaly type candidate set and fed back into the model, guiding it to select the most appropriate type from the predefined taxonomy. Finally, an online annotation and verification tool (as shown in \cref{online_tool}) is used to manually review and refine both anomaly locations and anomaly types, ensuring spatial and semantic consistency. This stage required approximately 120 hours of human annotation and verification effort in total.

\subsection{Sample Difficulty Definition and Reflection-Aware Data Construction}
To support reflection-aware learning, M3-AD-FT proposes a difficulty-aware data construction strategy based on decision correctness. We employ Qwen2.5-VL-72B \cite{bai2025qwen2} as a base model and use its performance on the core decision of anomaly presence (yes/no) as the sole criterion for difficulty assignment.
If the base model produces a correct decision, the corresponding sample is defined as an easy sample. If the base model produces an incorrect decision, the sample is defined as a hard sample, regardless of whether the image is anomalous or normal. This definition grounds sample difficulty in real failure cases encountered during model inference, rather than relying on heuristic rules or subjective human judgment.
Based on this difficulty partition, M3-AD-FT constructs two types of reasoning trajectories in the training data. The first is Thinking Mode, where the model directly produces a final decision after a single reasoning step. The second is Reflective Mode, where the model first generates an initial decision, then performs reflection analysis, and finally outputs a revised decision.
\subsection{Reflection-Aware Reasoning Trajectories}
The assignment of reasoning trajectories in M3-AD-FT is conditioned on sample difficulty rather than uniformly applied. For easy samples, Thinking Mode is the dominant trajectory, while approximately 30\% of the samples are constructed with Reflective Mode. In this case, reflection is not intended to correct errors, but to strengthen the explicit articulation of evidence underlying correct decisions. Through reflection, the model is encouraged to re-examine the visual cues supporting its judgment and convert implicit discriminative signals into more explicit and interpretable semantic representations, thereby improving the stability of representations for normal structures and salient anomaly patterns.

For hard samples, Reflective Mode is the primary reasoning trajectory, accounting for approximately 70\% of the instances. This design explicitly models the learning process of error–reflection–correction, enabling the model to understand the causes of incorrect predictions and learn how to revise unreliable initial decisions. Meanwhile, a small portion of hard samples retain Thinking Mode as a control group, preventing reflection from degenerating into the only decision-making pathway.
A.5 Design of Thinking and Reflection Templates

In the reflection-aware data construction of M3-AD-FT, both thinking and reflection processes are constrained by structured templates to ensure semantic consistency and controllability of reasoning trajectories. The thinking stage focuses on perception and initial judgment, including identifying the object type and industrial scenario in the image, analyzing defect regions in anomalous samples, and verifying structural integrity, surface continuity, and layout rationality in normal samples. The analysis is summarized into a concise natural language decision, serving as the initial prediction.
The reflection stage emphasizes self-examination and decision revision rather than re-describing image content. During reflection, the model is guided to evaluate the correctness of its initial decision with respect to anomaly presence, analyze the reasons for correctness or error by referencing anomaly type definitions and visual evidence, and determine whether the initial decision should be retained or revised.

Specifically, the textual reasoning captions for both Thinking Mode and Reflective Mode are generated using Claude-Sonnet-4.5 \cite{anthropic2025claudesonnet45}, and are subsequently organized into structured reasoning trajectories according to the assigned difficulty levels.

\section{Details of Reward Function Design}
\label{detail_reward}
In the reflection-aware reinforcement learning stage, we design a reward function to guide the model toward reliable decision-making, interpretable explanations, and effective self-reflection in industrial anomaly detection.
The overall reward consists of three components: a consistency reward, an accuracy reward, and a reflection reward.
Among them, the consistency reward serves as a structural constraint, while the accuracy reward and reflection reward provide the primary optimization signals during training.
This section focuses on the detailed design of the latter two components.
\subsection{Consistency Reward}
The consistency reward is proposed to enforce basic structural validity and logical coherence of model outputs.
It ensures that the generated responses contain the required reasoning and decision fields, and that anomaly-specific fields are provided only when appropriate.
This reward is implemented in a binary form and is used solely as a structural constraint; it does not directly evaluate the semantic correctness of anomaly understanding or localization.
Its main role is to provide stable and parsable outputs for subsequent reward computation.
\subsection{Accuracy Reward}
The accuracy reward evaluates the correctness of model predictions in industrial anomaly detection and constitutes the core training signal in the reinforcement learning stage.
It measures prediction quality from three complementary aspects: anomaly existence (\textit{Answer}), anomaly type (\textit{Type}), and anomaly localization (\textit{Location}).

\textbf{Answer Reward}
A gated design is adopted to reflect the decision hierarchy in industrial inspection, where type identification and localization are meaningful only when anomaly existence is correctly predicted.

Let $y$ denote the model-generated output and $g$ denote the ground-truth annotation.
Let $a(\cdot)\in\{\text{yes},\text{no}\}$ represent the parsed anomaly existence decision from the output text.
The answer reward is defined as:
\begin{small}
\begin{equation}
R_{\text{ans}}(y,g) = \mathbb{I}\big[a(y)=a(g)\big],
\end{equation}
\end{small}
where $\mathbb{I}[\cdot]$ is the indicator function.
When the anomaly existence prediction is incorrect, both type and localization rewards are set to zero, preventing the model from receiving positive feedback for fine-grained descriptions under an incorrect base decision.
\textbf{Type Reward}
When the anomaly existence is correctly predicted and the sample is truly anomalous, the type reward and localization reward are further computed.
The type reward is defined as:
\begin{small}
\begin{equation}
R_{\text{type}}(y,g)=
\mathbb{I}\big[a(y)=a(g)=\text{yes}\big]\cdot S_{\text{type}}\big(t(y),t(g)\big),
\end{equation}
\end{small}
where $t(\cdot)$ extracts the set of anomaly types from the \texttt{<type>} field, and
$S_{\text{type}}(\cdot,\cdot)\in[0,1]$ denotes a type matching score.
In practice, $S_{\text{type}}$ is computed based on a predefined hierarchical anomaly taxonomy, allowing partial credit for semantically related or hierarchically consistent type predictions.
\textbf{Location Reward}
Similarly, the localization reward is defined as:
\begin{small}
\begin{equation}
R_{\text{loc}}(y,g)=
\mathbb{I}\big[a(y)=a(g)=\text{yes}\big]\cdot S_{\text{loc}}\big(l(y),l(g)\big),
\end{equation}
\end{small}
where $l(\cdot)$ extracts the set of predicted anomaly bounding boxes from the \texttt{<location>} field, and
$S_{\text{loc}}(\cdot,\cdot)\in[0,1]$ measures spatial consistency between predicted and ground-truth bounding boxes.

The final accuracy reward is computed as:
\begin{small}
\begin{equation}
R_{\text{acc}}(y,g)
=
R_{\text{ans}}(y,g)
+
\frac{1}{2}\Big(R_{\text{type}}(y,g)+R_{\text{loc}}(y,g)\Big).
\end{equation}
\end{small}

This formulation explicitly differentiates between merely correct anomaly detection and correct detection accompanied by accurate semantic and spatial explanations.

\subsection{Reflection Reward}
\label{sec:reflection_reawrd}
The reflection reward is designed to explicitly model the model’s self-reflection and error-correction capability.
Rather than encouraging the unconditional generation of reflection text, this reward evaluates whether reflection leads to a substantive improvement in the model’s decision outcome.

When a reflection field \texttt{<reflection>} is present in the model output, we compare the anomaly predictions before and after reflection.
Let $y_0\in\{0,1\}$ denote the pre-reflection prediction inferred from the initial reasoning process, and
$y_1\in\{0,1\}$ denote the post-reflection prediction derived from the final output, where $1$ indicates an anomalous sample, and $0$ indicates a normal sample.
Let $y\in\{0,1\}$ denote the ground-truth anomaly label.

The reflection reward is defined according to the change in prediction correctness induced by reflection:
\begin{small}
\begin{equation}
R_{\text{ref}}(y_0,y_1,y)=
\begin{cases}
+1.0, & y_0 \neq y \;\land\; y_1 = y, \\
-1.0, & y_0 = y \;\land\; y_1 \neq y, \\
-0.5, & \text{otherwise}.
\end{cases}
\end{equation}
\end{small}

A positive reward is assigned only when reflection successfully corrects an initially incorrect prediction.
If reflection degrades a previously correct prediction, a larger penalty is applied.
For reflections that fail to improve prediction quality—including both unnecessary reflections that preserve correctness and ineffective reflections that remain incorrect—a uniform reflection cost is imposed to discourage meaningless or excessive reflection.
When no reflection field is present in the model output, the reflection reward is set to zero.

\section{Details of Experiments}
\label{detail_experiments}
\subsection{Compared Models}
To comprehensively evaluate the performance of the proposed method in industrial anomaly detection and anomaly analysis tasks, we compare against a diverse set of representative multimodal large language models (MLLMs), covering lightweight commercial models, mainstream open-source general-purpose models, as well as reasoning-enhanced variants. For commercial models, we include GPT-5.1-Nano \cite{openai2025gpt51}, GPT-5.1-Mini \cite{openai2025gpt51}, and Gemini-2.5-Flash-Lite \cite{google2025gemini25flashlite}, which are representative of efficiency-oriented deployments with low inference cost and fast response, reflecting practical requirements in real-world industrial scenarios. In addition, we include Anomaly-R1 and IAD-R1 (IAD-R1(Qwen2.5-VL-Instruct-3B) and IAD-R1(Qwen2.5-VL-Instruct-7B) \cite{anomalyr1,li2025iad} as a representative task-oriented method for anomaly detection, enabling comparison between general-purpose multimodal models and anomaly-specific approaches. For open-source baselines, we select several widely adopted instruction-following MLLMs, including Qwen-2-VL-Instruct \cite{team2024qwen2}, Qwen-2.5-VL-Instruct \cite{bai2025qwen2}, Qwen-3-VL-Instruct \cite{bai2025qwen3vltechnicalreport}, LLaVA-OneVision-SI \cite{li2024llavaonevisioneasyvisualtask}, and Intern-VL-3.5 \cite{wang2025internvl35advancingopensourcemultimodal}. These models exhibit strong cross-modal perception and instruction-following capabilities and serve as representative baselines for contemporary multimodal understanding and reasoning. Furthermore, we include the reasoning-enhanced variant Qwen-3-VL-Thinking \cite{bai2025qwen3vltechnicalreport} to examine whether explicitly strengthened reasoning processes can be directly translated into more reliable anomaly detection and analysis performance. Overall, this set of comparison models spans different model scales, inference paradigms, and application orientations, providing a comprehensive and fair basis for validating the effectiveness and generality of the proposed method.
\subsection{Compare with IAD-R1}
IAD-R1 \cite{li2025iad} is trained on the Real-IAD \cite{real-iad} dataset, while M3-AD-Bench includes image samples originating from Real-IAD.
To avoid potential training--testing overlap and the resulting performance overestimation, we exclude all Real-IAD samples from the test split when evaluating IAD-R1 on M3-AD-Bench. 

\subsection{Detailed Definitions of Reflection Reward Configurations}
\cref{refl_ablation} compares four reflection reward configurations, denoted as (a)--(d), which differ in how rewards or penalties are assigned to three types of reflection outcomes: \emph{correct reflection}, \emph{ineffective reflection}, and \emph{erroneous reflection}. The detailed definitions are as follows:
\begin{itemize}
    \item \textbf{Configuration (a):}
    Correct reflections are rewarded with $+1$, ineffective reflections receive a weak positive reward of $+0.5$, and no penalty is imposed on erroneous reflections. This configuration encourages frequent reflection triggering but lacks explicit constraints on unnecessary or ineffective reflections.

    \item \textbf{Configuration (b):}
    Correct reflections are rewarded with $+1$, ineffective reflections receive a positive reward of $+0.5$, while erroneous reflections are penalized with $-1$. This setting partially suppresses erroneous reflections but still tends to incentivize redundant ineffective reflections.

    \item \textbf{Configuration (c):}
    Correct reflections are rewarded with $+1$, erroneous reflections are penalized with $-1$, and ineffective reflections are neither rewarded nor penalized (0). This configuration reduces incentives for ineffective reflections but does not explicitly introduce a reflection cost.

    \item \textbf{Configuration (d):}
    Correct reflections are rewarded with $+1$, erroneous reflections are penalized with $-1$, and ineffective reflections incur an explicit reflection cost of $-0.5$. By jointly penalizing erroneous reflections and discouraging ineffective ones, this configuration effectively constrains reflection triggering and encourages beneficial self-corrections only when necessary.
\end{itemize}

\begin{figure}[h]
  \begin{center}
    \centerline{\includegraphics[width=\linewidth]{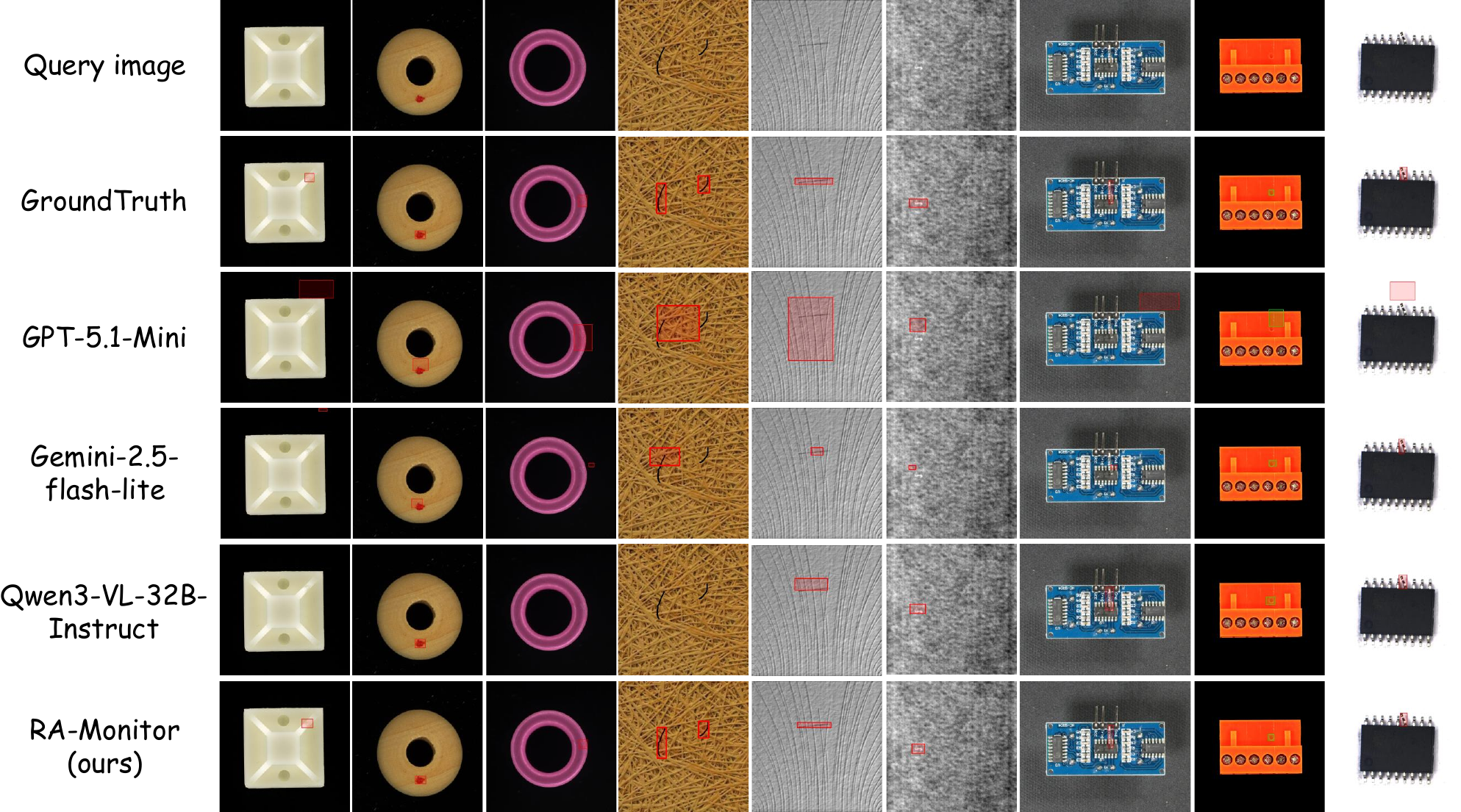}}
    \caption{
    Qualitative comparison of anomaly localization results across different multimodal large language models.
    For each query image, we show the ground-truth bounding boxes and the predicted bounding boxes generated by different models.
    Red boxes represent model predictions, and yellow boxes denote ground-truth annotations.
    Baseline models often exhibit inaccurate or incomplete localization under fine-grained and ambiguous anomaly patterns, whereas RA-Monitor consistently produces more precise and reliable bounding boxes, benefiting from its reflection-aware reasoning mechanism.
    }
    \label{bbox}
  \end{center}

\end{figure}

\subsection{Analysis of Bounding Box Localization Quality}
As shown in \cref{bbox}, RA-Monitor consistently produces more accurate and stable bounding box localization across diverse industrial scenarios. This improvement can be attributed to two key factors. First, the accuracy reward introduced in reflection-aware training not only supervises the correctness of anomaly identification, but also explicitly rewards precise localization of anomalous regions. This encourages the model, once an anomaly is correctly recognized, to further refine the spatial extent of the detected region rather than yielding coarse or incomplete bounding boxes. Second, the proposed multi-level anomaly taxonomy provides a more faithful characterization of different anomaly manifestations by explicitly distinguishing surface defects, structural anomalies, and logical irregularities. Such fine-grained and semantically aligned anomaly definitions offer clearer localization priors, enabling the model to better associate anomaly semantics with specific spatial regions. Together, these design choices allow RA-Monitor to achieve more reliable and precise anomaly localization, even under fine-grained and ambiguous anomaly patterns.
\section{Details of Inference Metrics}
\label{detail_inference}
To comprehensively evaluate the performance of multimodal large models in zero-shot industrial anomaly detection, we design evaluation metrics from two complementary aspects: anomaly detection capability and anomaly analysis capability. Anomaly detection capability measures whether a model can correctly determine the presence of anomalies in an image, while anomaly analysis capability further evaluates the model’s ability to understand and characterize anomalies in terms of their semantic types and spatial locations.

At the anomaly detection level, we report both accuracy and balanced accuracy. Since the proportion of normal and anomalous samples varies across different industrial scenarios, using accuracy alone may lead to biased evaluations due to class imbalance. Balanced accuracy computes the average of recall over normal and anomalous samples, providing a more objective measure of the model’s discriminative ability across both classes. Therefore, we jointly adopt accuracy and balanced accuracy to mitigate the influence of class imbalance on the evaluation results.

For anomaly type analysis, we adopt a Hard Micro-F1 metric with a type matching threshold $\tau$. Anomaly type prediction is inherently a discrete semantic decision problem, where valid outputs are constrained by a predefined anomaly taxonomy. As described earlier, we construct fine-grained and hierarchical anomaly type definitions for different industrial scenarios and explicitly specify the structured semantic space of anomalies through multi-level anomaly type lists. Under this setting, a predicted type is considered a valid match only if it satisfies predefined similarity conditions with the ground-truth type within this taxonomy. The threshold $\tau$ is introduced to distinguish fine-grained semantically consistent predictions from those that remain at a high-level category (e.g., \textit{Texture anomalies}, \textit{Structural anomalies}) or provide vague descriptions, ensuring that Type Hard-F1 faithfully reflects the model’s ability to precisely identify anomaly semantics under zero-shot conditions.

For anomaly localization analysis, we also adopt Hard Micro-F1 and define the matching relationship between predicted and ground-truth bounding boxes based on IoU. In the localization metric, we use IoU $\geq 0.3$ as the hard matching threshold. Similar to the Proportion Localised (PL) metric proposed in VisionAD \cite{taylor2024visionad}, this threshold is designed to exclude unsatisfactory or incidental overlaps between predicted and ground-truth boxes, while allowing reasonable discrepancies between predicted regions and the true anomaly shapes. VisionAD demonstrates through visual analysis that when IoU reaches 0.3 or higher, the spatial overlap between prediction and annotation is sufficient to cover the main anomalous region, thereby visually distinguishing between ``unsuccessful localization'' and ``successful localization.'' Based on this factual evidence, together with our own threshold sensitivity analysis, IoU $= 0.3$ is adopted as the minimum acceptable matching criterion for localization Hard-F1.

\section{System Prompt}
\label{system_prompt}
\subsection{System Prompt for M3-AD-FT Construction}
\vspace{8pt}
\begin{fancycodebox}{System Prompt for M3-AD-FT Construction (Thinking Mode)}
You are an expert vision-language inspector for industrial anomaly detection.

Your task is to look at the image and the correct anomaly information, and then write a concise final diagnosis in English, as if you personally inspected the sample.

You will be given:\\
- an image,(note: in abnormal samples, the image may contain one or more red bounding boxes that highlight where to inspect; the red boxes themselves are not defects, and normal samples never contain such boxes),\\
- the anomaly status of this sample (normal or abnormal),\\
- optional fine-grained defect types with their textual definitions,\\
- and optional defect box coordinates or a short location description (if a red bounding box is present, the actual defect—when present—lies inside the boxed region and not in the red outline itself).

Write 1–2 sentences that follow these rules:

- If the sample is abnormal:\\
  - briefly describe the object or scene,\\
  - clearly state that it is abnormal,\\
  - name the defect type(s),\\
  - and localize the defect using simple spatial terms based on the region indicated by the red box (e.g., “center”, “upper-left corner”, “along the right edge”), while remembering the red frame is not part of the defect.

- If the sample is normal:\\
  - briefly describe the object or scene,\\
  - and state that its structure, surface, and component layout appear intact and consistent, with no visible anomaly in any region, 

Important\\
- Do not mention bounding boxes, highlighted regions, marked areas, or any visual indicators that are not part of the physical object. Even if such information is provided for internal reasoning, your final reflection must describe the defect location only using natural spatial terms (e.g., “upper right portion”, “left edge”, “near the center”).\\
- Do NOT mention “labels”, “ground truth”, “JSON”, or any internal fields.
Just speak as an inspector describing what you see and your final judgment.
\end{fancycodebox}
\vspace*{10pt}
\begin{fancycodebox}{System Prompt for M3-AD-FT Construction (Reflective Mode)}
You are an expert inspector for industrial anomaly detection.

For this task, you will only see samples where the base model's output is not fully aligned with the correct annotation. Your job is to rethink the base model's answer using the image and the correct information.

You will be given:\\
- an image,(Note: the image may contain one or more red bounding boxes overlaid for highlighting; these red boxes are not defects but only indicators of the region to inspect.)\\
- the original answer from a base model (including a yes/no decision inside \texttt{<answer>}...\texttt{</answer>} and a free-form explanation),\\
- and the correct information for this image:\\
  - whether there is an anomaly (yes/no),\\
  - one or more fine-grained defect types, each with a textual definition,\\
  - and one or more defect locations (If bounding boxes are provided, the actual defect—when present—appears inside the highlighted box region and not in the red box lines themselves.), or an explicit note that the image is normal.

Your task is to generate a short reflection paragraph (2–3 sentences, at most 70 words) that:

1. Compares the base answer with the correct information in three aspects:\\
   - whether there is an anomaly (normal vs abnormal),\\
   - the defect type(s), if any,\\
   - and the defect location(s) at a rough, human-understandable level (you do not need to repeat exact coordinates, when referencing the location, treat the red bounding box as a pointer to the region where the defect should appear, while remembering that the red frame itself is not a defect.
).

2. Explains, using the defect definitions and visual evidence in the image, why the base answer is right or wrong on these aspects.\\
   - If the image is normal, focus on why the object looks structurally intact, surface-continuous, and logically consistent, \\
   - If the image is abnormal, focus on where the structure/surface/logic is broken in a way that matches the given defect type(s), especially within the region indicated by the bounding box.

3. States which parts of the base decision should be kept and which parts should be revised
   (for example: change from “no defect” to “abnormal with multiple scratches”, keep “abnormal” but fix the type from contamination to scratch, correct a false alarm to “normal with no anomaly”, etc.).

Important:\\
- Do not mention bounding boxes, highlighted regions, marked areas, or any visual indicators that are not part of the physical object. Even if such information is provided for internal reasoning, your final reflection must describe the defect location only using natural spatial terms (e.g., “upper right portion”, “left edge”, “near the center”).\\
- Do NOT output JSON, tags, or label lists.\\
- Do NOT mention “ground truth”, “labels”, “annotation”, “rethink”, or any JSON fields.\\
- Do NOT copy long definitions verbatim; just use them to support your reasoning.\\
- Just write the reflection as plain English sentences describing why the base answer is correct or incorrect and how it should be updated.
\end{fancycodebox}
\subsection{Inference Time Prompt}
\vspace{8pt}
\begin{fancycodebox}{System Prompt for Inference with Open-Source and Commercial Models}
        If you find anomalies in the test image, structure your response with the following format:\\
        \texttt{<think>}[Your process of observation and reasoning is here]\texttt{</think>}\\
        \texttt{<location>}[x1,y1,x2,y2]\texttt{</location>}\\
        where x1,y1,x2,y2 are NORMALIZED to [0,1] (not pixel coordinates),
        (x1,y1) is top-left, (x2,y2) is bottom-right, and x1<x2, y1<y2.
        \texttt{type}[The label chosen from the predefined anomaly list]\texttt{</type>}\\
        \texttt{<answer>}yes\texttt{</answer>}\\
        If no anomalies are detected in the test image, structure your response with the following format:
        \texttt{<think>}[Your process of observation and reasoning is here]</think>\\
        \texttt{<answer>}no\texttt{</answer>}\\
        Location requirements:\\
        - \texttt{<location>} must be a single normalized bounding box in [0,1] with the format x1,y1,x2,y2.\\
        - (x1,y1) is the TOP-LEFT corner and (x2,y2) is the BOTTOM-RIGHT corner.\\
        - Ensure x1 $<$ x2 and y1 $<$ y2.\\
        Type requirements:\\
        - \texttt{<type>} must be EXACTLY ONE label selected from the following list (use leaf labels only, do not invent new labels):\\
        discontinuity; indentation; deviation; pitting; dent; irregularity; roughness; protrusion;
        corrosion; moisture; oxidation corrosion; rusty; tear; scratch; abrasion; scrape; scuff; bent; warping; distortion; broken; breakage; crack;
        gap; fracture; fragmentation; hole; peeling; delamination; debris; contamination; foreign object intrusion; component misalignment; displacement; component missing; quantity errors;
        wrong combination; layout error; assembly error; size errors; color error.\\
        Output constraints:\\
        - Output ONLY the tags shown above. Do NOT include any extra text outside the tags.\\
\end{fancycodebox}

\section{Case Study}
\label{supply_case}
Using RA-Monitor (4B) as a representative example, we analyze its output behaviors under Thinking Mode and Reflective Mode. For a more intuitive comparison, we evaluate the same test samples using GPT-5.2 \cite{openai2025gpt52} and Gemini-3-Pro-Preview \cite{gemini3}.
These examples illustrate four representative cases: (i) cases where the model correctly identifies the presence of an anomaly directly after the initial reasoning process; (ii) cases where the initial decision fails to recognize the anomaly but the reflective process helps the model refocus on critical visual regions and correct the predicted anomaly type; and (iii) cases where, despite an initially correct decision, reflection further improves evidence articulation and interpretability.
In the model outputs, text segments related to anomaly existence judgment are highlighted in red, those associated with anomaly type analysis are highlighted in yellow, and those corresponding to anomaly localization are highlighted in blue.
In addition, to visually illustrate the localization performance of RA-Monitor (4B), both the ground-truth bounding boxes and the model-predicted bounding boxes are overlaid on the test images. Yellow boxes indicate ground-truth annotations, while red boxes denote the model’s predictions.
\begin{figure*}[htbp]
  \begin{center}
    \centerline{\includegraphics[width=\linewidth]{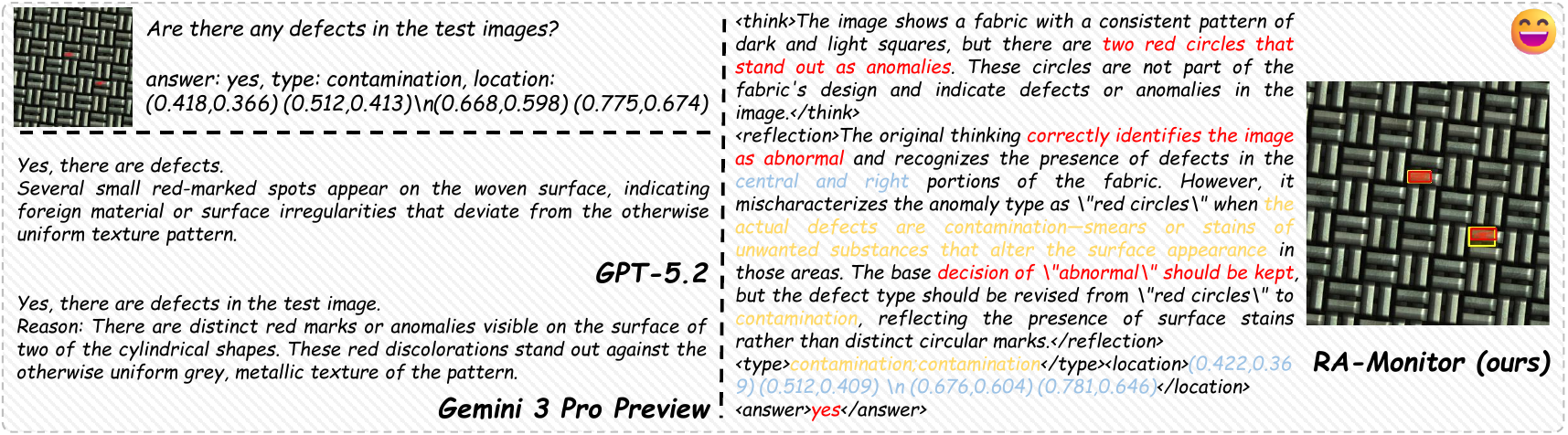}}
    \caption{
    Qualitative example on industrial woven inspection. The left panel shows outputs from GPT-5.2 and Gemini-3 Pro-Preview, while the right panel shows the output of RA-Monitor. Yellow boxes indicate ground-truth annotations, and red boxes denote anomaly regions predicted by RA-Monitor. Yellow boxes indicate ground-truth annotations, while red boxes denote the RA-Monitor’s predictions.
    }
  \end{center}
\end{figure*}

\begin{figure*}[htbp]
  \begin{center}
    \centerline{\includegraphics[width=\linewidth]{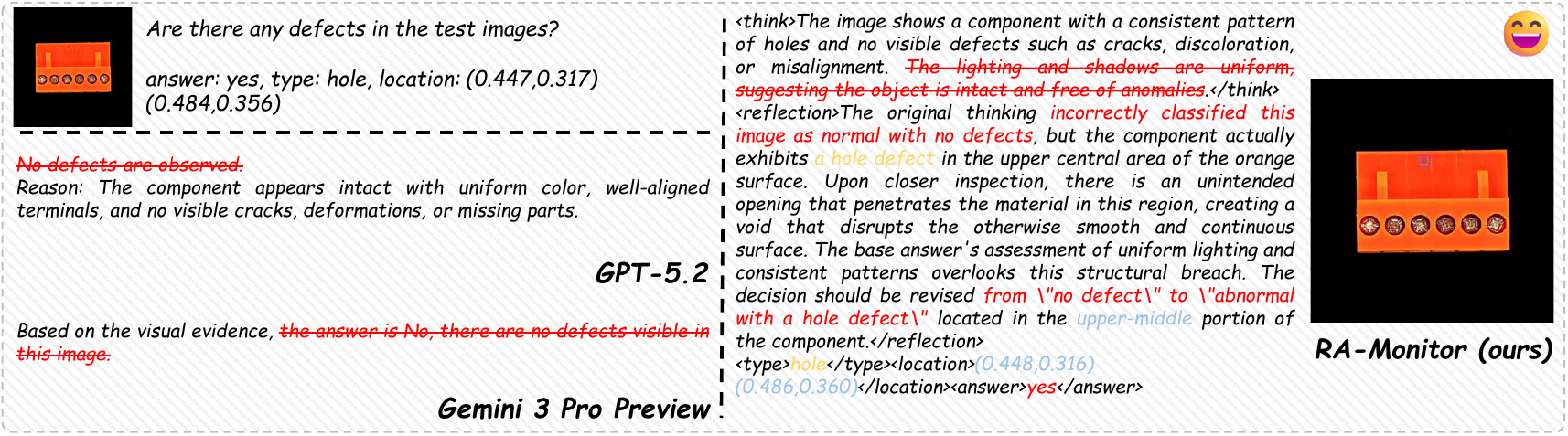}}
        \caption{
    Qualitative example on industrial pcb terminal block inspection. The left panel shows outputs from GPT-5.2 and Gemini-3 Pro-Preview, while the right panel shows the output of RA-Monitor. Yellow boxes indicate ground-truth annotations, and red boxes denote anomaly regions predicted by RA-Monitor. Yellow boxes indicate ground-truth annotations, while red boxes denote the RA-Monitor’s predictions.
    }
  \end{center}
\end{figure*}

\begin{figure*}[htbp]
  \begin{center}
    \centerline{\includegraphics[width=\linewidth]{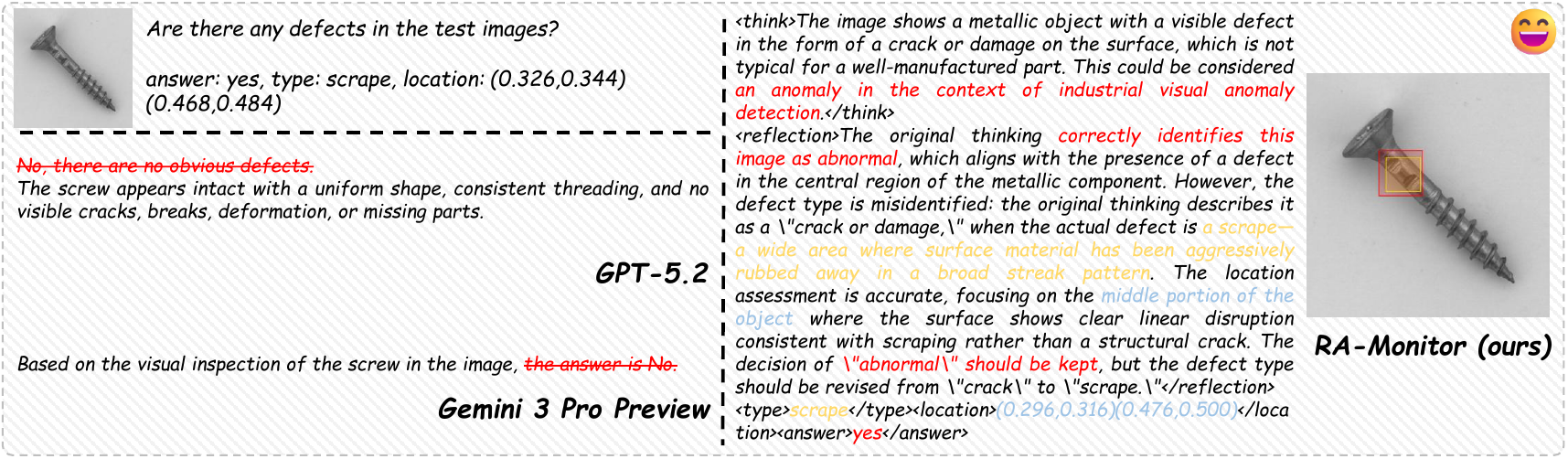}}
        \caption{
        Qualitative example on industrial screw inspection. The left panel shows outputs from GPT-5.2 and Gemini-3 Pro-Preview, while the right panel shows the output of RA-Monitor. Yellow boxes indicate ground-truth annotations, and red boxes denote anomaly regions predicted by RA-Monitor. Yellow boxes indicate ground-truth annotations, while red boxes denote the RA-Monitor’s predictions.
    }
  \end{center}
\end{figure*}

\begin{figure*}[htbp]
  \begin{center}
    \centerline{\includegraphics[width=\linewidth]{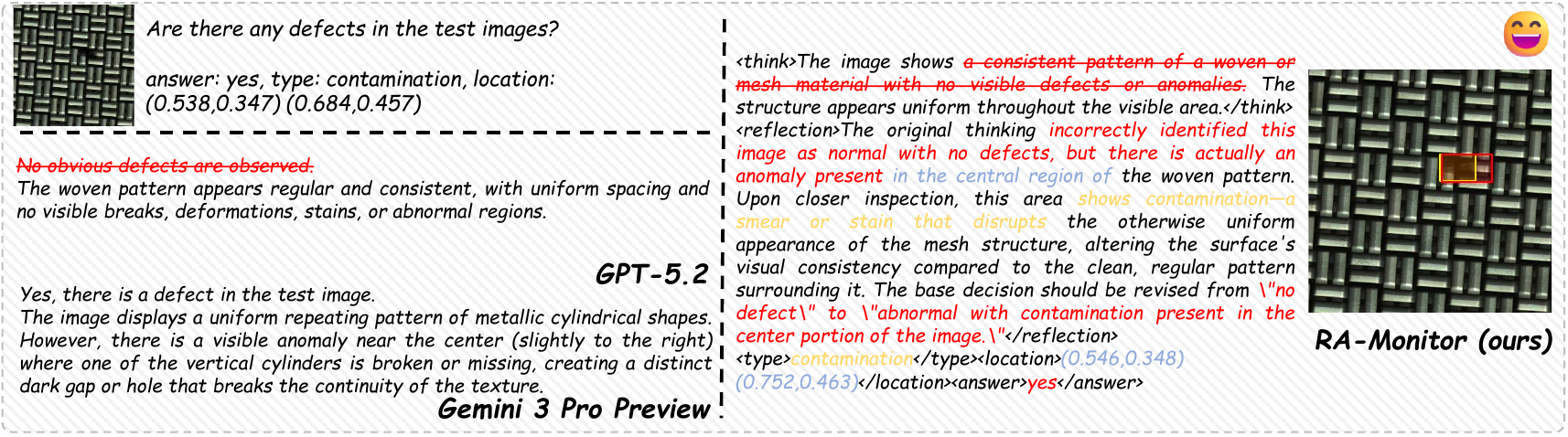}}
        \caption{
    Qualitative example on industrial woven inspection. The left panel shows outputs from GPT-5.2 and Gemini-3 Pro-Preview, while the right panel shows the output of RA-Monitor. Yellow boxes indicate ground-truth annotations, and red boxes denote anomaly regions predicted by RA-Monitor. Yellow boxes indicate ground-truth annotations, while red boxes denote the RA-Monitor’s predictions.
    }
  \end{center}
\end{figure*}

\begin{figure*}[htbp]
  \begin{center}
    \centerline{\includegraphics[width=\linewidth]{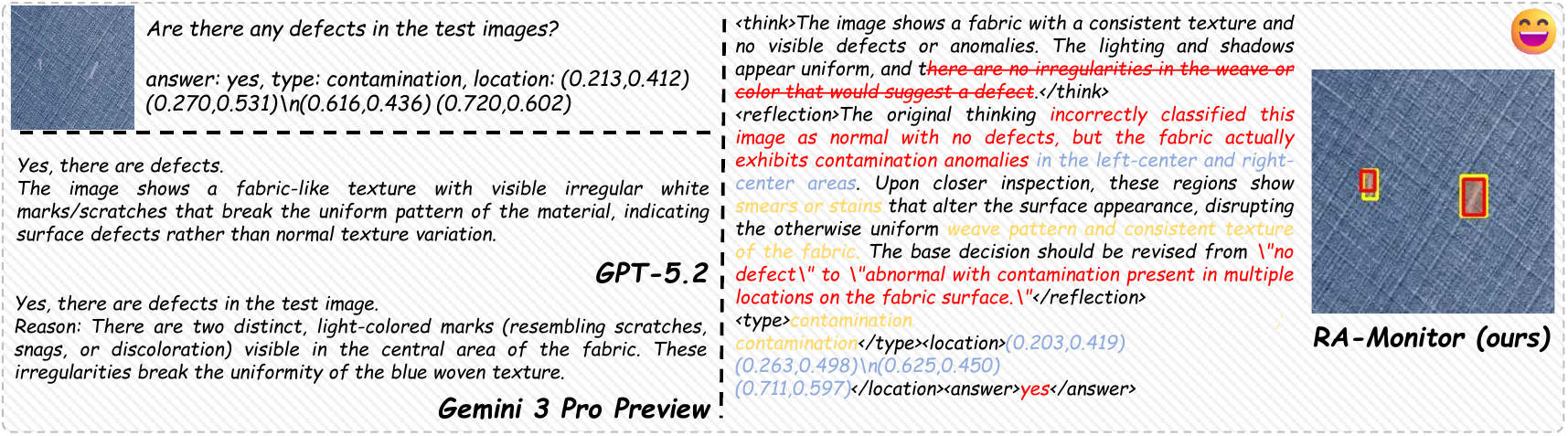}}
        \caption{
    Qualitative example on industrial textile fabrics inspection. The left panel shows outputs from GPT-5.2 and Gemini-3 Pro-Preview, while the right panel shows the output of RA-Monitor. Yellow boxes indicate ground-truth annotations, and red boxes denote anomaly regions predicted by RA-Monitor. Yellow boxes indicate ground-truth annotations, while red boxes denote the RA-Monitor’s predictions.
    }
  \end{center}
\end{figure*}

\begin{figure*}[htbp]
  \begin{center}
    \centerline{\includegraphics[width=\linewidth]{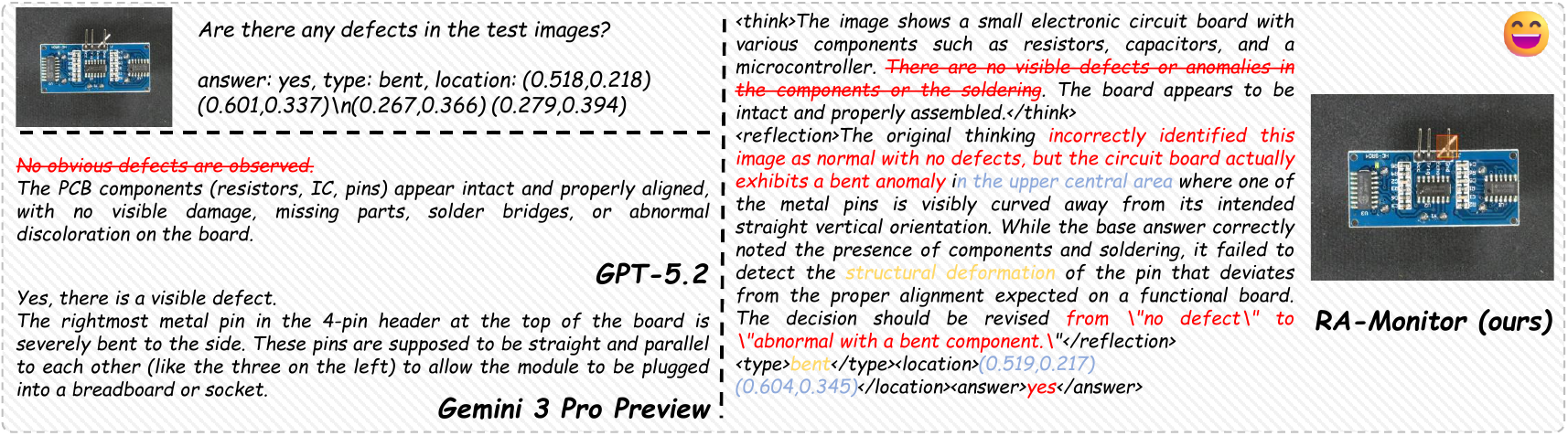}}
        \caption{
    Qualitative example on industrial pcb board inspection. The left panel shows outputs from GPT-5.2 and Gemini-3 Pro-Preview, while the right panel shows the output of RA-Monitor. Yellow boxes indicate ground-truth annotations, and red boxes denote anomaly regions predicted by RA-Monitor. Yellow boxes indicate ground-truth annotations, while red boxes denote the RA-Monitor’s predictions.
    }
  \end{center}
\end{figure*}

\begin{figure*}[htbp]
  \begin{center}
    \centerline{\includegraphics[width=\linewidth]{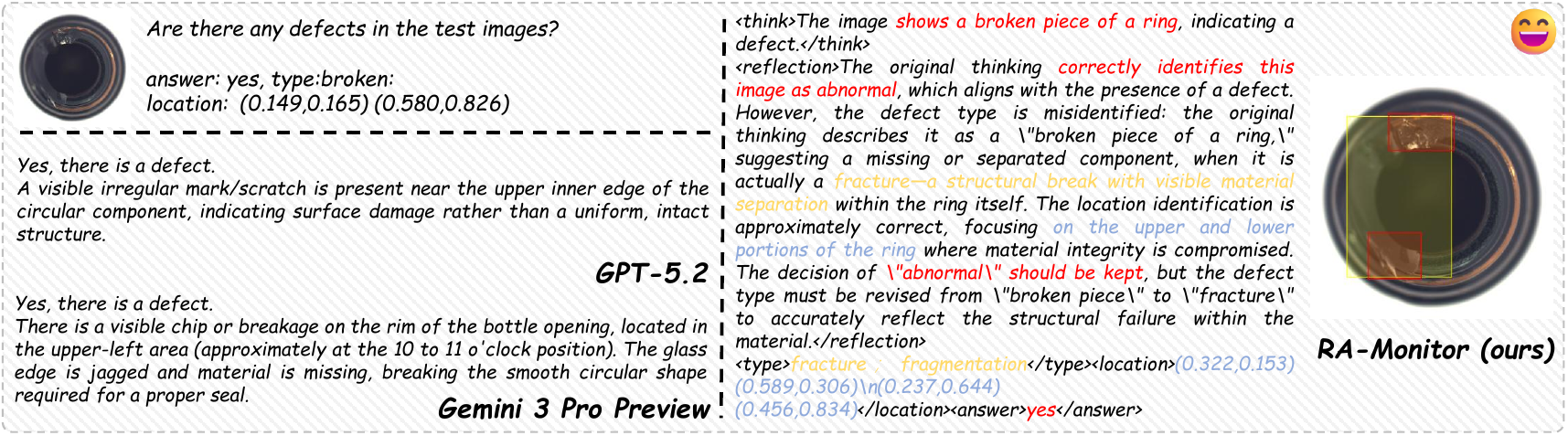}}
        \caption{
    Qualitative example on industrial bottle inspection. The left panel shows outputs from GPT-5.2 and Gemini-3 Pro-Preview, while the right panel shows the output of RA-Monitor. Yellow boxes indicate ground-truth annotations, and red boxes denote anomaly regions predicted by RA-Monitor. Yellow boxes indicate ground-truth annotations, while red boxes denote the RA-Monitor’s predictions.
    }
  \end{center}
\end{figure*}

\begin{figure*}[htbp]
  \begin{center}
    \centerline{\includegraphics[width=\linewidth]{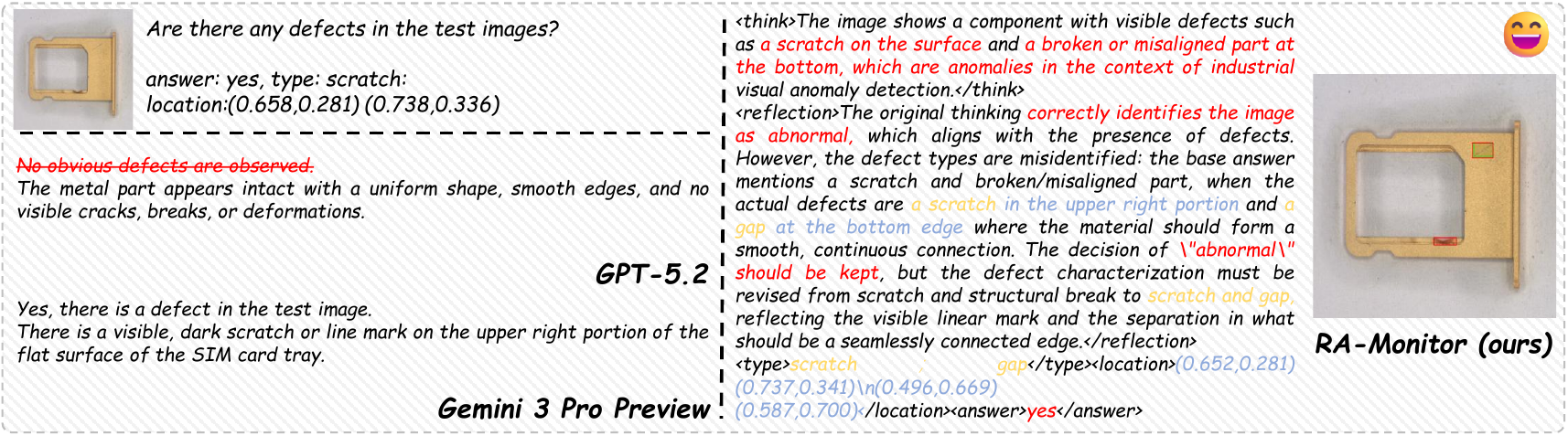}}
        \caption{
    Qualitative example on industrial mobile phone sim card slot inspection. The left panel shows outputs from GPT-5.2 and Gemini-3 Pro-Preview, while the right panel shows the output of RA-Monitor. Yellow boxes indicate ground-truth annotations, and red boxes denote anomaly regions predicted by RA-Monitor. Yellow boxes indicate ground-truth annotations, while red boxes denote the RA-Monitor’s predictions.
    }
  \end{center}
\end{figure*}

\begin{figure*}[htbp]
  \begin{center}
    \centerline{\includegraphics[width=\linewidth]{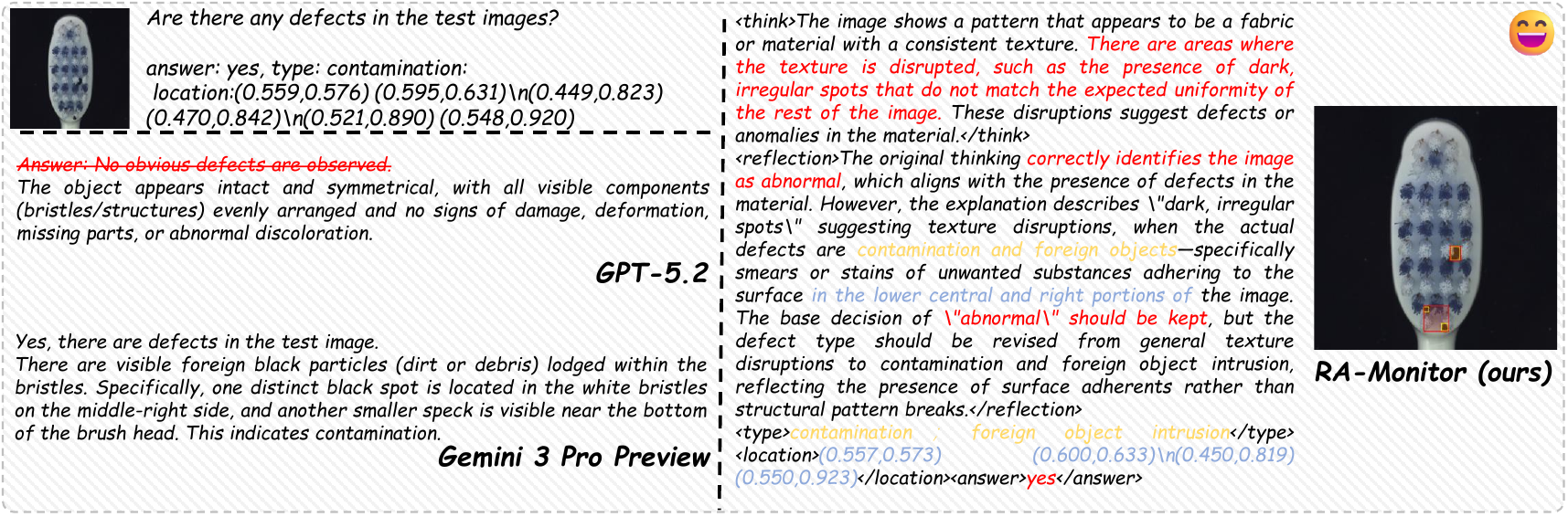}}
        \caption{
    Qualitative example on industrial toothbrush inspection. The left panel shows outputs from GPT-5.2 and Gemini-3 Pro-Preview, while the right panel shows the output of RA-Monitor. Yellow boxes indicate ground-truth annotations, and red boxes denote anomaly regions predicted by RA-Monitor. Yellow boxes indicate ground-truth annotations, while red boxes denote the RA-Monitor’s predictions.
    }
  \end{center}
\end{figure*}

\begin{figure*}[htbp]
  \begin{center}
    \centerline{\includegraphics[width=\linewidth]{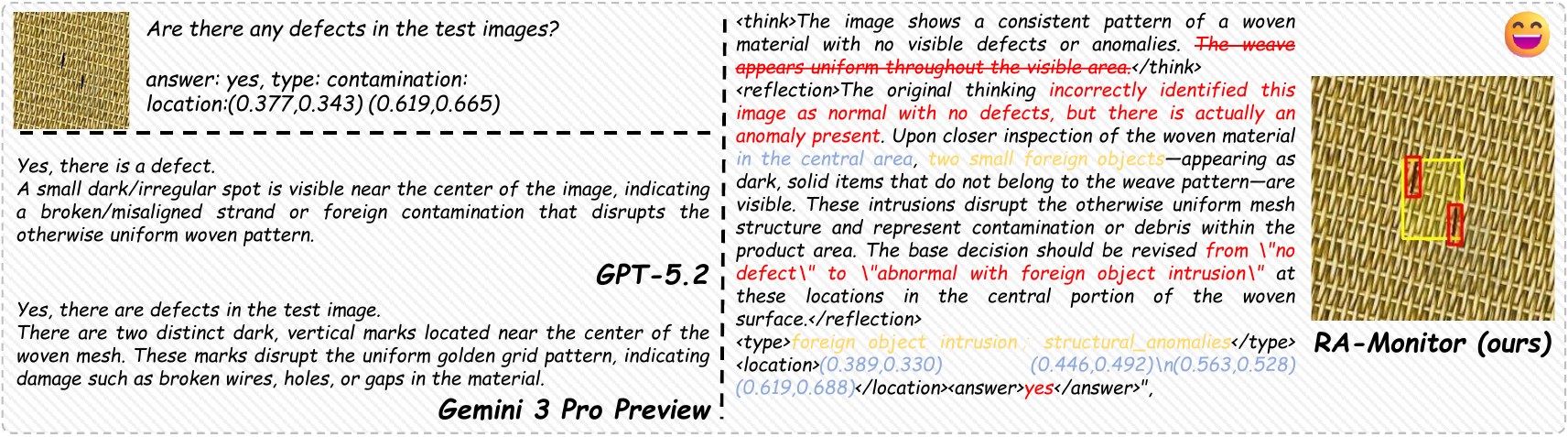}}
        \caption{
    Qualitative example on industrial surface inspection. The left panel shows outputs from GPT-5.2 and Gemini-3 Pro-Preview, while the right panel shows the output of RA-Monitor. Yellow boxes indicate ground-truth annotations, and red boxes denote anomaly regions predicted by RA-Monitor. Yellow boxes indicate ground-truth annotations, while red boxes denote the RA-Monitor’s predictions.
    }
  \end{center}
\end{figure*}

\begin{figure*}[htbp]
  \begin{center}
    \centerline{\includegraphics[width=\linewidth]{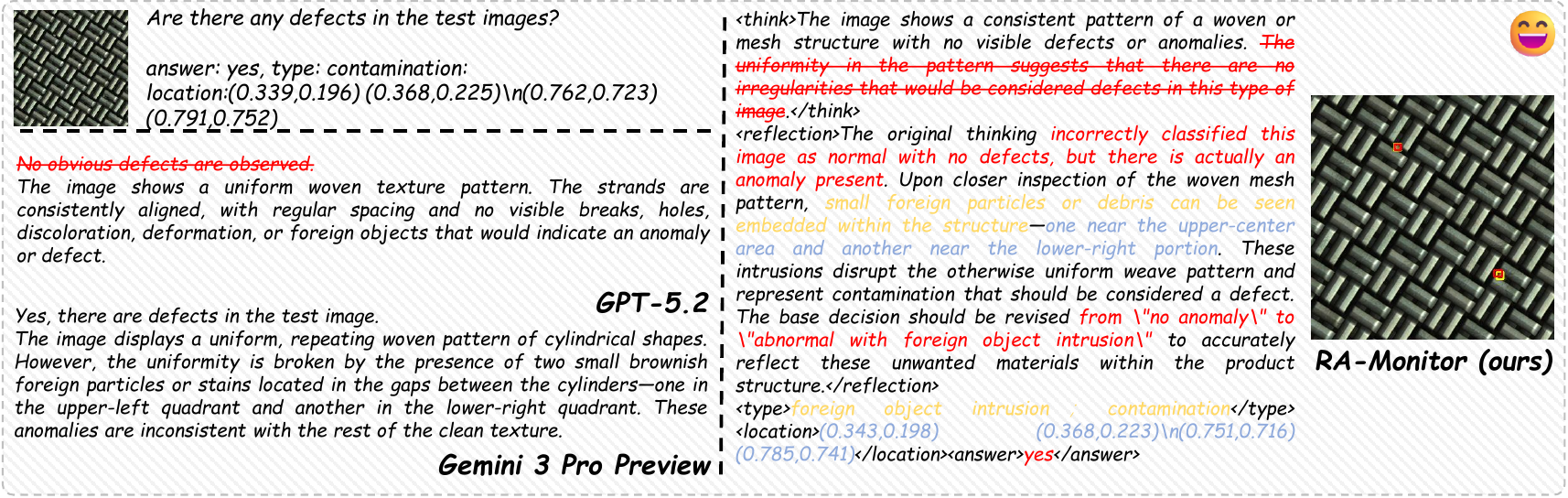}}
        \caption{
    Qualitative example on industrial woven inspection. The left panel shows outputs from GPT-5.2 and Gemini-3 Pro-Preview, while the right panel shows the output of RA-Monitor. Yellow boxes indicate ground-truth annotations, and red boxes denote anomaly regions predicted by RA-Monitor. Yellow boxes indicate ground-truth annotations, while red boxes denote the RA-Monitor’s predictions.
    }
  \end{center}
\end{figure*}

\begin{figure*}[htbp]
  \begin{center}
    \centerline{\includegraphics[width=\linewidth]{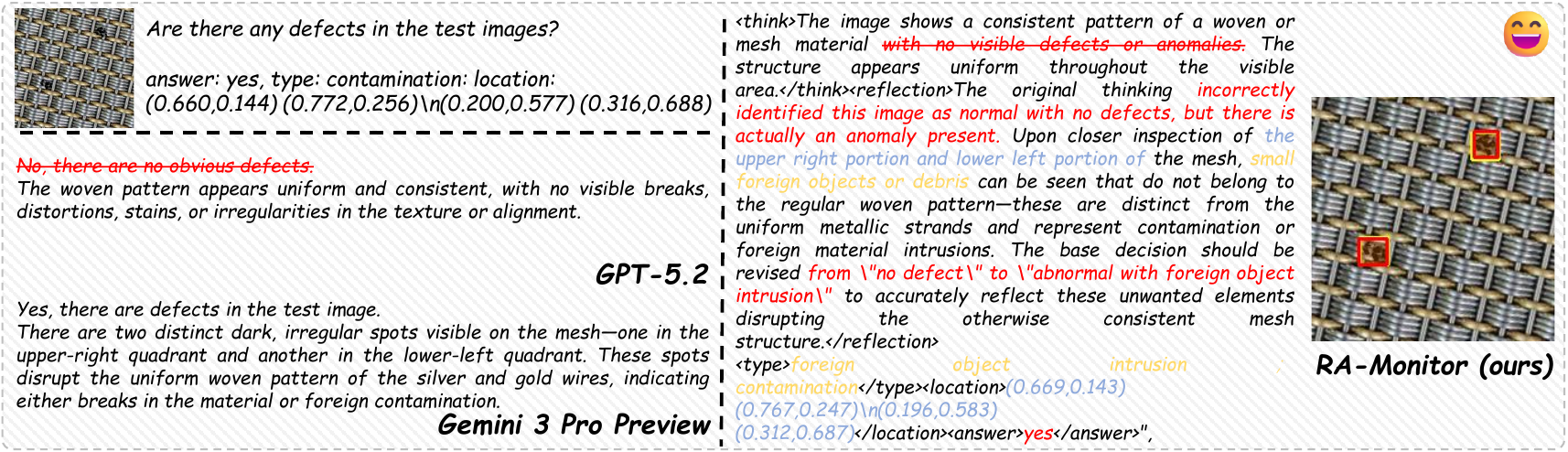}}
        \caption{
    Qualitative example on industrial surface inspection. The left panel shows outputs from GPT-5.2 and Gemini-3 Pro-Preview, while the right panel shows the output of RA-Monitor. Yellow boxes indicate ground-truth annotations, and red boxes denote anomaly regions predicted by RA-Monitor. Yellow boxes indicate ground-truth annotations, while red boxes denote the RA-Monitor’s predictions.
    }
  \end{center}
\end{figure*}

\begin{figure*}[htbp]
  \begin{center}
    \centerline{\includegraphics[width=\linewidth]{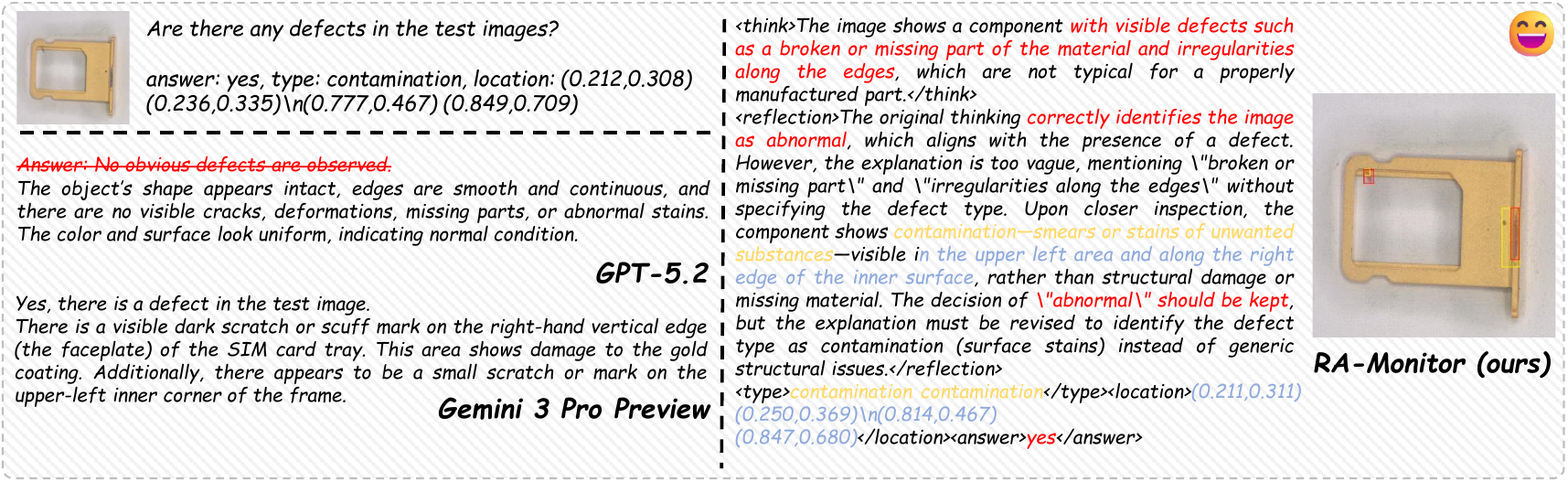}}
        \caption{
    Qualitative example on industrial mobile phone sim card slot inspection. The left panel shows outputs from GPT-5.2 and Gemini-3 Pro-Preview, while the right panel shows the output of RA-Monitor. Yellow boxes indicate ground-truth annotations, and red boxes denote anomaly regions predicted by RA-Monitor. Yellow boxes indicate ground-truth annotations, while red boxes denote the RA-Monitor’s predictions.
    }
  \end{center}
\end{figure*}

\begin{figure*}[htbp]
  \begin{center}
    \centerline{\includegraphics[width=\linewidth]{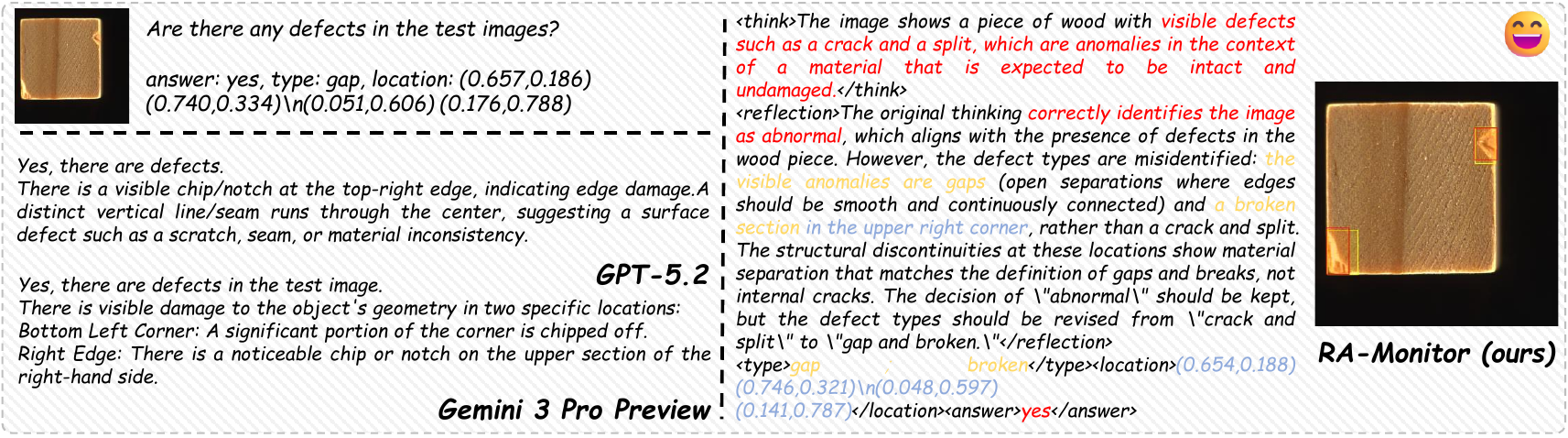}}
        \caption{
    Qualitative example on industrial toy brick inspection. The left panel shows outputs from GPT-5.2 and Gemini-3 Pro-Preview, while the right panel shows the output of RA-Monitor. Yellow boxes indicate ground-truth annotations, and red boxes denote anomaly regions predicted by RA-Monitor. Yellow boxes indicate ground-truth annotations, while red boxes denote the RA-Monitor’s predictions.
    }
  \end{center}
\end{figure*}

\begin{figure*}[htbp]
  \begin{center}
    \centerline{\includegraphics[width=\linewidth]{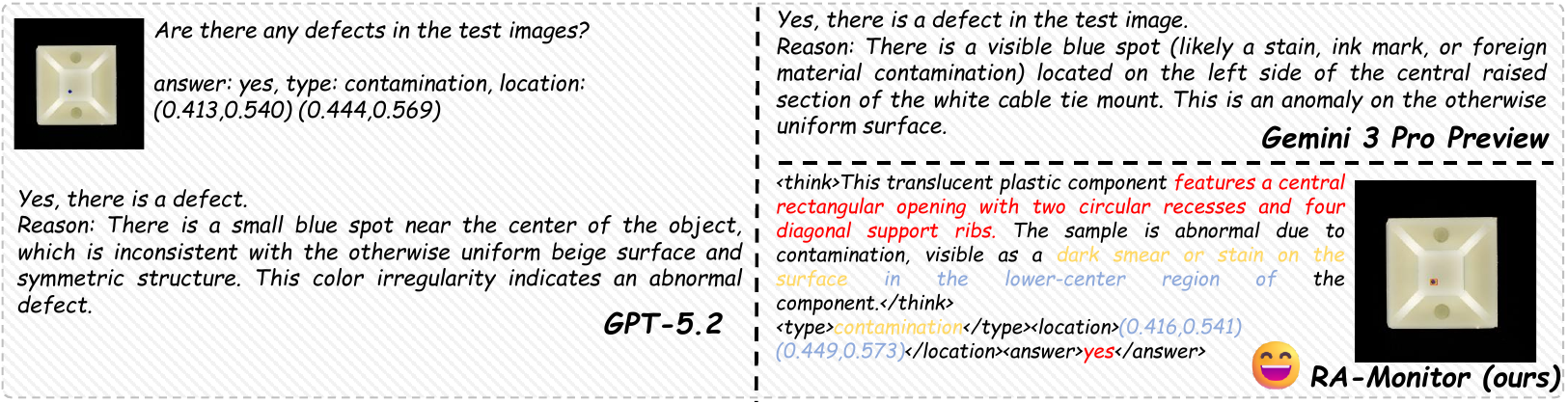}}
        \caption{
    Qualitative example on industrial rolled strip base inspection. The left panel shows outputs from GPT-5.2 and Gemini-3 Pro-Preview, while the right panel shows the output of RA-Monitor. Yellow boxes indicate ground-truth annotations, and red boxes denote anomaly regions predicted by RA-Monitor. Yellow boxes indicate ground-truth annotations, while red boxes denote the RA-Monitor’s predictions.
    }
  \end{center}
\end{figure*}

\begin{figure*}[htbp]
  \begin{center}
    \centerline{\includegraphics[width=\linewidth]{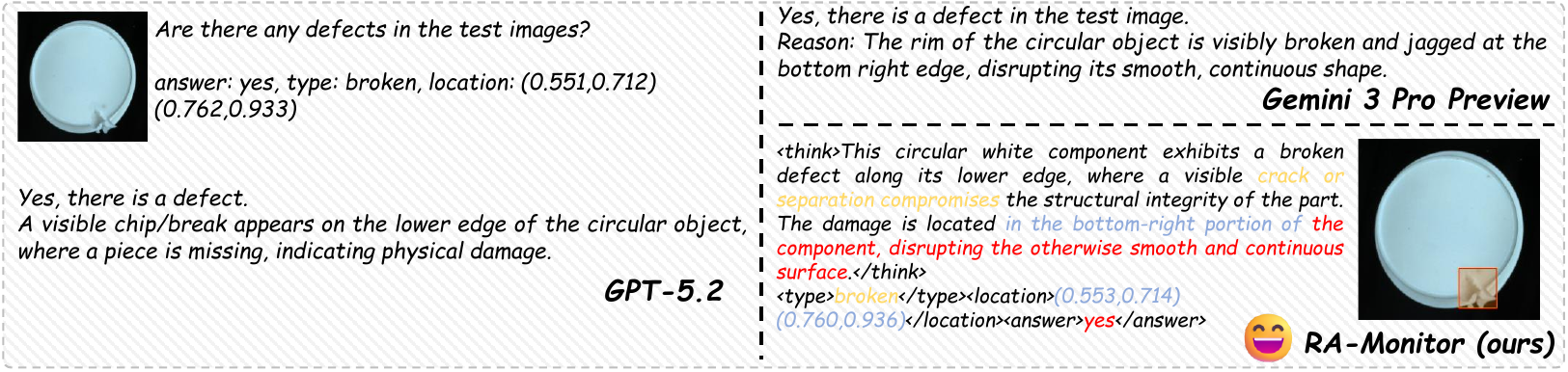}}
        \caption{
    Qualitative example on industrial round button cap inspection. The left panel shows outputs from GPT-5.2 and Gemini-3 Pro-Preview, while the right panel shows the output of RA-Monitor. Yellow boxes indicate ground-truth annotations, and red boxes denote anomaly regions predicted by RA-Monitor. Yellow boxes indicate ground-truth annotations, while red boxes denote the RA-Monitor’s predictions.
    }
  \end{center}
\end{figure*}

\begin{figure*}[htbp]
  \begin{center}
    \centerline{\includegraphics[width=\linewidth]{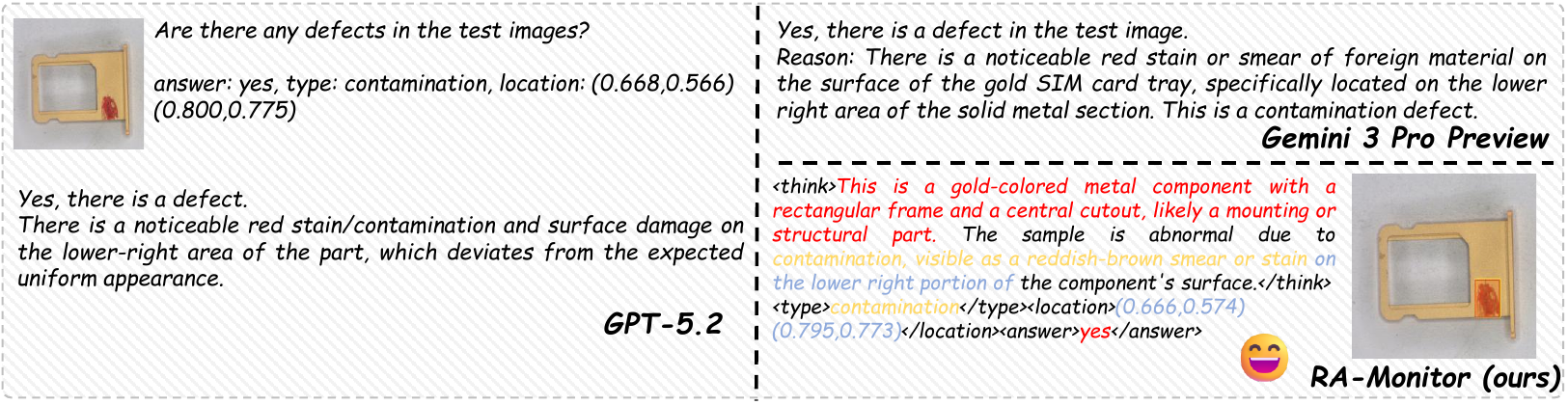}}
        \caption{
    Qualitative example on industrial mobile phone sim card slot inspection. The left panel shows outputs from GPT-5.2 and Gemini-3 Pro-Preview, while the right panel shows the output of RA-Monitor. Yellow boxes indicate ground-truth annotations, and red boxes denote anomaly regions predicted by RA-Monitor. Yellow boxes indicate ground-truth annotations, while red boxes denote the RA-Monitor’s predictions.
    }
  \end{center}
\end{figure*}

\begin{figure*}[htbp]
  \begin{center}
    \centerline{\includegraphics[width=\linewidth]{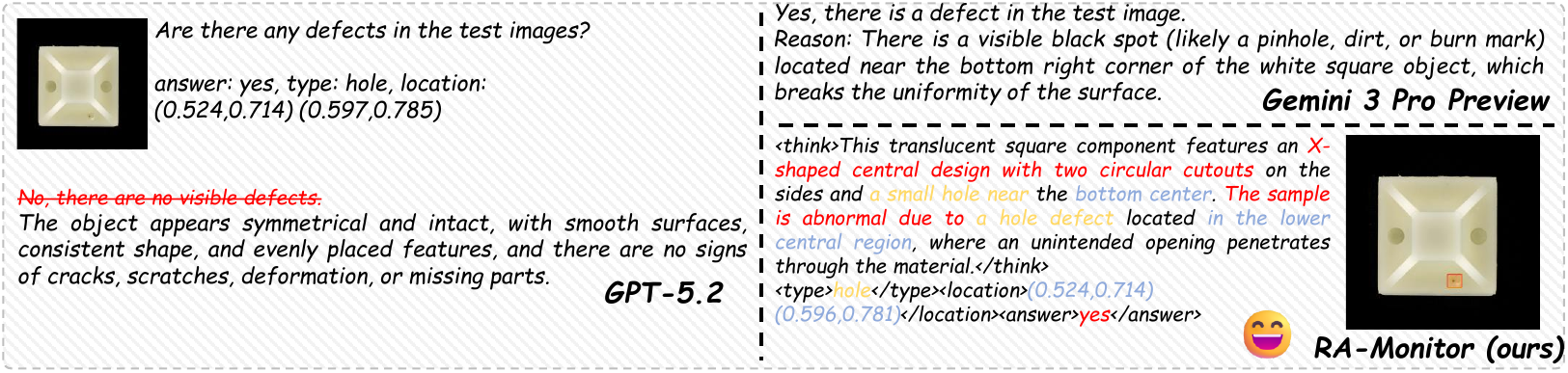}}
        \caption{
    Qualitative example on industrial rolled strip base inspection. The left panel shows outputs from GPT-5.2 and Gemini-3 Pro-Preview, while the right panel shows the output of RA-Monitor. Yellow boxes indicate ground-truth annotations, and red boxes denote anomaly regions predicted by RA-Monitor. Yellow boxes indicate ground-truth annotations, while red boxes denote the RA-Monitor’s predictions.
    }
  \end{center}
\end{figure*}

\begin{figure*}[htbp]
  \begin{center}
    \centerline{\includegraphics[width=\linewidth]{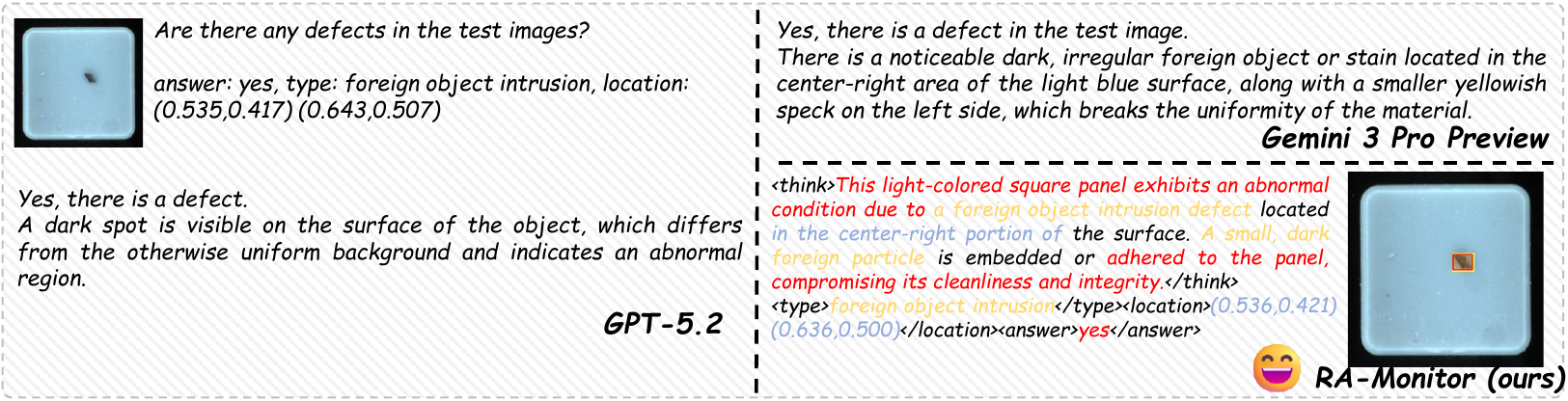}}
        \caption{
    Qualitative example on industrial square button cap inspection. The left panel shows outputs from GPT-5.2 and Gemini-3 Pro-Preview, while the right panel shows the output of RA-Monitor. Yellow boxes indicate ground-truth annotations, and red boxes denote anomaly regions predicted by RA-Monitor. Yellow boxes indicate ground-truth annotations, while red boxes denote the RA-Monitor’s predictions.
    }
  \end{center}
\end{figure*}

\begin{figure*}[htbp]
  \begin{center}
    \centerline{\includegraphics[width=\linewidth]{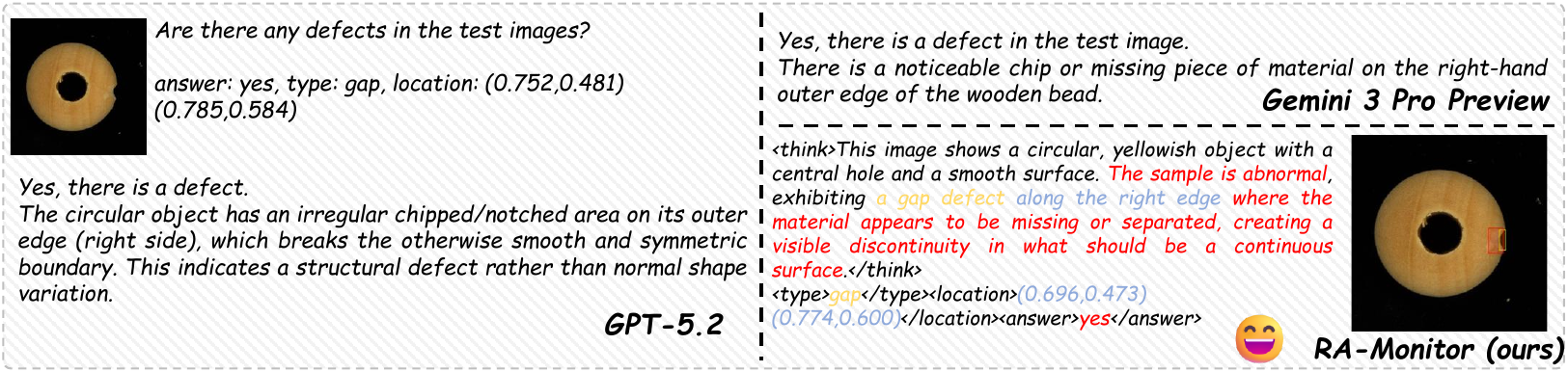}}
        \caption{
    Qualitative example on industrial wooden beads inspection. The left panel shows outputs from GPT-5.2 and Gemini-3 Pro-Preview, while the right panel shows the output of RA-Monitor. Yellow boxes indicate ground-truth annotations, and red boxes denote anomaly regions predicted by RA-Monitor. Yellow boxes indicate ground-truth annotations, while red boxes denote the RA-Monitor’s predictions.
    }
  \end{center}
\end{figure*}

\begin{figure*}[htbp]
  \begin{center}
    \centerline{\includegraphics[width=\linewidth]{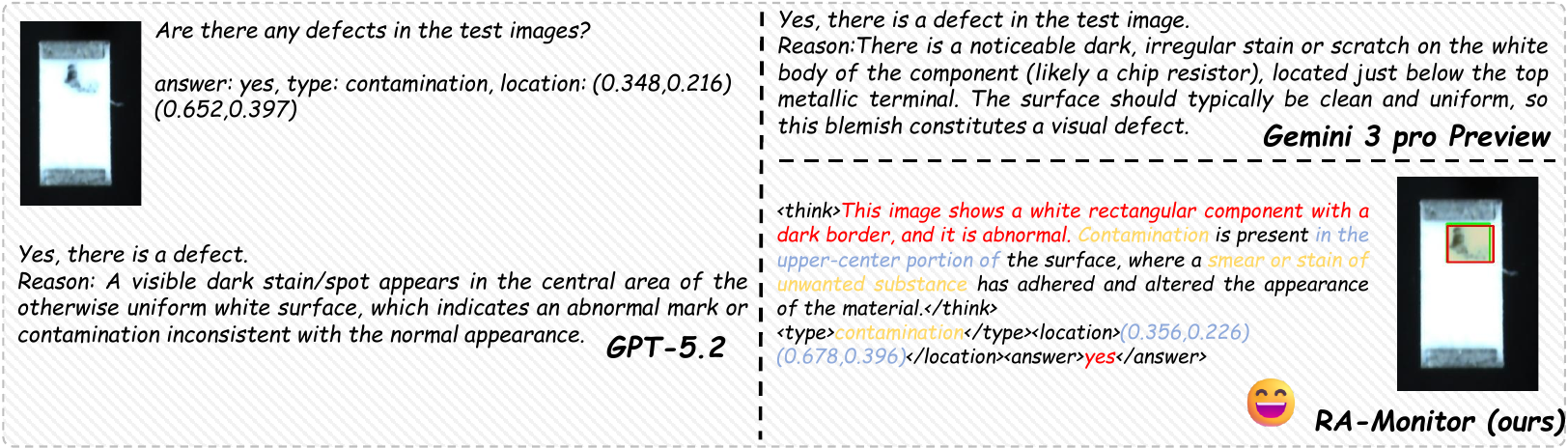}}
        \caption{
    Qualitative example on industrial thin resistor inspection. The left panel shows outputs from GPT-5.2 and Gemini-3 Pro-Preview, while the right panel shows the output of RA-Monitor. Yellow boxes indicate ground-truth annotations, and red boxes denote anomaly regions predicted by RA-Monitor. Yellow boxes indicate ground-truth annotations, while red boxes denote the RA-Monitor’s predictions.
    }
  \end{center}
\end{figure*}

\begin{figure*}[htbp]
  \begin{center}
    \centerline{\includegraphics[width=\linewidth]{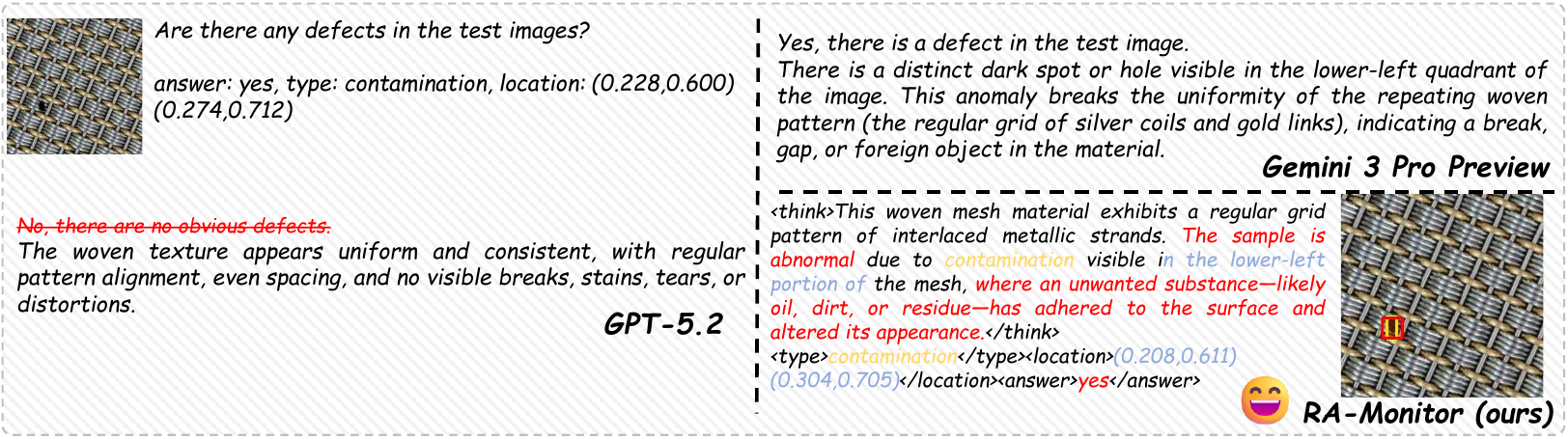}}
        \caption{
    Qualitative example on industrial surface inspection. The left panel shows outputs from GPT-5.2 and Gemini-3 Pro-Preview, while the right panel shows the output of RA-Monitor. Yellow boxes indicate ground-truth annotations, and red boxes denote anomaly regions predicted by RA-Monitor. Yellow boxes indicate ground-truth annotations, while red boxes denote the RA-Monitor’s predictions.
    }
  \end{center}
\end{figure*}

\begin{figure*}[htbp]
  \begin{center}
    \centerline{\includegraphics[width=\linewidth]{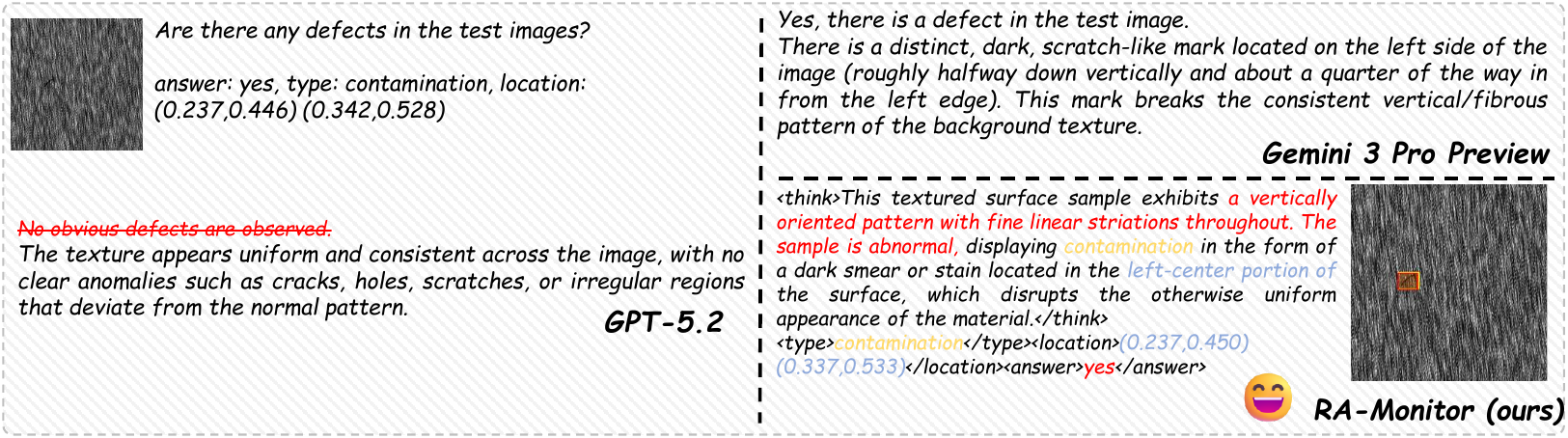}}
        \caption{
    Qualitative example on industrial surface inspection. The left panel shows outputs from GPT-5.2 and Gemini-3 Pro-Preview, while the right panel shows the output of RA-Monitor. Yellow boxes indicate ground-truth annotations, and red boxes denote anomaly regions predicted by RA-Monitor. Yellow boxes indicate ground-truth annotations, while red boxes denote the RA-Monitor’s predictions.
    }
  \end{center}
\end{figure*}

\begin{figure*}[htbp]
  \begin{center}
    \centerline{\includegraphics[width=\linewidth]{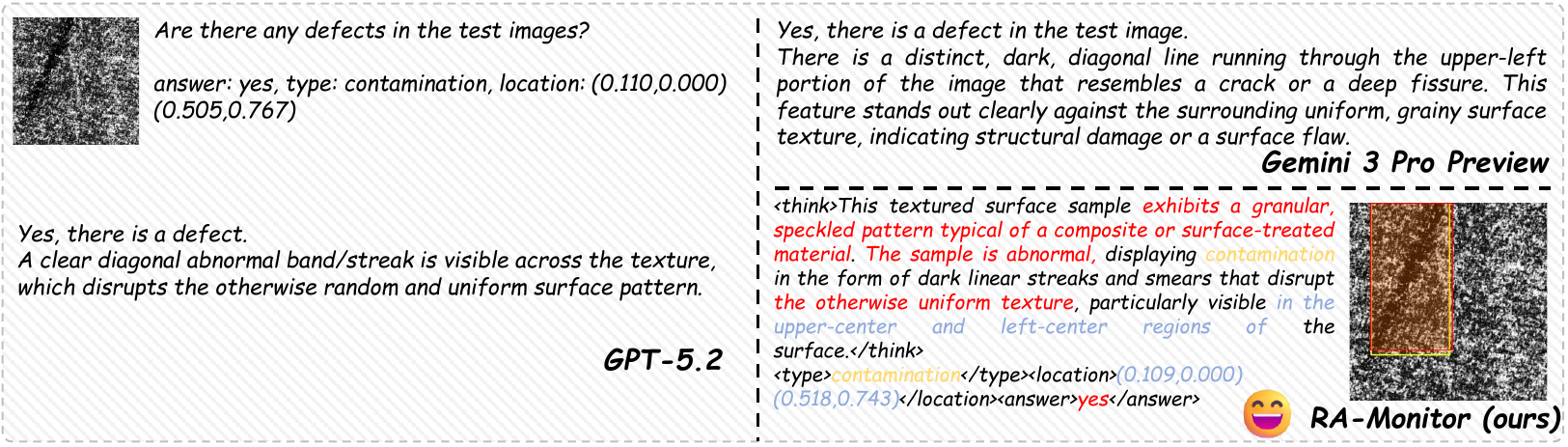}}
        \caption{
    Qualitative example on industrial surface inspection. The left panel shows outputs from GPT-5.2 and Gemini-3 Pro-Preview, while the right panel shows the output of RA-Monitor. Yellow boxes indicate ground-truth annotations, and red boxes denote anomaly regions predicted by RA-Monitor. Yellow boxes indicate ground-truth annotations, while red boxes denote the RA-Monitor’s predictions.
    }
  \end{center}
\end{figure*}

\begin{figure}[htbp]
  \begin{center}
    \centerline{\includegraphics[width=\linewidth]{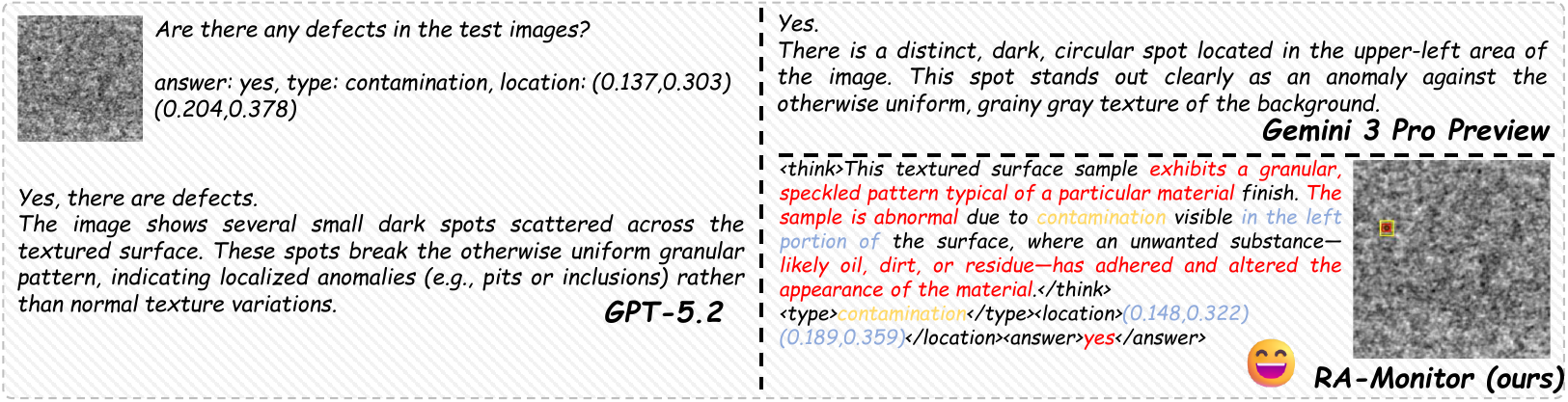}}
        \caption{
    Qualitative example on industrial surface inspection. The left panel shows outputs from GPT-5.2 and Gemini-3 Pro-Preview, while the right panel shows the output of RA-Monitor. Yellow boxes indicate ground-truth annotations, and red boxes denote anomaly regions predicted by RA-Monitor. Yellow boxes indicate ground-truth annotations, while red boxes denote the RA-Monitor’s predictions.
    }
  \end{center}
\end{figure}

\begin{figure}[htbp]
  \begin{center}
    \centerline{\includegraphics[width=\linewidth]{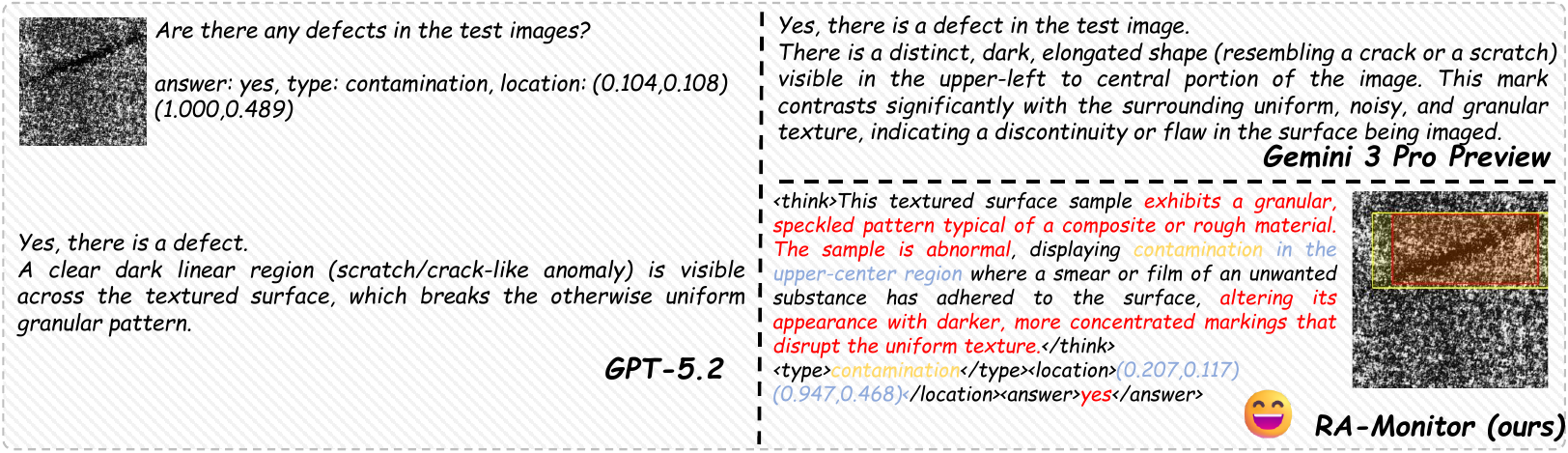}}
        \caption{
    Qualitative example on industrial surface inspection. The left panel shows outputs from GPT-5.2 and Gemini-3 Pro-Preview, while the right panel shows the output of RA-Monitor. Yellow boxes indicate ground-truth annotations, and red boxes denote anomaly regions predicted by RA-Monitor. Yellow boxes indicate ground-truth annotations, while red boxes denote the RA-Monitor’s predictions.
    }
  \end{center}
\end{figure}

\begin{figure}[htbp]
  \begin{center}
    \centerline{\includegraphics[width=\linewidth]{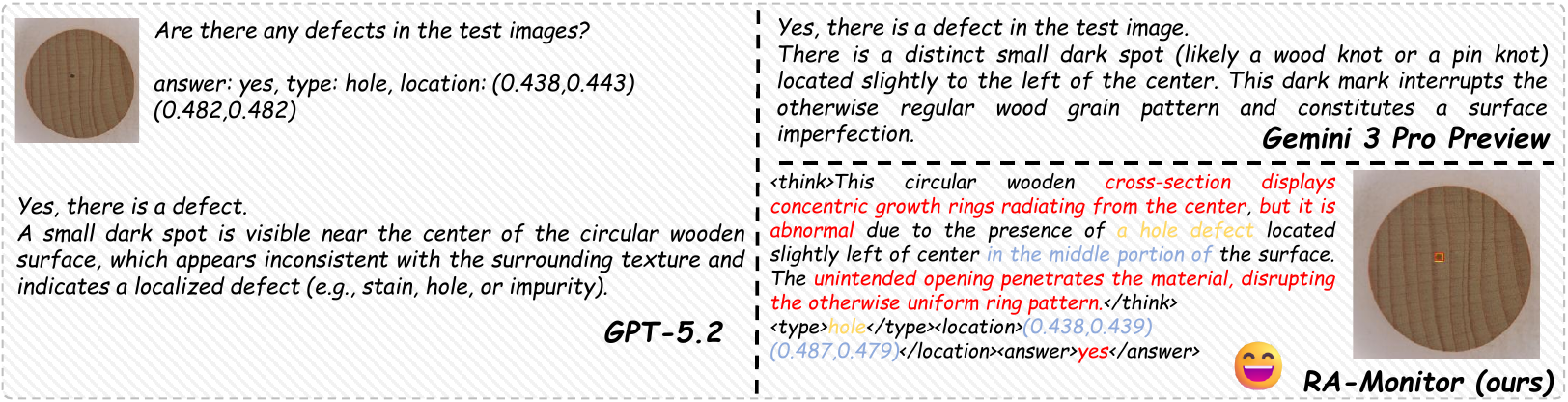}}
        \caption{
    Qualitative example on industrial woodstick inspection. The left panel shows outputs from GPT-5.2 and Gemini-3 Pro-Preview, while the right panel shows the output of RA-Monitor. Yellow boxes indicate ground-truth annotations, and red boxes denote anomaly regions predicted by RA-Monitor. Yellow boxes indicate ground-truth annotations, while red boxes denote the RA-Monitor’s predictions.
    }
  \end{center}
\end{figure}

\begin{table*}[t]
    \centering
    \caption{Surface anomaly taxonomy with fine-grained definitions.}
    \label{tab:taxonomy_surface}
    \small
    \setlength{\tabcolsep}{4pt}
    \renewcommand{\arraystretch}{1.15}
    \begin{tabularx}{\textwidth}{@{}l l l X@{}}
        \toprule
        \textbf{Level-1} & \textbf{Level-2} & \textbf{Fine-grained Type} & \textbf{Description} \\
        \midrule

        \multirow{39}{*}{\makecell[l]{Surface\\Anomaly}}
        & \multirow{6}{*}{Foreign Material}
        & debris & Loose unwanted particles or fragments (dust, chips, fibers) deposited on the surface and not originally part of the product. \\
        &  & contamination & Smears, stains, or films of unwanted substances (oil, dirt, liquid, glue, etc.) that adhere to the surface and alter its appearance. \\
        &  & foreign object intrusion & A foreign solid object partially or fully intruding into the product area (e.g., hair, metal chip) that clearly does not belong there. \\
        \cmidrule(lr){2-4}

        & \multirow{15}{*}{Irregular Surface}
        & discontinuity & A local break or interruption in an otherwise continuous surface, such as a notch, or abrupt edge. \\
        &  & indentation & A localized sunken area where the surface is pressed inward, often with a clear boundary. \\
        &  & deviation & A local surface region that visibly deviates from the intended shape or profile, even if not clearly dented or protruding. \\
        &  & pitting & Multiple small, deep pits or pinholes on the surface, often densely distributed in a region. \\
        &  & dent & A noticeable depression in the surface caused by impact or pressure, usually larger and more obvious than a minor indentation. \\
        &  & irregularity & Any non-uniform, uneven, or inconsistent surface area that does not match the expected smoothness or pattern. \\
        &  & roughness & An area where the surface texture is coarser than expected, with visible grain, bumps, or unevenness. \\
        &  & protrusion & A localized bulge or bump where material rises above the normal surface level. \\
        \cmidrule(lr){2-4}

        & \multirow{8}{*}{Material Anomaly}
        & corrosion & Material degradation with visible damage such as pits, flaking, or discoloration caused by chemical reactions (e.g., with moisture or chemicals). \\
        &  & moisture & Visible presence of water or liquid traces on or inside the surface (droplets, fogging, damp patches). \\
        &  & oxidation corrosion & Oxide layers or spots (e.g., white, grey, dark films) formed on the material due to oxidation, often spreading over time. \\
        &  & rusty & Brown or reddish corrosion products typical of rust on metal surfaces, often flaky or powdery in texture. \\
        \cmidrule(lr){2-4}

        & \multirow{10}{*}{Surface Damage}
        & tear & A ripped or torn area where the surface material is pulled apart, often with irregular, jagged edges. \\
        &  & scratch & A narrow, linear mark where the surface coating or material is cut or scraped away along a line. \\
        &  & abrasion & A worn area where the surface material has been gradually removed by friction, usually over a broader region. \\
        &  & scrape & A relatively wide area where the surface has been aggressively rubbed, removing material in a broad streak. \\
        &  & scuff & A superficial, often blurry or smudged mark caused by light rubbing, typically without deep material removal. \\
        \bottomrule
    \end{tabularx}
\end{table*}

\begin{table*}[t]
    \centering
    \caption{Structural anomaly taxonomy with fine-grained definitions.}
    \label{tab:taxonomy_structural}
    \small
    \setlength{\tabcolsep}{4pt}
    \renewcommand{\arraystretch}{1.15}
    \begin{tabularx}{\textwidth}{@{}l l l X@{}}
        \toprule
        \textbf{Level-1} & \textbf{Level-2} & \textbf{Fine-grained Type} & \textbf{Description} \\
        \midrule

        \multirow{21}{*}{\makecell[l]{Structural\\Anomaly}}
        & \multirow{5}{*}{Deformation}
        & bent & The component is visibly curved or bent away from its intended straight or flat shape. \\
        &  & \multirow{2}{*}{warping} & Gradual, non-uniform deformation where a part twists or curls, causing surfaces or edges to no longer lie in the same plane. \\
        &  & \multirow{2}{*}{distortion} & The overall geometry is visibly distorted, with shape, angles, or proportions changed from the design, not just a simple bend. \\
        \cmidrule(lr){2-4}

        & \multirow{12}{*}{Damage}
        & broken & The part is visibly separated into pieces or has a major rupture that interrupts its integrity. \\
        &  & \multirow{2}{*}{breakage} & A condition where the component has experienced breaking, leaving sharp or irregular broken edges or missing sections. \\
        &  & \multirow{2}{*}{crack} & A linear fissure in the material that may not fully separate the part but clearly breaks continuity. \\
        &  & gap & A gap appearing in edges or corners that were supposed to be smooth. \\
        &  & \multirow{2}{*}{fracture} & A structural break in the material, often more severe than a crack, with visible separation or branching crack patterns. \\
        &  & \multirow{2}{*}{fragmentation} & The component has shattered or broken into multiple small fragments instead of remaining as a single piece. \\
        &  & \multirow{2}{*}{hole} & An unintended opening that penetrates through the material or creates a significant void in the part. \\
        \cmidrule(lr){2-4}

        & \multirow{4}{*}{Separation}
        & \multirow{2}{*}{peeling} & Layers such as coating, paint, tape, or labels are lifting, curling, or coming away from the substrate. \\
        &  & \multirow{2}{*}{delamination} & Internal or inter-layer separation where bonded layers of material split apart, often visible as bubbles, steps, or layer edges. \\
        \bottomrule
    \end{tabularx}
\end{table*}

\begin{table*}[t]
    \centering
    \caption{Logical anomaly taxonomy with fine-grained definitions.}
    \label{tab:taxonomy_logical}
    \small
    \setlength{\tabcolsep}{4pt}
    \renewcommand{\arraystretch}{1.15}
    \begin{tabularx}{\textwidth}{@{}l l l X@{}}
        \toprule
        \textbf{Level-1} & \textbf{Level-2} & \textbf{Fine-grained Type} & \textbf{Description} \\
        \midrule

        \multirow{17}{*}{\makecell[l]{Logical\\Anomaly}}
        & \multirow{4}{*}{Position Errors}
        & \multirow{2}{*}{component misalignment} & The component is installed but its orientation or alignment relative to reference features is incorrect (tilted, rotated, not centered). \\
        &  & displacement & The component is shifted from its intended position, while remaining roughly parallel and correctly oriented. \\
        \cmidrule(lr){2-4}

        & \multirow{3}{*}{Completeness Errors}
        & component missing & A required part or subcomponent is completely absent from the assembly. \\
        &  & \multirow{2}{*}{quantity errors} & The number of components (e.g., screws, washers) does not match the specified quantity---too many or too few. \\
        \cmidrule(lr){2-4}

        & \multirow{6}{*}{Configuration Errors}
        & \multirow{2}{*}{wrong combination} & Incompatible or incorrect types of components are used together (e.g., mismatched parts, wrong model or material). \\
        &  & \multirow{2}{*}{layout error} & The spatial arrangement of multiple components is incorrect, such as wrong order, spacing, or relative placement. \\
        &  & \multirow{2}{*}{assembly error} & General assembly mistake, such as reversed orientation, incorrect insertion, missing fastening, or wrong connection. \\
        \cmidrule(lr){2-4}

        & \multirow{4}{*}{Specification Errors}
        & \multirow{2}{*}{size errors} & The actual size or dimension of a component visibly deviates from the required specification or tolerance (too large or too small). \\
        &  & \multirow{2}{*}{color error} & The color of a component or area does not match the specified standard (wrong hue, shade, or brightness). \\
        \bottomrule
    \end{tabularx}
\end{table*}
\end{document}